\RequirePackage{fix-cm}
\documentclass[twocolumn]{svjour3}          %
\smartqed  %
\usepackage[table,dvipsnames]{xcolor}
\usepackage{tikz}  %
\usepackage{times}
\usepackage{graphicx}
\usepackage{amsmath}
\usepackage{amssymb}
\usepackage{textcomp}
\usepackage{natbib}
\usepackage{url}
\renewcommand{\cite}{\citep}

\usepackage{subfiles}
\usepackage[normalem]{ulem}
\usepackage{pgfplots} %
\usepackage{pgfplotstable} %
\pgfplotsset{compat=newest} %
\usetikzlibrary{plotmarks} %
\usetikzlibrary{colorbrewer} %
\usepackage{multirow}
\usepackage{booktabs} %
\usepackage{etoolbox}
\usepackage[group-separator={,},output-decimal-marker = {.}, group-minimum-digits={3}]{siunitx} %

\robustify\bfseries
\robustify\itshape
\usepackage{paralist} %

\definecolor{rowblue}{RGB}{220,230,240}
\definecolor{olivegreen}{RGB}{0,170,0}
\definecolor{darkred}{RGB}{220,100,10}
\definecolor{tealblue}{RGB}{20,100,200}
\definecolor{darkishgreen}{RGB}{0,200,0}

\newcommand{\AlinaReplaced}[2]{{#2}}
\newcommand{\new}[1]{#1}

\newcommand{\oi}{Open Images}
\newcommand{\oid}{Open Images Dataset}
\newcommand{\pascal}{PASCAL}
\newcommand{\coco}{COCO}
\newcommand{\ilsvrc}{ILSVRC}

\newcommand{\ie}{i.e.\ } 
\newcommand{\eg}{e.g.\ }
\newcommand{\etc}{etc.\ } 
\newcommand{\smallsection}[1]{\paragraph*{#1}\rule{0mm}{1mm}\\[2mm]}
\newcommand{\cls}[1]{\texttt{#1}}

\newcommand{\lara}[1]{$\langle\,$#1$\,\rangle$}

\newcommand{\bigcdot}{\ \tikz[baseline=-0.5ex]{\draw[fill=black]  circle(1pt);} \ }

\newcommand{\mypar}[1]{\vspace{-0mm}\paragraph{#1}}

\newcommand{\putpin}[4]{\addplot[forget plot,mark=none] coordinates {#1} node[pin={[pin distance=#2,inner sep=0pt,outer sep=0pt,fill=white,label={[inner sep=0pt]#4}]#3:{}}]{};}
\newcommand{\p}{p}

\journalname{IJCV}
\begin{document}

\title{The Open Images Dataset V4%
}
\subtitle{Unified image classification, object detection, and visual relationship detection at scale}

\author{\mbox{Alina Kuznetsova \bigcdot
        Hassan Rom \bigcdot
        Neil Alldrin \bigcdot
        Jasper Uijlings \bigcdot
        Ivan Krasin \bigcdot
        Jordi Pont-Tuset \bigcdot}
\mbox{Shahab Kamali \bigcdot
        Stefan Popov \bigcdot
        Matteo Malloci \bigcdot
        Alexander Kolesnikov \bigcdot
        Tom Duerig \bigcdot
        Vittorio Ferrari} \includegraphics[width=0.15\linewidth]{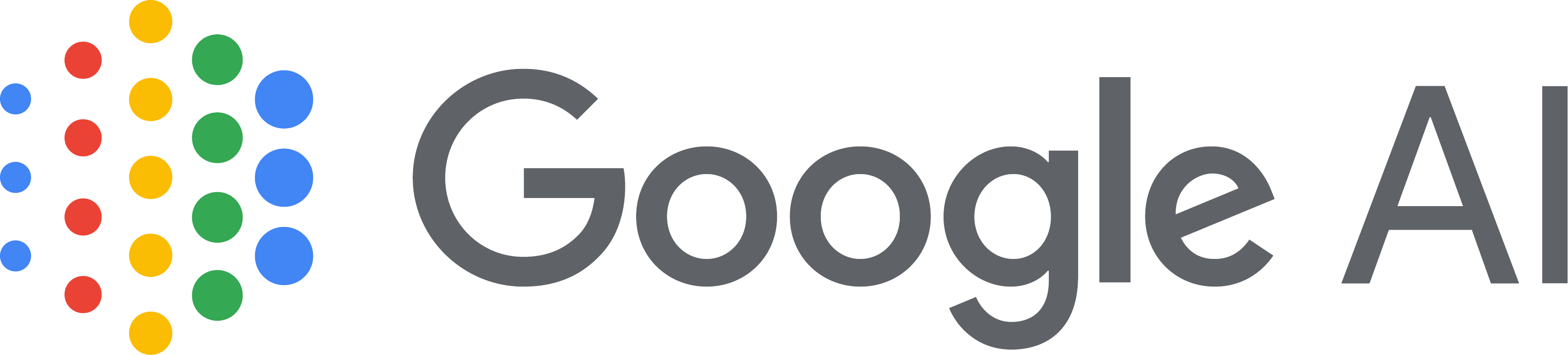} \rule{0mm}{10mm}}

\authorrunning{Kuznetsova, Rom, Alldrin, Uijlings, Krasin, Pont-Tuset, Kamali, Popov, Malloci, Kolesnikov, Duerig, and Ferrari} %
\institute{}
\date{}

\maketitle

\begin{abstract}
We present \oi{} V4, a dataset of \num{9.2}M images with unified annotations for image classification, object detection and visual relationship detection.
The images have a Creative Commons Attribution license that allows to share and adapt the material, and they have been collected from Flickr without a predefined list of class names or tags, leading to natural class statistics and avoiding an initial design bias.
\oi{} V4 offers large scale across several dimensions: \num{30.1}M image-level labels for \num{19.8}k concepts, \num{15.4}M bounding boxes for \num{600} object classes, and \num{375}k %
visual relationship annotations involving \num{57} classes.
For object detection in particular, we provide $15\times$ more bounding boxes than the next largest datasets (\num{15.4}M boxes on \num{1.9}M images).
The images often show complex scenes with several objects (8 annotated objects per image on average).
We annotated visual relationships between them, which support visual relationship detection, an emerging task that requires structured reasoning.
We provide in-depth comprehensive statistics about the dataset,
we validate the quality of the annotations,
we study how the performance of several modern models evolves with increasing amounts of training data,
\new{and we demonstrate two applications made possible by having unified annotations of multiple types coexisting in the same images.}
We hope that the scale, quality, and variety of \oi{} V4 will foster further research and innovation even beyond the areas of image classification, object detection, and visual relationship detection.
\end{abstract}

\section{Introduction}
\label{sec:intro}
Deep learning is revolutionizing many areas of computer vision.
Since its explosive irruption in the ImageNet challenge~\cite{russakovsky15ijcv} in 2012, performance of models has been improving at an unparalleled speed.
At the core of their success, however, lies the need of gargantuan amounts of annotated data to learn from.
Larger and richer annotated datasets are a boon for leading-edge research in computer vision to enable the next generation of state-of-the-art algorithms.

Data is playing an especially critical role in enabling computers to interpret images as compositions of objects, an achievement that humans can do effortlessly while it has been elusive for machines so far.
In particular, one would like machines to automatically identify what objects are present in the image (\textit{image classification}), where are they precisely located (\textit{object detection}), and which of them are interacting and how (\textit{visual relationship detection}).

\begin{figure*}
\scriptsize
\resizebox{\linewidth}{!}{%
\begin{minipage}{0.3\textwidth}
\centering
\fbox{\includegraphics[width=0.99\textwidth]{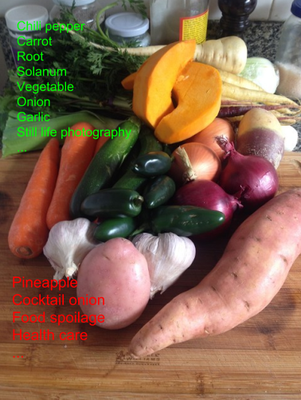}}\\[1pt]
Image classification
\end{minipage}
\begin{minipage}{0.3\textwidth}
\centering
\includegraphics[width=\textwidth]{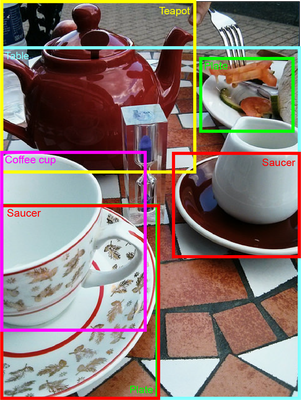}\\[1pt]
Object detection
\end{minipage}
\begin{minipage}{0.3\textwidth}
\centering
\fbox{\includegraphics[width=\textwidth]{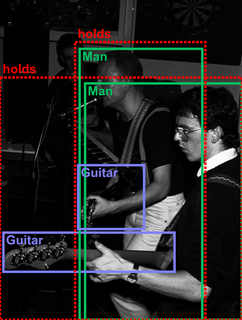}}\\[1pt]
Visual relationship detection
\end{minipage}}
\caption{\textbf{Example annotations in \oi{}} for image classification, object detection, and visual relationship detection. For image classification, positive labels (present in the image) are in green while negative labels (not present in the image) are in red. For visual relationship detection, the box with a dashed contour groups the two objects that hold a certain visual relationship.}
\label{fig:example_intro}
\end{figure*}

This paper presents the \oid{} V4, which contains images and ground-truth annotations for the three tasks above (Figure~\ref{fig:example_intro}). \oi{} V4 has several attractive characteristics, compared to previously available datasets in these areas~\cite{krizhevsky09,fei-fei:pami06,caltech256,deng09cvpr,russakovsky15ijcv,pascal-voc-2012,gupta2015arxiv,krishna17ijcv}.
The images were collected from Flickr\footnote{Image hosting service (\url{flickr.com})} without a predefined list of class names or tags, 
leading to natural class statistics and avoiding the initial design bias on what should be in the dataset.
They were released by the authors under a \textit{Creative Commons Attribution} (CC-BY) license that allows to share and adapt the material, even commercially; 
particularly so for models trained on these data, since it makes them more easily usable in any context.
Also, we removed those images that appear elsewhere in the internet to reduce bias towards \new{web image search engines}, favoring complex images containing several objects.
Complex images open the door to visual relationship detection, an emerging topic at the frontier of computer vision that requires structured reasoning of the contents of an image.
Section~\ref{sec:generation} further explains all the specifics about how we collected and annotated \oi{}.

\oi{} V4 is large scale in terms of images (\num{9178275}), annotations (\num{30113078} image-level labels, \num{15440132} bounding boxes, \num{374768} %
visual relationship triplets) and the number of visual concepts (\textit{classes}) (\num{19794} for image-level labels and \num{600} for bounding boxes). This makes it ideal for pushing the limits of the data-hungry methods that dominate the state of the art.
For object detection in particular, the scale of the annotations is unprecedented (\num{15.4} million bounding boxes for \num{600} categories on \num{1.9} million images). The number of bounding boxes we provide is more than $15\times$ greater than the next largest datasets (\coco{} and ImageNet).
Also, there are 8 annotated bounding boxes per image on average, demonstrating the complexity of the images and the richness of our annotations.
We hope this will stimulate research into more sophisticated detection models that will exceed current state-of-the-art performance and will enable assessing more precisely in which situations different detectors work best.
Section~\ref{sec:stats} provides an in-depth comprehensive set of statistics about \oi{} V4 and compare them to previous datasets.
Finally, \oi{} V4 goes beyond previous datasets also in that it is {\em unified}: the annotations for image classification, object detection, and visual relationship detection all coexist in the same set of images.
\new{This allows for cross-task training and analysis, potentially supporting deeper insights about each of the three tasks, enabling tasks that require multiple annotation types, and stimulating progress towards genuine scene understanding.}

To validate the quality of the annotations, in Section~\ref{sec:quality} we study the geometric accuracy of the bounding boxes and the recall of the image-level annotations by comparing them to annotations done by two experts and by comparing the annotators' consistency.
In Section~\ref{sec:baselines} we analyze the performance of several modern models for image classification and object detection, studying how their performance evolves with increasing amounts of training data,
\new{and we also report several baselines for visual relationship detection.}
\new{Finally, to demonstrate the value of having unified annotations, we report in Section~\ref{sec:unification-experiments} two experiments that are made possible by them (fine-grained object detection without fine-grained box labels, and zero-shot visual relationship detection).}

All the annotations, up-to-date news, box visualization tools, \etc are available on the \oi{} website: \url{https://g.co/dataset/openimages/}.
This is the first paper about \oi{}, there is no previous conference or journal version.

\section{Dataset Acquisition and Annotation}
\label{sec:generation}

This section explains how we collected the images in the \oid{}
(Sec.~\ref{sec:im_acq}), which classes we labeled (Sec.~\ref{sec:classes}),
and how we annotated (i) image-level labels (Sec.~\ref{sec:image_labels}),
(ii) bounding boxes (Sec.~\ref{sec:bboxes}), and (iii) visual relationships
(Sec.~\ref{sec:vis_rel}).

\subsection{Image Acquisition}
\label{sec:im_acq}

Images are the foundation of any good vision dataset.
The \oid{} differs in three key ways from most other datasets.
First, all images have Creative Commons Attribution (CC-BY) license and can
therefore be more easily used, with proper attribution (\eg{}in commercial
applications, or for crowdsourcing).
Second, the images are collected starting from Flickr and then removing images
that appear elsewhere on the internet.
This removes simple images that appear in search engines such as Google Image
Search, and therefore the dataset contains a high proportion of interesting,
complex images with several objects.
Third, the images are not scraped based on a predefined list of class names or
tags, leading to natural class statistics and avoiding the initial design bias
on what should be in the dataset.

The \texttildelow9 million images in the \oid{} were collected using the following procedure:

\begin{enumerate}
\item Identify all Flickr images with CC-BY license. This was done in November 2015.

\item Download the original version\footnote{In Flickr terms, images are served at different sizes (Thumbnail, Large, Medium, etc.). The Original size is a pristine copy of the image that was uploaded by the author.} of these images and generate a copy at two resolutions:

\begin{itemize}
\item \textit{1600HQ}: Images have at most \num{1600} pixels on their longest side and \num{1200} pixels on their shortest. JPEG quality of \num{90}.
\item \textit{300K}: Images have roughly \num{300000} pixels.
JPEG quality of 72.

\end{itemize}
\item Extract relevant metadata of all images to give proper attribution:

\begin{itemize}
\item \textit{OriginalURL}: Flickr direct original image url.
\item \textit{OriginalLandingURL}: Flickr image landing page.
\item \textit{License}: Image license, a subtype of CC-BY.
\item \textit{Author}: Flickr name of the author of the photo.
\item \textit{Title}: Title given by the author in Flickr.
\item \textit{AuthorProfileURL}: Link to the Flickr author profile.
\item \textit{OriginalMD5}: MD5 hash of the original JPEG-encoded image.
\end{itemize}

\item Remove images containing inappropriate content (porn, medical, violence, memes, etc.) using the safety filters on Flickr and Google SafeSearch.

\item Remove near-duplicate images, based on low-level visual similarity.

\item Remove images that appear elsewhere on the internet.
This was done for two reasons: to prevent invalid CC-BY attribution and to reduce bias towards \new{web image search engines}.

\item Recover the user-intended image orientation by comparing each original downloaded image to one of the Flickr resizes.\footnote{More details at https://storage.googleapis.com/openimages/web/2018-05-17-rotation-information.html.}

\item Partition the images into train (\num{9011219} images), validation (\num{41620}) and test (\num{125436}) splits (Tab.~\ref{tab:subsets_global}).
\end{enumerate}

\subsection{Classes}
\label{sec:classes}

\begin{figure}
\setlength{\fboxsep}{0pt}
\resizebox{\linewidth}{!}{%
\fbox{\includegraphics{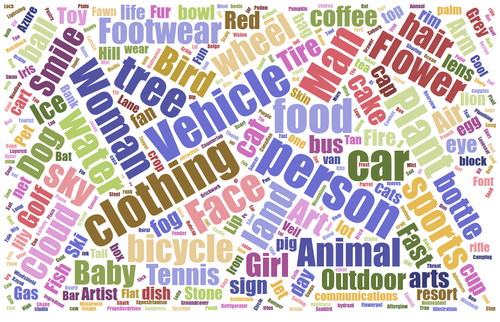}}}
\caption{\textbf{Most-frequent image-level classes}. Word size is proportional to the class counts in the train set.}
\label{fig:image_level_word_cloud}
\end{figure}

\begin{figure}
\setlength{\fboxsep}{0pt}
\resizebox{\linewidth}{!}{%
\fbox{\includegraphics{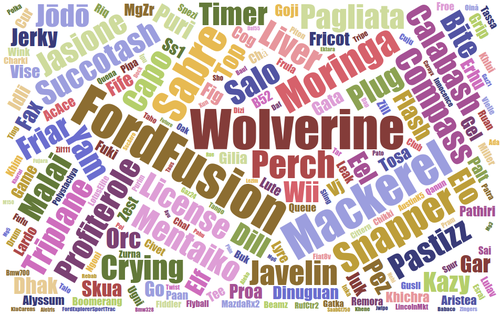}}}
\caption{\textbf{Infrequent image-level classes}. Word size is inversely proportional to the class counts in the train set.}
\label{fig:image_level_infrequent_word_cloud}
\end{figure}

The set of classes included in the \oid{} is derived from JFT, an internal dataset
at Google with millions of images and thousands of classes~\cite{hinton14nips,chollet17cvpr, sun17iccv}.
We selected \num{19794} classes from JFT, spanning a very wide range of
concepts, which serve as the image-level classes in the \oid{}:
\begin{itemize}
\item Coarse-grained object classes (\eg{}\cls{animal}).
\item Fine-grained object classes (\eg{}\cls{Pembroke} \cls{welsh} \cls{corgi}).
\item Scene classes (\eg{}\cls{sunset} and \cls{love}).
\item Events (\eg{}\cls{birthday}).
\item Materials and attributes (\eg{}\cls{leather} and \cls{red}).
\end{itemize}
An overview of the most frequent and infrequent classes is shown in Figures~\ref{fig:image_level_word_cloud} and~\ref{fig:image_level_infrequent_word_cloud}.

\begin{figure}
\resizebox{\linewidth}{!}{%
\fbox{\includegraphics[height=6cm]{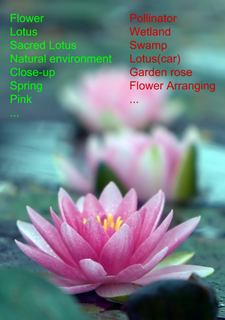}}%
\hspace{1pt}
\fbox{\includegraphics[height=6cm]{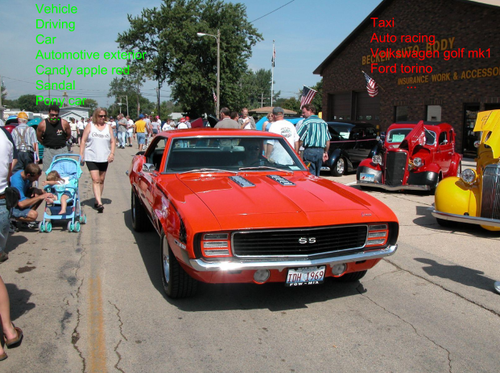}}%
}
\caption{\textbf{Examples of image-level labels}. Positive (green) and negative (red) image-level labels.}
\label{fig:image_level_examples}
\end{figure}

\begin{figure*}
\setlength{\fboxsep}{0pt}
\resizebox{\linewidth}{!}{%
\includegraphics{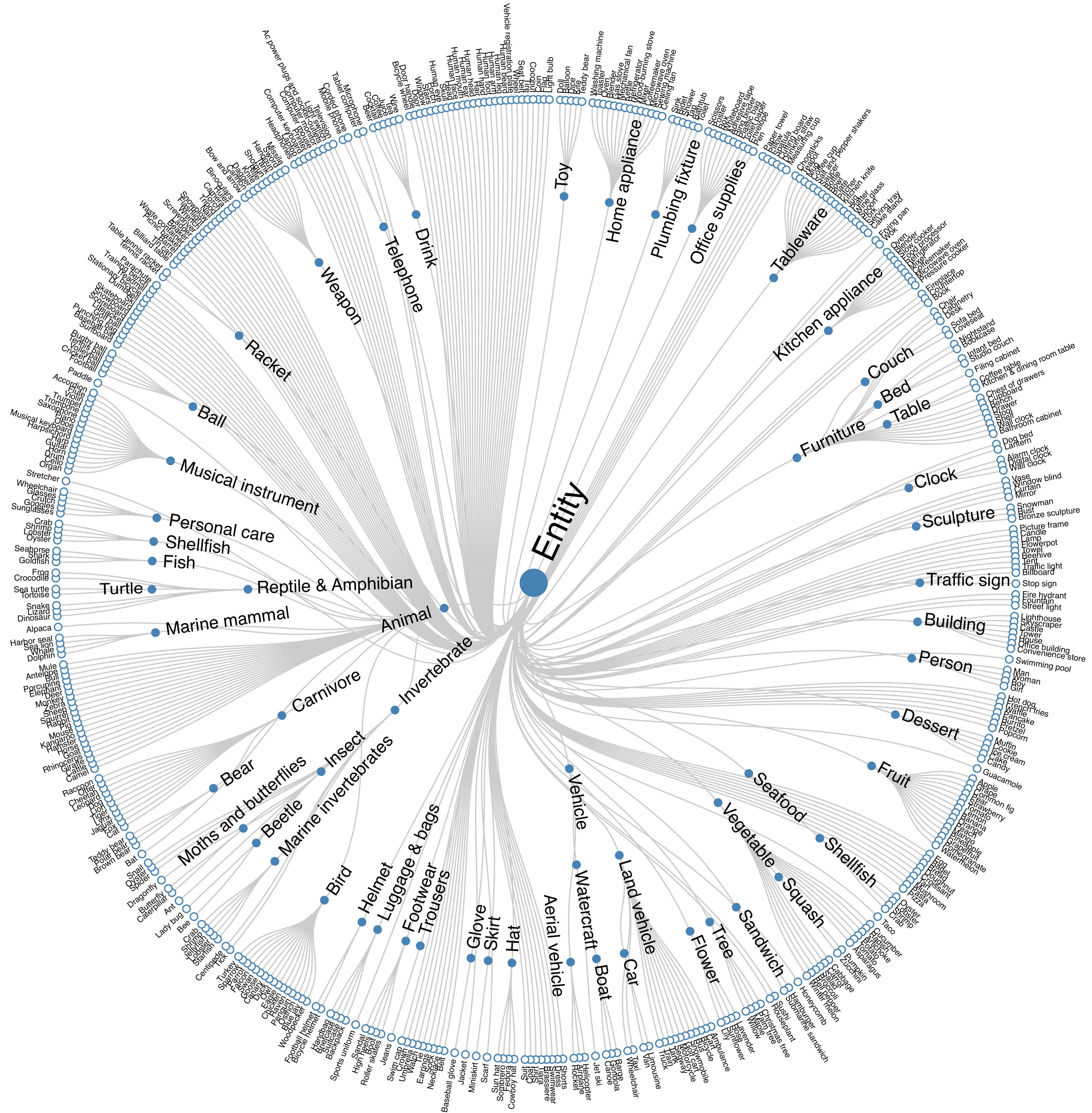}}
\caption{\new{\textbf{The boxable class hierarchy}. Parent nodes represent more generic concepts than their children.}}
\label{fig:boxable_class_hierarchy}
\end{figure*}

Out of the image-level classes, we selected \num{600} object classes we deemed important and with a clearly defined spatial extent as \textit{boxable}: these are classes for which we collect bounding box annotations (Sec.~\ref{sec:bboxes}).
A broad range of object classes are covered including animals, clothing, vehicles, food, people, buildings, sports equipment, furniture, and kitchenware.
The boxable classes additionally form a hierarchy, shown in Figure~\ref{fig:boxable_class_hierarchy}.
Figure~\ref{fig:lots_of_classes} shows two example images with a wide variety of boxable classes present.

\subsection{Image-Level Labels}
\label{sec:image_labels}

Manually labeling a large number of images with the presence or absence of \num{19794} different classes is not feasible not only because of the amount of time one would need, but also because of the difficulty for a human to learn and remember that many classes.
In order to overcome this, we apply a computer-assisted protocol.
We first apply an image classifier to generate candidate labels for all images (Sec.~\ref{sec:candidate_labels_valtest} and~\ref{sec:candidate_labels_train}),
and then ask humans to verify them (Sec.~\ref{sec:human_verification}).

For each image, this process results in several positive (the class is present) and negative (the class is absent) labels. The presence of any other label (which has not been verified) is unknown. The negative labels are therefore valuable, as they enable to properly train discriminative classifiers even in our incomplete annotation setting. We will investigate this further in Section~\ref{subsec:image_classification}.
Examples of positive and negative image-level labels are shown in Figure~\ref{fig:image_level_examples}.

\subsubsection{Candidate labels for test and validation}
\label{sec:candidate_labels_valtest}
For test and validation we generate predictions for each of the \num{19794} classes using a google-internal variant of the Inception-V2-based image classifier~\cite{szegedy16cvpr}, which is publicly available through the Google Cloud Vision API.
This model is trained on the JFT dataset -- an internal Google dataset with more than 300 million images with noisy labels~\cite{hinton14nips,chollet17cvpr}. We applied this model to the \textit{300K} resized images in the test and validation splits.
For each image, we retain all labels with a confidence score above \num{0.5} as candidates. We then ask humans to verify these candidate labels (Sec.~\ref{sec:human_verification}).

\subsubsection{Candidate labels for train}
\label{sec:candidate_labels_train}
For the train split, we generate predictions by applying dozens of image classifiers. To do this, we trained various image classification models on the JFT dataset. The classification models are Google-internal and use a variety of architectures such as Inception and ResNet families. We applied all models to each of the \textit{300K} resized images in the train split. These model predictions were used to select candidate labels to be verified by humans through stratified sampling, as explained next.

For each model we take the predictions for each image for all classes and distribute them in strata according to percentiles of their score.
We then sample a certain amount of images from each class and strata to verify.
The rationale behind this strategy is to have all ranges of classification scores represented in the verified sample.

Formally, for each class $c$, image $i$ is assigned to strata according to the logit scores output by the classifier $\bold{m}$ as:
\begin{equation}
  \mathrm{stratum}(i, c; \bold{m}) = \Bigl\lfloor\mathrm{logit}(i, c; \bold{m}) \cdot \frac{1}{w}\Bigr\rfloor
\end{equation}
where $w$ is the stratum width, $\mathrm{logit}(i, c; \bold{m})$ is the logit
score for class $c$ from model $\bold{m}$ applied to image $i$, and $\lfloor\cdot\rfloor$ is the \textit{floor} operator (i.e. rounding down to the nearest integer).
Within each stratum, we sample $k$ images to be verified.

Since we perform this process for multiple classification models, the sampling of images within each stratum is not done randomly, but by selecting the $k$ images with lowest image id\footnote{Image ids are generated based on hashes of the data so effectively the sampling within a stratum is pseudo-random and deterministic.}.
This way, the overall process results in far fewer than $m \cdot k$ verifications since there is high overlap of sampled image ids between models.
Moreover, it encourages verifying multiple different classes on the same images: the low image ids will have high probability to be sampled for many classes, while high image ids will only be sampled for rare classes with higher-confidence model predictions.

This sampling strategy yields a good variety of examples: high confidence strata lead to a mix of easy positives and hard negatives, while low confidence strata lead to a mix of hard positives and easy negatives.
We repeated this procedure for dozens of classifier models $\bold{m}$ using $w=2$ and $k=10$.\footnote{Note that while in theory logit scores are unbounded, we rarely observe values outside of $[-8,8]$ so the number of strata is bounded in practice.}

Additionally, to obtain denser annotations for the 600 boxable classes, we repeated the approach in Section~\ref{sec:candidate_labels_train} on the \num{1.74} million training images where we annotated bounding boxes (Sec.~\ref{sec:bboxes}). This generates a denser set of candidate labels for the boxable classes, to which we want to give stronger emphasis.

\subsubsection{Human verification of candidate labels}
\label{sec:human_verification}

We presented each candidate label with its corresponding image to a human annotator, who \emph{verifies} whether the class is present in that image or not.
We use two pools of annotators for such verification questions: a Google-internal pool and a crowdsource external pool.
Annotators in the Google-internal pool are trained and we can provide them with extensive guidance on how to interpret and verify the presence of classes in images.
The latter are Internet users that provide verifications through a crowdsourcing platform over which we cannot provide such training.

For each verification task, we use majority voting over multiple annotators.
We varied the number of annotators depending on the annotator pool and the difficulty of the class (which depends on how objective and clearly defined it is).
More precisely, we used the majority of 7 annotators for crowdsource annotators.
For the internal pool, we used 3 annotators for difficult classes
(\eg{}\cls{ginger beer}) and 1 annotator for easy classes (including the boxable ones).

\subsection{Bounding Boxes}
\label{sec:bboxes}

We annotated bounding boxes for the 600 boxable object classes (Sec.~\ref{sec:classes}).
In this section, we first describe the guidelines which we used to define what a good bounding box is on an object (Sec.~\ref{sec:perfect_box}).
Then we describe the two annotation techniques which we used (Secs.~\ref{sec:extreme_clicking}~and~\ref{sec:box_verification}), followed by hierarchical de-duplication (Sec.~\ref{sec:hierarchical_dedup}), and attribute annotation (Sec.~\ref{sec:attributes}).

\subsubsection{What is a perfect bounding box?}
\label{sec:perfect_box}

\begin{figure}[thpb]
  \centering
  \fbox{\includegraphics[height=3cm]{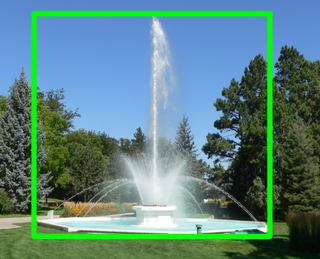}}
  \fbox{\includegraphics[height=3cm]{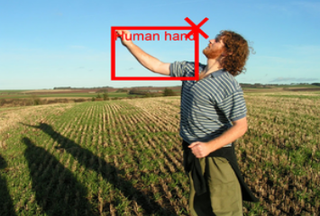}}
  \caption{\textbf{Example boxes shown to annotators}. The left shows a perfect box for \cls{fountain}. The right shows a common mistake for \cls{human hand}, caused by cultural differences. Only examples of the target class were shown before annotating that class.}
  \label{fig:perfect_boxes}
\end{figure}

As instruction, our annotators were given the following general definition: Given a target class, a perfect box is the smallest possible box that contains all visible parts of the object (Figure~\ref{fig:perfect_boxes} left). While this definition seems simple enough at first sight, there are quite a few class-dependent corner cases such as: \textit{are straps part of a camera?}, \textit{is water part of a fountain?}. Additionally, we found unexpected cultural differences, such as a \cls{human hand} including the complete human arm in some parts of the world (Figure~\ref{fig:perfect_boxes} right).
To ensure different annotators would consistently mark the same spatial extent, we manually annotated a perfect bounding box on two examples for each of the 600 object classes. Additionally, for \num{20}\% of the classes we identified common mistakes in pilot studies.
Annotators always worked on a single class at a time, and were shown the positive examples and common mistakes directly before starting each annotation session (Figure~\ref{fig:perfect_boxes}). This helps achieving high quality and consistency.

Sometimes object instances are too close to each other to put individual boxes on them.
Therefore, we also allowed annotators to draw one box around five or more heavily overlapping instances (Fig.~\ref{fig:attributes} right) and mark that box with the \cls{GroupOf} attribute (Sec.~\ref{sec:attributes}).

\subsubsection{Extreme Clicking}
\label{sec:extreme_clicking}

We annotated 90\% of all bounding boxes using extreme clicking, a fast box drawing technique introduced in~\cite{papadopoulos17iccv}.
The traditional method of drawing a bounding box~\cite{su12aaai}, used to annotate \ilsvrc{}~\cite{russakovsky15ijcv}, involves clicking on imaginary corners of a tight box around the object. This is difficult as these corners are often outside the actual object and several adjustments are required to obtain a tight box. In extreme clicking, annotators are asked to click on four physical points on the object: the top, bottom, left- and right-most points. This task is more natural and these points are easy to find.

\begin{figure}[htpb]
  \centering
  \includegraphics[width=\linewidth]{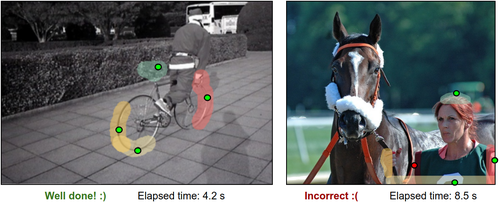}
  \caption{\textbf{Feedback during extreme clicking training}.
  Left: the annotator correctly annotated a box for \cls{bicycle}.
  Right: the annotator incorrectly annotated \cls{person} (wrong point shown in red).
  In both cases we display the valid area for each extreme point.}
  \label{fig:extreme_clicking_feedback}
\end{figure}

\mypar{Training annotators.}
We use Google-internal annotators for drawing all boxes on the \oid{}.
We found it crucial to train annotators using an automated process, in the spirit of~\cite{papadopoulos17iccv}.
Our training consists of two parts.
Part one is meant to teach extreme clicking.
Here annotators draw boxes on 10 objects for each of the 20 \pascal{} VOC classes~\cite{everingham15ijcv}.
After each class we automatically provide feedback on which boxes were correctly
or incorrectly drawn and why, by showing valid possible positions of the extreme
points (Fig.~\ref{fig:extreme_clicking_feedback}).
Part two is a qualification task in which the annotators practice both speed and
accuracy.
They are asked to draw \num{800} boxes and pass if their intersection-over-union
(IoU) with the ground truth is higher than \num{0.84} and drawing time per box is
\num{20} seconds or less.
This sets high-quality standards, as the human expert agreement is \num{0.88}~\cite{papadopoulos17iccv}.

\mypar{Annotation time.}
On average over the complete dataset, it took \num{7.4} seconds to draw a single bounding box.
This is much faster than the median time of \num{42} seconds reported for \ilsvrc{}~\cite{russakovsky15ijcv,su12aaai},
broken down into \num{25.5} seconds for drawing a box, \num{9.0} seconds for
verifying its quality, and \num{7.8} seconds for checking if other instances
needed to be annotated in the same image.
Because of our automated training and qualification stages, we found it
unnecessary to verify whether a box was drawn correctly (Sec.~\ref{sec:box_quality} for a quality analysis).
Furthermore, annotators were asked to draw boxes around \textit{all} instances
of a single class in an image consecutively, removing the separate task of
checking if other instances needed to be annotated.
Finally, extreme clicking significantly reduced the box drawing time itself
from \num{25.5} to \num{7.4} seconds.

\subsubsection{Box Verification Series}
\label{sec:box_verification}

About \num{10}\% of the bounding boxes in the training set were annotated using box verification series, in which annotators verify bounding boxes produced automatically by a learning algorithm~\cite{papadopoulos16cvpr}.
Given image-level labels, this scheme iterates between retraining the detector, relocalizing objects in the training images, and having human annotators verify bounding boxes.
The verification signal is used in two ways.
First, the detector is periodically retrained using all bounding boxes accepted so far, making it stronger.
Second, the rejected bounding boxes are used to reduce the search space of possible object locations in subsequent iterations.
Since a rejected box has an IoU $<t$ with the true bounding box, we can eliminate all candidate boxes with an IoU $\geq$ $t$ with it ($t$ is the acceptance threshold).
This is guaranteed not to remove the correct box.
This strategy is effective because it eliminates those areas of the search space that matter: high scoring locations which are unlikely to contain the object.

\begin{figure}[htpb]
\resizebox{\linewidth}{!}{%
  \fbox{\includegraphics[height=2cm]{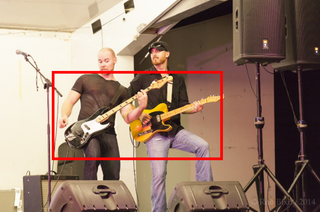}}
  \fbox{\includegraphics[height=2cm]{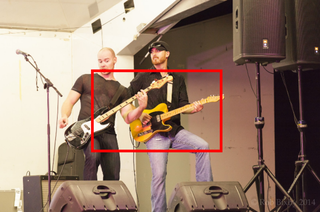}}
  \fbox{\includegraphics[height=2cm]{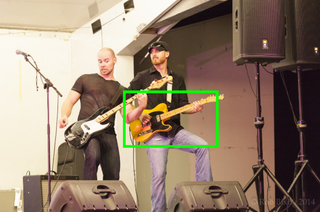}}}
  \caption{\textbf{Example of a box verification series for \cls{guitar}}.
  The highest scored \cls{guitar} box is shown to the annotator, who rejects it.
  Then the system proposes a second box, which the annotator rejects as well.
  Finally, the third proposed box is accepted and the process is completed.}
  \label{fig:box_verification_series}
\end{figure}

We adapted this general scheme to our operation in several ways.
We make up to four attempts to obtain a bounding box for a specific class in an image.
We set a higher quality criterion: we instruct annotators to accept a box if its IoU with an imaginary perfect box is greater than \new{$t=0.7$} (instead of \num{0.5} in~\cite{papadopoulos16cvpr}).
Additionally, to more efficiently use annotation time we did not verify boxes with a confidence score lower than \num{0.01}.
As detector we used Faster-RCNN~\cite{ren15nips} based on Inception-ResNet~\cite{szegedy17aaai} using the implementation of~\cite{huang17cvpr}.
We train our initial detector using the weakly-supervised technique with knowledge transfer described in~\cite{uijlings18cvpr}.
This uses image-level labels on the \oid{} and the \ilsvrc{} detection 2013 training set~\cite{russakovsky15ijcv}.
We retrained our detector several times during the annotation process based on all boxes accepted until that point in time.
Interestingly, the final detector was truly stronger than the initial one.
The annotators accepted \num{48}\% of the boxes the initial detector proposed (considering the highest-scored box for an image, if its score is $>0.01$).
This increased to \num{70}\% for the final detector.
A typical box verification series is shown in Figure~\ref{fig:box_verification_series}.

\begin{figure}[thpb]
\resizebox{\linewidth}{!}{%
  \includegraphics[height=2cm]{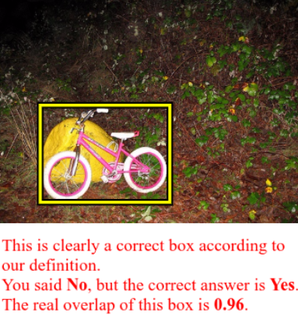}
  \includegraphics[height=2cm]{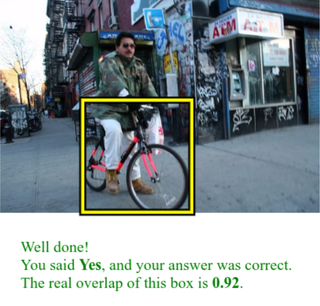}}
  \caption{\textbf{Example feedback during the training phase of box verification series}. The target class is \cls{bicycle}.}
  \label{fig:verification_feedback}
\end{figure}

\mypar{Training annotators.}
As for extreme clicking, we found it crucial to train the annotators using an automated process.
We performed several training rounds where the annotators verified 360 boxes on \pascal{} VOC 2012.
We automatically generated these boxes and calculated their IoU with respect to the ground truth.
Ideally, the annotator should accept all boxes with \mbox{IoU $>0.7$} and reject the rest.
In practice, we ignored responses on borderline boxes with \mbox{IoU $\in (0.6,0.8)$}, as these are too difficult to verify.
This helped relaxing the annotators, who could then focus on the important intervals of the IoU range ($[0.0, 0.6]$ and $[0.8,1.0]$).
To make training effective, after every 9 examples we provided feedback on which boxes were correctly or incorrectly verified and why (Figure~\ref{fig:verification_feedback}).

\begin{figure}[htpb]
  \centering
  \resizebox{\linewidth}{!}{%
 \begin{tikzpicture}
\begin{axis}[width=1.4\linewidth, height=6cm,
              ylabel=Acceptance rate,ylabel shift={-5pt},xlabel=IoU with ground truth,xlabel shift={-3pt},
              minor grid style={white!85!black},
              major grid style={white!60!black},
              xmin=0, xmax=1,
              ymin=0, ymax=1,
              legend pos=north west,
              ytick={0,0.1,...,1.1},
              minor ytick={0,0.05,...,1},
              yticklabels={0,.1,.2,.3,.4,.5,.6,.7,.8,.9,1},
              clip marker paths=true,
              enlargelimits=false,grid=both,grid style=densely dotted]
\addplot+[red,solid,mark=*, mark size=0.8, line width=1.8pt] coordinates {  (0.05,0.142631) (0.15,0.012719) (0.25,0.015280)  (0.35,0.052588)  (0.45,0.071379)  (0.55,0.149527)  (0.65,0.433089)  (0.75,0.774364)  (0.85,0.951268)  (0.95,0.988972)};
  \label{fig:annotator_training:first}
      \addlegendentry{First round}
  \addplot+[blue,solid,mark=*, mark size=0.8, line width=1.8pt] coordinates {
  (0.05,0.033937) (0.15,0.012939) (0.25,0.002302) (0.35,0.011024) (0.45,0.022703) (0.55,0.038543) (0.65,0.364333) (0.75,0.815808) (0.85,0.981436) (0.95,0.994193)};
  \label{fig:annotator_training:third}
  \addlegendentry{Third round}
    \addplot+[black,dashed,line width=1.8pt, mark=none] coordinates { (0,0.005) (0.7,0.005) (0.7,0.995) (1,0.995)};
      \label{fig:annotator_training:ideal}
      \addlegendentry{Ideal behaviour}
\end{axis}
\end{tikzpicture}}
  \caption{\textbf{IoU versus acceptance rate for the box verification training task}.
  Overall, annotators do much better after three training rounds with feedback.
  }
  \label{fig:annotator_training}
\end{figure}
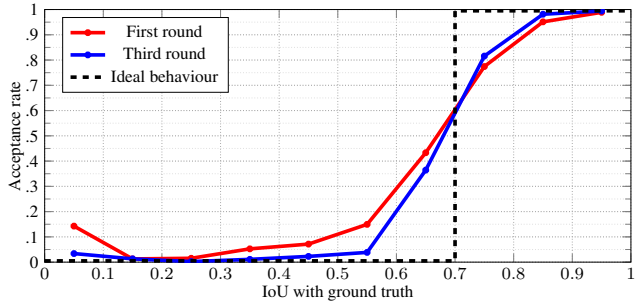

Figure~\ref{fig:annotator_training} demonstrates the importance of training.
It plots the acceptance rate versus the IoU for the first (\ref{fig:annotator_training:first}) and third (\ref{fig:annotator_training:third}) training rounds, and compares it to the ideal behavior (\ref{fig:annotator_training:ideal}).
In the first round, performance was not great: \num{15}\% of boxes with almost no overlap with the object were accepted, while only \num{95}\% of boxes with very high overlap were accepted (IoU $\in [0.8, 0.9]$).
Additionally, relatively poor poxes with IoU $\in [0.3, 0.6]$ were accepted \num{5}\% to \num{15}\% of the time.
In contrast, after three training rounds the acceptance rate of IoU \new{$\in [0.3, 0.6]$} was nicely below \num{4}\%, while high overlap boxes (IoU $\in [0.8, 0.9]$) were accepted \num{98}\% of the time.
\num{80}\% of the annotators were deemed qualified after three rounds of training.
The other \num{20}\% needed one extra training round to reach good quality.

\mypar{Annotation time.}
After a short period of time where annotators were getting used to the task, verifying a single box took \num{3.5} seconds on average.
By dividing the total number of accepted boxes by the total time spent on verifying boxes (including cases for which box verification series failed to produce a box), we measured an average time of 8.5 seconds per box produced.
This is much faster than the original annotation time for the \ilsvrc{} boxes (\num{25.5} seconds for manually drawing it, plus additional time to verify it, Sec.~\ref{sec:extreme_clicking}).

\mypar{Historical process.}
We initially annotated \num{1.5} million boxes in the training set with box verification series.
Afterwards, we co-invented extreme clicking~\cite{papadopoulos17iccv}.
Since extreme clicking takes about the same annotation time, but it is easier to deploy and delivers more accurate boxes (Sec.~\ref{sec:quality}), we used it to annotate all remaining boxes (\ie{} \num{13.1} million in the training set, and the whole validation and test sets).
Please note that in this second stage we asked annotators to draw all missing boxes for all available positive image-level labels in all images.
Hence, the final dataset has a box on an object even if box verification series failed (\ie{} after \num{4} rejected boxes).

\subsubsection{Hierarchical de-duplication}
\label{sec:hierarchical_dedup}

We annotated bounding boxes for each positively verified image-level label.
To prevent drawing two bounding boxes on the same object with two labels (\eg{}\cls{animal} and \cls{zebra}), we performed hierarchical de-duplication.
On the train set, before the box annotation process started, we removed all parents of another label already present in the set of image-level labels for a particular image.
For example, if an image had labels \cls{animal}, \cls{zebra}, \cls{car}; we annotated boxes for \cls{zebra} and \cls{car}.
On the validation and test splits we used a stricter, and more expensive, protocol.
We first asked annotators to draw all boxes for all available labels on the image.
Then we only removed a parent box (\eg{}\cls{animal}) if it overlapped with a box of a child class (\eg{}\cls{zebra}) by IoU $>0.8$.

\subsubsection{Attributes}
\label{sec:attributes}

We asked annotators to mark the following attributes if applicable:
\begin{description}
  \item[\cls{GroupOf}:] the box covers more than 5 instances of the same class which heavily occlude each other.
  \item[\cls{Partially occluded}:] the object is occluded by another object in the image.
  \item[\cls{Truncated}:] the object extends outside of the image.
  \item[\cls{Depiction}:] the object is a depiction such as a cartoon, drawing, or statue.
  \item[\cls{Inside}:] the box captures the inside of an object (\eg{}inside of an \cls{aeroplane} or a \cls{car}).

  Additionally, we marked whether boxes were obtained through box verification series (Sec.~\ref{sec:box_verification}) or through extreme clicking (Sec.~\ref{sec:extreme_clicking}).
\end{description}

\cls{Truncated} and \cls{Occluded} were also marked in \pascal{}~\cite{everingham15ijcv}.
The purpose of \cls{GroupOf} is similar to \cls{crowd} in \coco{}~\cite{lin14eccv}, but its definition is different.
In \coco{}, after having individually segmented \num{10}-\num{15} instances in an image, other instances in the same image were grouped together in a single, possibly disconnected, \cls{crowd} segment.

\subsection{Visual relationships}
\label{sec:vis_rel}

The \oid{} is rich in terms of the of number of classes and diversity of scenes, which motivated us to annotate {\em visual relationships}.
This will support research in the emerging topics of visual relationship detection~\cite{lu2016eccv,krishna17ijcv,gupta2015arxiv,dai2017cvpr}
and scene graphs~\cite{zellers18cvpr, xu2017scenegraph}.

\subsubsection{Selecting relationship triplets}
\label{ss:selecting-rels}

We explain here how we selected a set of relationship triplets to be annotated.
Each triplet has the form of \lara{\cls{class1}, relationship, \cls{class2}},
\eg{}\lara{\cls{woman}, playing, \cls{guitar}}, \lara{\cls{bottle}, on, \cls{table}}.
The challenge of selecting triplets lies in balancing several requirements:
(i) selecting frequent-enough relationships that can be found in real-world images,
(ii) generating enough data to be useful for developing future models, and
(iii) selecting non-trivial relationships that cannot be inferred from pure co-occurrence (\eg{}a \cls{car}
and a \cls{wheel} in the same image are almost always in a `part-of' relationship).

To meet these requirements, we select pairs of classes co-occurring sufficiently
frequently on the train set, and are not connected by trivial relationships, \eg{}\lara{\cls{man}, wears, \cls{shirt}}.
We also excluded all `part-of' relationships, \eg{}\lara{\cls{window}, part-of, \cls{building}}.
To make the task more interesting we make sure several relationships can connect
the same pair of objects, so the task cannot be solved simply by detecting a pair
of objects: the correct relationship between them must be recognized as well.
Finally, we make sure relationship triplets are well defined, \ie{} if we select
the triplet \lara{\cls{class1}, relationship, \cls{class2}},
then we do not include triplet \lara{\cls{class2}, relationship, \cls{class1}},
which would make it difficult to disambiguate these two triplets in evaluation.
In total, we selected \num{326} candidate triplets
After annotation we found that \num{287} of them have at least one instance in
the train split of \oi{} (Tab.~\ref{tab:vis_rel_comparison}).

Some examples of the selected relationships are: \lara{\cls{man}, hits, \cls{tennis ball}},
\lara{\cls{woman}, holds, \cls{tennis ball}}, \lara{\cls{girl}, on, \cls{horse}},
\lara{\cls{boy}, plays, \cls{drum}}, \lara{\cls{dog}, inside of, \cls{car}},
\lara{\cls{girl}, interacts with, \cls{cat}}, \lara{\cls{man}, wears, \cls{backpack}},
\lara{\cls{chair}, at, \cls{table}}.

We also introduced further attributes in the dataset, which we represent using
the `is' relationship for uniformity, \eg{}\lara{\cls{chair}, is, wooden}.
In total we consider \num{5} attributes corresponding to different material
properties (`wooden', `transparent', `plastic', `made of textil', `made of leather')
leading to \num{42} distinct \lara{object, is, attribute} triplets
(all of them turned out to have instances in the train split after verification process).

\subsubsection{Annotation process}
\label{ss:annotating-rels}

Several prior works tried different schemes for annotating visual relationships.
\cite{lu2016eccv} proposed a controlled protocol,
whereas~\cite{krishna17ijcv} gives the annotators almost complete freedom.
On \oi{} we have collected extensive bounding box annotations (Sec.~\ref{sec:bboxes}),
so we leverage them in the visual relationship annotation process as follows.
For each image and relationship triplet \lara{\cls{class1}, relationship, \cls{class2}}
we perform the following steps:
\begin{enumerate}
\item Select all pairs of object bounding boxes that can potentially be connected in this relationship triplet.
As a criterion we require that their classes match those specified in the triplet and that the two boxes overlap after enlarging them by $20\%$
\new{(our relationships assume objects to have physical contact in 3D space and consequently overlap in their 2D projections).}
\item Ask human annotators to verify that the two objects are indeed connected by this relationship.
\end{enumerate}
Note that two objects can be connected by several relationships at the same time, since they are not mutually exclusive.

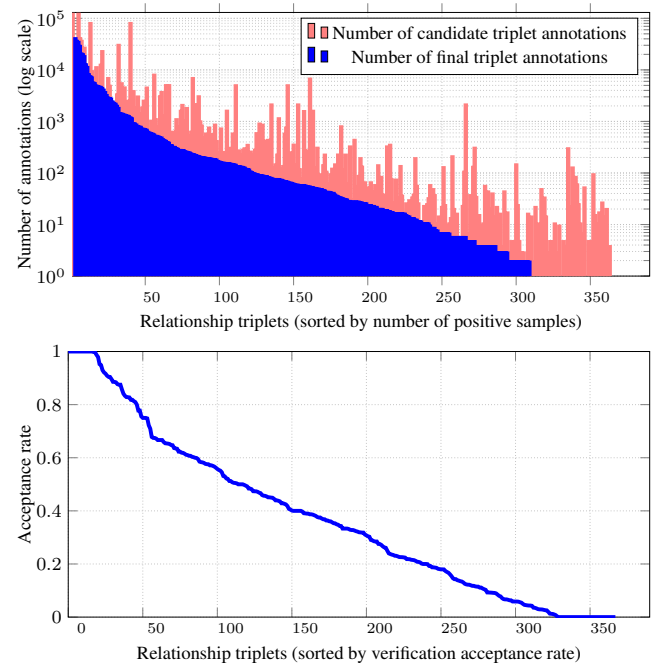
\begin{figure}[h]
\pgfplotsset{every x tick label/.append style={font=\scriptsize, xshift=0.8ex}}
  \resizebox{\linewidth}{!}{%
\begin{tikzpicture}
 \begin{axis}[ybar,width=1.3\linewidth,height=0.7\linewidth,
              ylabel=Number of annotations (log scale),ylabel shift={-5pt},
              xlabel=Relationship triplets (sorted by number of positive samples),
              ymode=log,
              xmin=1,
              xmax=390,
              ymin=0,
              ymax=130000,
              enlargelimits=false,grid=both,grid style=densely dotted]
  \addplot [ybar, bar width=2pt,draw=none, fill=red!50] table[x=index,y=total,col sep=tab] {data/vrd/num_anno_vs_acceptance.tsv};
  \addplot [ybar, bar width=2pt,draw=none, fill=blue!100] table[x=index,y=positive,col sep=tab] {data/vrd/num_anno_vs_acceptance.tsv};
  \legend{Number of candidate triplet annotations, Number of final triplet annotations}
\end{axis}
\end{tikzpicture}}
\pgfplotsset{every x tick label/.append style={font=\scriptsize, xshift=0.8ex}}
  \resizebox{\linewidth}{!}{%
\begin{tikzpicture}
\begin{axis}[width=1.3\linewidth,height=0.7\linewidth,
              ylabel=Acceptance rate,ylabel shift={-5pt},
              xlabel=Relationship triplets (sorted by verification acceptance rate),
              xmin=0,
              xmax=390,
              ymin=0,
              ymax=1,
              enlargelimits=false,grid=both,grid style=densely dotted]
  \addplot+ [blue, solid,mark=none,line width=1.8pt] table[x=index,y=rate,col sep=tab] {data/vrd/acceptance_rate.tsv};
\end{axis}
\end{tikzpicture}}
\caption{\textbf{Top:} The number of candidate relationship triplet annotations and the number of positively verified ones.
The overlap between two object bounding boxes does not guarantee that they are connected by a particular relationship. \textbf{Bottom:} Acceptance rate per distinct relationship triplet;
note that triplets with acceptance rate \num{100}\% have no more than \num{30} samples in the training split.}
\label{fig:vrd_acceptance_stats}
\end{figure}

We report the acceptance rates of the triplet verification process in Table~\ref{tab:rel_stats}.
These acceptance rates are rather low, which shows that the selected relationships are hard to predict based just on co-occurrence and spatial proximity of objects.
The acceptance rate and the total number of final annotations per relationship triplet is detailed in Figure~\ref{fig:vrd_acceptance_stats}.
On average the annotators took \num{2.6} seconds to verify a single candidate triplet.

\begin{table}[h]
\centering
\resizebox{\linewidth}{!}{%
 \begin{tabular}{lcccc}
 \toprule
 & at & holds & under & all  \\
\midrule
Acceptance rate \rule{18mm}{0mm}& $58.9\%$ & $27.9\%$ & $2.3\%$ & $28.2\%$ \\
\bottomrule
 \end{tabular}}
 \caption{\textbf{Acceptance rates for the relationship annotation process}. Displaying the relationship with the highest acceptance rate (`at'), the one with the median acceptance rate (`holds'), the one with the lowest acceptance rate (`under'), and the acceptance rate across all triplet candidates.}\label{tab:rel_stats}
\end{table}

Note that, due to the annotation process, for each pair of positive image-level labels in an image,
we annotate all relationships between all objects with those labels. Therefore, we can have multiple instances of the same relationship triplet in the same image,
connecting different pairs of objects (\eg{}different men on playing different guitars, and even different chairs {\em at the same table}, Fig.~\ref{fig:relationships}).
Note, that we excluded \cls{GroupOf} objects from the annotation process.

\section{Statistics}
\label{sec:stats}

The \oid{} consists of \num{9178275} images, split into \textit{train}, \textit{validation}, and \textit{test} (Tab.~\ref{tab:subsets_global}).

\begin{table}[h]
\centering
\resizebox{\linewidth}{!}{%
\begin{tabular}{lrrr}
\toprule
       & Train & Validation & Test\\
\midrule
Images  \rule{18mm}{0mm}&	 \num{9011219}	&    \num{41620}	&   \num{125436}\\
\bottomrule
\end{tabular}}
\vspace{2mm}
\caption{Split sizes.}
\label{tab:subsets_global}
\end{table}

As explained in Section~\ref{sec:generation}, the images have been annotated with image-level labels, bounding boxes, and visual relationships, spanning different subsets of the whole dataset.
Below we give more detailed statistics about the span, complexity, and sizes of the subsets of images annotated with human-verified image-level labels (Sec.~\ref{sec:stats:imagelevel}), with bounding boxes (Sec.~\ref{sec:stats:bbox}), and with visual relationships (Sec.~\ref{sec:stats:vis_res}).

\subsection{Human-Verified Image-Level Labels}
\label{sec:stats:imagelevel}

We assigned labels at the image level for \num{19794} classes.
Each label can be positive (indicating the class is present in the image) or negative (indicating the class is absent). %
Figure~\ref{fig:image_level_examples} shows examples and Table~\ref{tab:image_level_rated_stats} provides general statistics.

\begin{table}[h]
\centering
\resizebox{\linewidth}{!}{%
\begin{tabular}{l@{\hspace{10mm}}rrr}
\toprule
       & Train & Validation & Test \\
\midrule
Images                               &	   \num{5655108}	 &   \num{41620}   &   \num{125436}	  \\
Positive labels     \rule{4mm}{0mm}      &    \num{13444569}   &   \num{365772}  &   \num{1105052}	  \\
\rule{3mm}{0mm}\it per image & \textit{2.4} & \textit{8.8} & \textit{8.8} \\
Negative labels           &    \num{14449720}   &   \num{185618}  &   \num{562347}	      \\
\rule{3mm}{0mm}\it per image & \textit{2.6} & \textit{4.5} & \textit{4.5} \\
\bottomrule
\end{tabular}}
\caption{\textbf{Human-verified image-level labels}: Split sizes and their label count.
}
\label{tab:image_level_rated_stats}
\end{table}

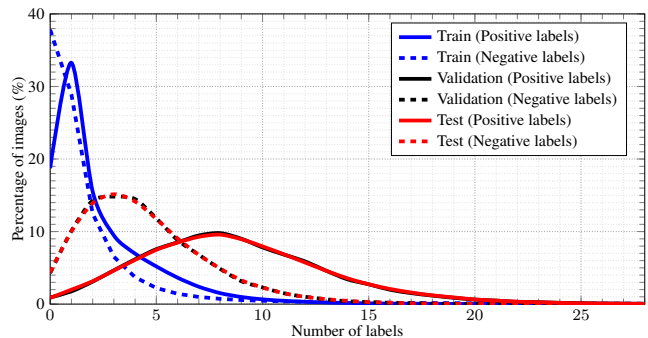
\begin{figure}[h]
\resizebox{\linewidth}{!}{%
\begin{tikzpicture}
\begin{axis}[width=1.5\linewidth, height=7cm,
              ylabel={Percentage of images (\%)},xlabel={Number of labels},
              yticklabels={0,0,10,20,30,40},
              minor xtick={0,1,...,30},
              minor ytick={0,0.01,...,0.4},
              minor grid style={white!85!black},
              major grid style={white!60!black},
              legend style={
                legend cell align=left,
              },
              legend pos = north east,
              ymax=0.4,xlabel shift={-3pt},ylabel shift={-5pt},
	      ymin=0,
              enlargelimits=false,grid=both,grid style=densely dotted]

  \addplot+[blue,solid,,smooth=false,mark=none, line width=2pt] table[x=Number_of_labels,y=Positive_labels_train] {data/image_level/image-level-label-counts.txt};
  \addlegendentry{Train (Positive labels)}
  \label{fig:image_level:label_per_image:train_pos}
  \addplot+[blue,dashed,mark=none, line width=2pt] table[x=Number_of_labels,y=Negative_labels_train] {data/image_level/image-level-label-counts.txt};
  \addlegendentry{Train (Negative labels)}
  \label{fig:image_level:label_per_image:train_neg}
  \addplot+[black,solid,smooth,mark=none, line width=2pt] table[x=Number_of_labels,y=Positive_labels_validation] {data/image_level/image-level-label-counts.txt};
  \label{fig:image_level:label_per_image:val_pos}
  \addlegendentry{Validation (Positive labels)}
  \addplot+[black,dashed,smooth,mark=none, line width=2pt] table[x=Number_of_labels,y=Negative_labels_validation] {data/image_level/image-level-label-counts.txt};
  \addlegendentry{Validation (Negative labels)}
  \addplot+[red,solid,smooth,mark=none, line width=2pt] table[x=Number_of_labels,y=Positive_labels_test] {data/image_level/image-level-label-counts.txt};
    \label{fig:image_level:label_per_image:test_pos}
  \addlegendentry{Test (Positive labels)}
  \addplot+[red,dashed,smooth,mark=none, line width=2pt] table[x=Number_of_labels,y=Negative_labels_test] {data/image_level/image-level-label-counts.txt};
  \addlegendentry{Test (Negative labels)}
\end{axis}
\end{tikzpicture}}
\caption{\textbf{Human-verified image-level labels}: Histogram of number of labels per image.
}
\label{fig:image_level:label_per_image}
\end{figure}

To further study how these labels are distributed, Figure~\ref{fig:image_level:label_per_image} shows the percentage of images with a certain number of positive (solid line) and negative (dashed line) labels.
In the train split there are 2.4 positive labels per image on average, while the validation and test splits have 8.8.
This discrepancy comes from the fact that we generated candidate labels in the validation and test splits more densely (Sec.~\ref{sec:candidate_labels_valtest}) than in the train split (Sec.~\ref{sec:candidate_labels_train}).
Please also note that the distribution of labels for validation and test are the same, since the annotation strategies are the same for both splits.

\begin{figure}[h]
\resizebox{\linewidth}{!}{%
\begin{tikzpicture}
\begin{axis}[width=1.5\linewidth, height=7cm,
              ylabel=Percentage of labels,xlabel={Class rank, sorted by frequency},
              legend pos = north east,
              legend style={
                legend cell align=left,
              },
              minor xtick={0,100,...,6000},
              minor grid style={white!85!black},
              major grid style={white!60!black},
	      xmax=6000,
              ymax=0.008,
	      ymin=0.000005,
              ymode=log,
              xlabel shift={-3pt},ylabel shift={-5pt},
              enlargelimits=false,grid=both,grid style=densely dotted]

  \addplot+[blue,solid,,smooth=false,mark=none, line width=1.8pt] table[x=Index,y=Train_Positives] {data/image_level/label-distributions.tsv};
  \addlegendentry{Train (Positive labels)}
  \addplot+[blue,dashed,mark=none, line width=1.5pt] table[x=Index,y=Train_Negatives] {data/image_level/label-distributions.tsv};
  \addlegendentry{Train (Negative labels)}
  \addplot+[black,solid,mark=none, line width=1.8pt] table[x=Index,y=Validation_Positives] {data/image_level/label-distributions.tsv};
  \addlegendentry{Validation (Positive labels)}
  \addplot+[black, dashed,mark=none, line width=1.5pt] table[x=Index,y=Validation_Negatives] {data/image_level/label-distributions.tsv};
  \addlegendentry{Validation (Negative labels)}
  \addplot+[red,solid,mark=none, line width=1.8pt] table[x=Index,y=Test_Positives] {data/image_level/label-distributions.tsv};
  \addlegendentry{Test (Positive labels)}
  \addplot+[red,dashed,mark=none, line width=1.5pt] table[x=Index,y=Test_Negatives] {data/image_level/label-distributions.tsv};
  \addlegendentry{Test (Negative labels)}
\end{axis}
\end{tikzpicture}}
\caption{\textbf{Percentage of human-verified image-level labels for each class}. The horizontal axis represents the rank of each class when sorted by frequency, the vertical axis is in logarithmic scale.}
\label{fig:image_level:label_distribution}
\end{figure}
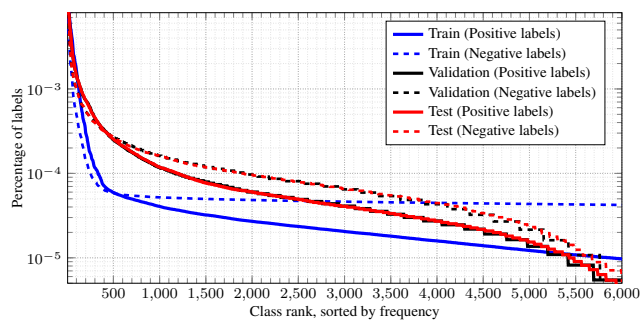

Some classes are more commonly captured in images than others, and this is also reflected in the counts of annotated labels for different classes.
Figure~\ref{fig:image_level:label_distribution} shows the percentage of labels for the top \num{6000} classes (sorted by decreasing frequency). As expected, the \texttildelow300 most frequent classes cover the majority of the samples for all three splits of the dataset. %

As mentioned in Section~\ref{sec:human_verification}, label verification is done by annotators from two different pools: internal and crowd-sourced.
Table~\ref{tab:human_annotator_pool_stats} shows the number of human-verified labels coming from each pool.
We can see that in train, crowd-sourced labels represent about 20\% of all verified labels, whereas for validation and test, they represent less than 1\%.

\begin{table}[h]
\centering
\resizebox{\linewidth}{!}{%
\begin{tabular}{l@{\hspace{5mm}}rrr}
\toprule
       & Train & Validation & Test \\
\midrule
Internal annotators      \rule{4mm}{0mm}  &  \num{22351016}  &  \num{547291}  &  \num{1655384} \\
Crowd-source annotators                   &  \num{5543273}   &  \num{4099}    &  \num{12015}   \\
\bottomrule
\end{tabular}}
\caption{\textbf{Internal versus crowd-source human-verified image-level labels}: Number of image-level labels (positive and negative) coming from the two pools of human annotators.}
\label{tab:human_annotator_pool_stats}
\end{table}

\subsection{Bounding Boxes}
\label{sec:stats:bbox}

\smallsection{General statistics}
We annotated bounding boxes around objects of \num{600} boxable classes on the whole validation and test splits, and on a subset of the train split (Tab.~\ref{tab:subsets_bboxes}).

\begin{table}[h]
\centering
\resizebox{\linewidth}{!}{%
\begin{tabular}{lrrr}
\toprule
       & Train & Validation & Test \\
\midrule
Images   \rule{20mm}{0mm}    &	 \num{1743042}	&    \num{41620}	&   \num{125436}	   \\
Boxes                       &	\num{14610229}	&   \num{204621}	&   \num{625282}	   \\
\rule{3mm}{0mm}\it per image & \textit{8.4} & \textit{4.9} & \textit{5.0} \\

\bottomrule
\end{tabular}}
\caption{\textbf{Split sizes with annotated bounding boxes}. For each split, number of images and boxes (also normalized per image in italics). These statistics are only over the 600 boxable classes.
}
\label{tab:subsets_bboxes}
\end{table}

Table~\ref{tab:size_comparison} shows the number of classes, images and bounding boxes in the \oid{} compared to  other well-known datasets for object detection: \coco{}~\cite{lin14eccv} (2017 version), \pascal{}~\cite{everingham15ijcv} (2012 version), and \ilsvrc{}~\cite{russakovsky15ijcv} (2014 detection version).
In this comparison we only consider images in \oi{} with bounding boxes, not the full dataset. As it is common practice, we ignore the objects marked as difficult in \pascal{}.
As the table shows, \oid{} is much larger than previous datasets and offers $17\times$ more object bounding boxes than \coco{}. Moreover, it features complex images with several objects annotated (about the same as COCO on average).

\begin{table}[h]
\centering
\resizebox{\linewidth}{!}{%
\begin{tabular}{lrrrrr}
\toprule
       & \pascal{} & \coco{} & \multicolumn{2}{r}{\ilsvrc{}-Det} & \oi{} \\
       &  &  & All & Dense &\\
\midrule
Classes     &      \num{20} & \num{80} & \num{200} & \num{200} & \num{600} \\
Images      &   \num{11540} & \num{123287} & \num{476688}& \num{80462}& \num{1910098}\\
Boxes       &   \num{27450} & \num{886284} & \num{534309}& \num{186463} & \num{15440132}\\
Boxes/im.  &   \num{2.4} & \num{7.2} & \num{1.1} & \num{2.3} &\num{8.1}\\
\bottomrule
\end{tabular}}
\caption{\textbf{Global size comparison to other datasets}. We take the dataset splits with publicly available ground truth, that is, \texttt{train+val} in all cases except \oi{}, where we also add the \texttt{test} set which is publicly available. Please note that in \ilsvrc{} train, only a subset of $\sim$\num{80000} images are densely annotated with all 200 classes ($\sim$\num{60000} train and $\sim$\num{20000} validation). The other images are more sparsely annotated, with mostly one class per image.}
\label{tab:size_comparison}
\end{table}

\begin{figure}[h]
\input{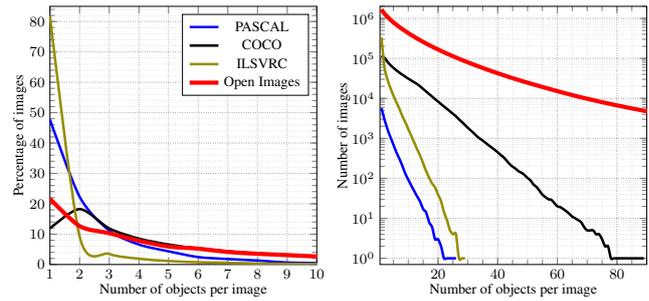}
\caption{\textbf{Number of objects per image}.
Percentage of images with exactly a certain number of objects (left).
Number of images with at least a certain number of objects (right).
Train set for all datasets.}
\label{fig:num_objs_hist}
\end{figure}

To further study how objects are distributed over the images, Figure~\ref{fig:num_objs_hist} (left) counts the percentage of images with a certain number of bounding boxes.
We can observe that \coco{} and \oi{} are significantly less biased towards single-object images.
Figure~\ref{fig:num_objs_hist} (right) displays the number of images that contain at least a certain number of bounding boxes.
\oi{} has significantly more images than the other datasets in the whole the range of number of boxes per image,
and especially so at high values, where it covers some regime unexplored before (more than 80 bounding boxes per image, up to 742).

\begin{figure*}
\setlength{\fboxsep}{0pt}
\resizebox{\linewidth}{!}{%
\fbox{\includegraphics[height=3cm]{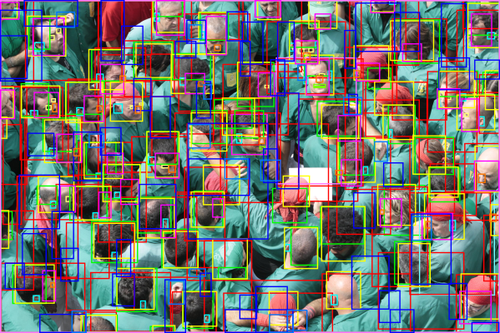}}%
\hspace{1mm}
\fbox{\includegraphics[height=3cm]{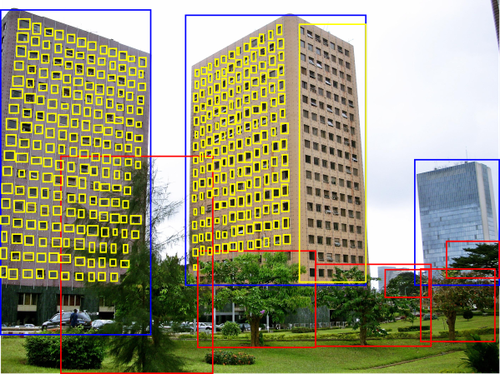}}%
\hspace{1mm}
\fbox{\includegraphics[height=3cm]{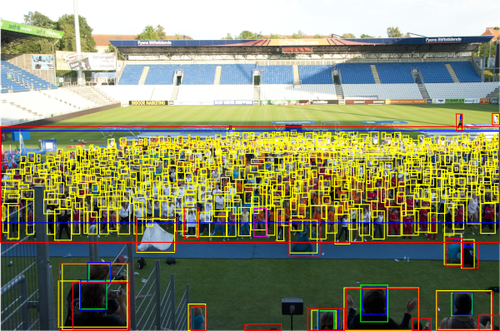}}%
}
\caption{\textbf{Examples of large number of annotated boxes}: Images with 348, 386, and 743, respectively.
\cls{GroupOf} could have been used in many of these cases, but nevertheless they still have interest in practice.}
\label{fig:lots_instances}
\end{figure*}

\begin{figure*}
\setlength{\fboxsep}{0pt}
\resizebox{\linewidth}{!}{%
\fbox{\includegraphics[height=3cm]{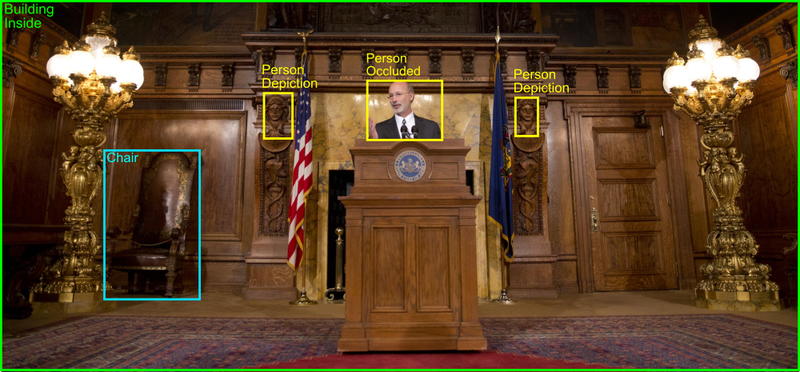}}%
\hspace{1mm}
\fbox{\includegraphics[height=3cm]{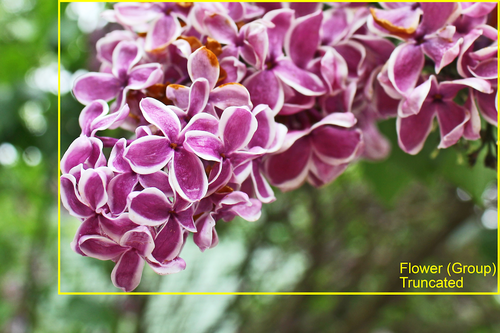}}%
}
\caption{\textbf{Examples of box attributes}: \cls{GroupOf}, \cls{Occluded}, \cls{Depiction}, \cls{Truncated}, and \cls{Inside}.}
\label{fig:attributes}
\end{figure*}

Figure~\ref{fig:lots_instances} shows some images with a large number of bounding boxes (348, 386, and 743, respectively).
In many of these cases, the \cls{GroupOf} attribute could have been used to reduce the annotation time (\eg{}the set of windows marked as \cls{GroupOf} on the right face of the center building).
Having some of these extreme cases exhaustively annotated, however, is also useful in practice.

\smallsection{Box attributes statistics}
As explained in Section~\ref{sec:attributes}, the bounding boxes in \oi{} are also labeled with five attributes. Table~\ref{tab:att_frequency} shows the frequency of these attributes in the annotated bounding boxes.

\begin{table}[h]
\centering
\resizebox{\linewidth}{!}{%
\begin{tabular}{lrrrrr}
\toprule
Attribute  & Occluded & Truncated & GroupOf & Depiction & Inside \\
\midrule
Frequency     & \num{66.06}\% & \num{25.09}\% & \num{5.99}\% & \num{5.45}\% & \num{0.24}\% \\
\bottomrule
\end{tabular}}
\caption{\textbf{Frequency of attributes}: Percentage of boxes with the five different attributes on \oi{} train.
As reference, the \cls{Crowd} attribute in \coco{} is present in \num{1.17}\% of their boxes.}
\label{tab:att_frequency}
\end{table}

\cls{Occluded} and \cls{Truncated} are the most common attributes, with a considerable portion of the objects being marked as such.
\cls{GroupOf} and \cls{Depiction} are still used in a significant proportion, whereas \cls{Inside} is rare.

Figure~\ref{fig:attributes} displays two images containing boxes labeled with
each of the five available attributes.
In the left image, the building is viewed from inside, the person is occluded by
the stand, and the two busts are depictions of people.
The right image shows a group of flowers that is truncated by the picture framing.

\smallsection{Box class statistics}
Not all object classes are equally common and equally captured in pictures, so the classes in \oi{} are not uniformly distributed in their number of instances and through the images.
We study both effects below.

\begin{figure}[h]
\resizebox{\linewidth}{!}{%
\begin{tikzpicture}
\begin{axis}[width=1.35\linewidth, height=6.5cm,
              ylabel=Number of boxes,ylabel shift={-5pt},xlabel=Class sorted position,xlabel shift={-3pt},
              ytick={0.1,1,10,100,1e3,1e4,1e5,1e6,1e7,1e8},
              legend pos = north east,
              clip marker paths=true,
              ymode=log,
              xmode=log,
              ymax=1e8,
              enlargelimits=false,grid=both,grid style=densely dotted]
  \putpin{(1,10.2*146102.30)}{8.5mm}{55}{Clothing}
  \putpin{(2,10.2*146102.30)}{1mm}{120}{Man}
  \putpin{(3,7.5*146102.30)}{5mm}{110}{Human face}
  \putpin{(4,7.1*146102.30)}{1.5mm}{85}{Person};
  \putpin{(5,5.6*146102.30)}{6mm}{68}{Tree}
  \putpin{(6,5.5*146102.30)}{5mm}{30}{Woman}
  \putpin{(7,5.3*146102.30)}{8mm}{2}{Footwear}

  \putpin{(1,257253)}{5mm}{12}{Person}

  \putpin{(1,74517)}{8mm}{7}{Dog}
  \putpin{(2,60255)}{7mm}{15}{Person}

  \putpin{(1,4194)}{5mm}{20}{Person}
  \putpin{(2,1178)}{3mm}{45}{Chair}

  \addplot+[blue,solid,mark=*, mark size=0.8, line width=1.5pt] table[x=Position,y=Count] {data/class_distrib/ClassDistrib_PASCAL_VOC2012_Main_train.txt};
  \addlegendentry{\pascal{}}
  \addplot+[black,solid,mark=*, mark size=0.8, line width=1.5pt] table[x=Position,y=Count] {data/class_distrib/ClassDistrib_COCO_train2017.txt};
  \addlegendentry{\coco{}}
    \addplot+[olive,solid,mark=*, mark size=0.8, line width=1.5pt] table[x=Position,y=Count] {data/class_distrib/ClassDistrib_ILSVRC-DET_train2014.txt};
  \addlegendentry{\ilsvrc{}}
  \addplot+[red,solid,mark=*, mark size=0.8, line width=2.5pt, mark options={fill=red}] table[x=Position,y=Count] {data/class_distrib/ClassDistrib_OpenImagesV4_train.txt};
\addlegendentry{\oi{}}
\label{fig:gt_class_distrib:oi}
\end{axis}
\end{tikzpicture}}
\caption{\textbf{Number of boxes per class}.
The horizontal axis is the rank of each class when sorted by the number of boxes,
represented in logarithmic scale for better readability.
We also report the name of the most common classes.
Train set for all datasets.}
\label{fig:gt_class_distrib}
\end{figure}
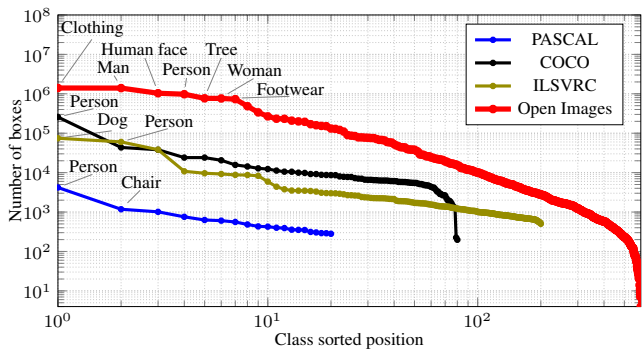

Figure~\ref{fig:gt_class_distrib} (\ref{fig:gt_class_distrib:oi}) plots the the
number of boxes annotated for each class, sorted by increasing frequency.
In order to visually compare to the other datasets with fewer classes, the horizontal axis is shown in logarithmic scale.

\oi{} is generally an order of magnitude larger than the other datasets.
There are 11 classes in \oi{} with more samples than the largest class in \coco{}.
As a particular example, the \cls{person} class has \num{257253} instances in \coco{}, while \oi{} has \num{3505362} instances of the agglomeration of classes referring to person (\cls{person}, \cls{woman}, \cls{man}, \cls{girl}, \cls{boy})\footnote{These are really unique objects: Each object is annotated only with its \textit{leafmost} label, \eg{}a \cls{man} has a single box, it is not annotated as \cls{person} also.}.

At the other end of the spectrum, \oi{} has 517 classes with more instances than the most infrequent class in \coco{} (198 instances), and 417 classes in the case of \ilsvrc{} (502 instances).

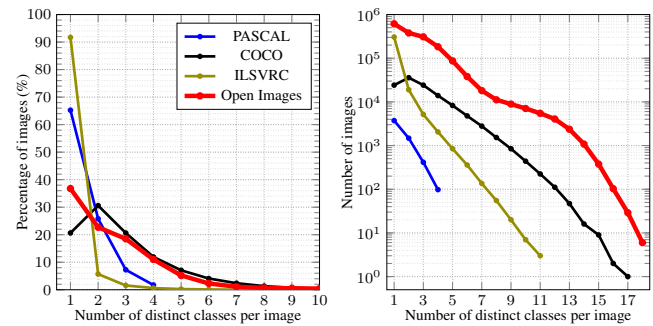
\begin{figure}[h]
\resizebox{\linewidth}{!}{%
\begin{tikzpicture}
\begin{axis}[width=0.8\linewidth, height=7cm,
              ylabel=Percentage of images (\%),ylabel shift={-8pt},xlabel=Number of distinct classes per image,xlabel shift={-3pt},
              legend pos = north east,
              minor grid style={white!85!black},
              major grid style={white!60!black},
              xmin=0.5, xmax=10,
              ymin=0, ymax=100,
              ytick={0,10,...,100},
              xtick={1,2,...,17},
              minor ytick={0,2,...,100},
              clip marker paths=true,
              enlargelimits=false,grid=both,grid style=densely dotted]
  \addplot+[blue,solid,mark=*, mark size=0.8, line width=1.5pt] coordinates {
 (1,65.244009) (2,25.800245) (3,7.241560) (4,1.714186)
};
  \addlegendentry{\pascal{}}
  \addplot+[black,solid,mark=*, mark size=0.8, line width=1.5pt] coordinates {
 (1,20.624904) (2,30.578343) (3,20.610407) (4,11.915645) (5,7.063428) (6,4.078761) (7,2.379206) (8,1.303873) (9,0.721437) (10,0.374363) (11,0.191019) (12,0.094657) (13,0.040080) (14,0.013644) (15,0.007675) (16,0.001706) (17,0.000853)
};
  \addlegendentry{\coco{}}

  \addplot+[olive,solid,mark=*, mark size=0.8, line width=1.5pt, mark options={fill=olive}] coordinates {
   (1,91.713297) (2,5.693997) (3,1.548247) (4,0.616540) (5,0.253993) (6,0.107655) (7,0.040783) (8,0.016493) (9,0.005997) (10,0.002099) (11,0.000900)
};
\addlegendentry{\ilsvrc{}}

  \addplot+[red,solid,mark=*, mark size=0.8, line width=2.5pt, mark options={fill=red}] coordinates {
   (1,36.772774) (2,22.711543) (3,18.508519) (4,10.991653) (5,5.207475) (6,2.281339) (7,1.089748) (8,0.669494) (9,0.534264) (10,0.423791) (11,0.328303) (12,0.243425) (13,0.142063) (14,0.064738) (15,0.022598) (16,0.006174) (17,0.001738) (18,0.000360)
};
\addlegendentry{\oi{}}
\end{axis}
\end{tikzpicture}
\begin{tikzpicture}
\begin{axis}[width=0.8\linewidth, height=7cm,
              ylabel=Number of images,ylabel shift={-5pt},xlabel=Number of distinct classes per image,xlabel shift={-3pt},
              minor grid style={white!85!black},
              major grid style={white!60!black},
              xmin=0.5, xmax=18.5,
              ymin=0.5, ymax=1e6,
			  ymode=log,
              xtick={1,3,...,18},
			  ytick={0,1,10,1e2,1e3, 1e4, 1e5, 1e6},
              clip marker paths=true,
              enlargelimits=false,grid=both,grid style=densely dotted]
  \addplot+[blue,solid,mark=*, mark size=0.8, line width=1.5pt] coordinates {
  (1,3730.00) (2,1475.00) (3,414.00) (4,98.00)
};
  \addplot+[black,solid,mark=*, mark size=0.8, line width=1.5pt] coordinates {
(1,24186.00) (2,35858.00) (3,24169.00) (4,13973.00) (5,8283.00) (6,4783.00) (7,2790.00) (8,1529.00) (9,846.00) (10,439.00) (11,224.00) (12,111.00) (13,47.00) (14,16.00) (15,9.00) (16,2.00) (17,1.00)
};
  \addplot+[olive,solid,mark=*, mark size=0.8, line width=1.5pt, mark options={fill=olive}] coordinates {
(1,305840.00) (2,18988.00) (3,5163.00) (4,2056.00) (5,847.00) (6,359.00) (7,136.00) (8,55.00) (9,20.00) (10,7.00) (11,3.00)
};

  \addplot+[red,solid,mark=*, mark size=0.8, line width=2.5pt, mark options={fill=red}] coordinates {
(1,613471.00) (2,378891.00) (3,308773.00) (4,183371.00) (5,86875.00) (6,38059.00) (7,18180.00) (8,11169.00) (9,8913.00) (10,7070.00) (11,5477.00) (12,4061.00) (13,2370.00) (14,1080.00) (15,377.00) (16,103.00) (17,29.00) (18,6.00)
};
\end{axis}
\end{tikzpicture}}
\caption{\textbf{Number of distinct classes per image}.
Normalized (left) and unnormalized (right) histogram of the number of distinct classes per image.
Train set for all datasets.}
\label{fig:co_occur_hist}
\end{figure}

Interactions between different object classes are also a reflection of the richness of the visual world. Figure~\ref{fig:co_occur_hist} (left) reports the percentage of images with boxes coming from a varying number of distinct classes.
We can see that \oi{} and \coco{} have a much richer distribution of images with co-occurring classes compared to \ilsvrc{} and \pascal{}, which are more biased to a single class per image.

\begin{figure}[h]
\setlength{\fboxsep}{0pt}
\resizebox{\linewidth}{!}{%
\fbox{\includegraphics[height=3cm]{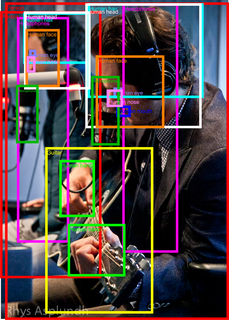}}%
\hspace{0.1pt}
\fbox{\includegraphics[height=3cm]{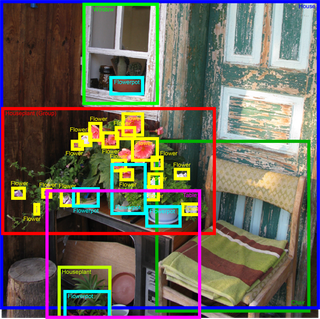}}%
}
\caption{Images with a large number of different classes annotated (11 on the left, 7 on the right).}
\label{fig:lots_of_classes}
\end{figure}

Figure~\ref{fig:co_occur_hist} (right) shows the unnormalized statistics (i.e. with the number of images instead of the percentage). It shows that \oi{} has at least one order of magnitude more images than \coco{} at any point of the curve.
As an example, \oi{} has about \num{1000} images with \num{14} distinct classes, while \coco{} has 20; \ilsvrc{} has no image with more than 11 classes, and \pascal{} no more than 4.
Figure~\ref{fig:lots_of_classes} displays two images with a large number of classes annotated, to illustrate the variety and granularity that this entails.

As further analysis, we compute the class co-occurrence matrix in \oi{} and sort the pairs of classes in decreasing order. We observe the following patterns. The most co-occurring pairs are human-related classes (\cls{Person}, \cls{Man}, \cls{Woman}) with their parts (\cls{Human face}, \cls{Human arm}, \cls{Human hair}, \cls{Human nose}, etc.) or with accessories (\cls{Clothing}, \cls{Footwear}, \cls{Glasses}, etc.); and other types of objects and their parts (\cls{Car}-\cls{Wheel}, \cls{House}-\cls{Window}).
Other interesting object pairs co-occurring in more than \num{100} images are \cls{Drum}-\cls{Guitar}, \cls{Chair}-\cls{Desk}, \cls{Table}-\cls{Drink}, \cls{Person}-\cls{Book}.
Please note that objects co-occurring in an image does not imply them being in any particular visual relationship (analyzed in Sec.~\ref{sec:stats:vis_res}).

\smallsection{Box size statistics}

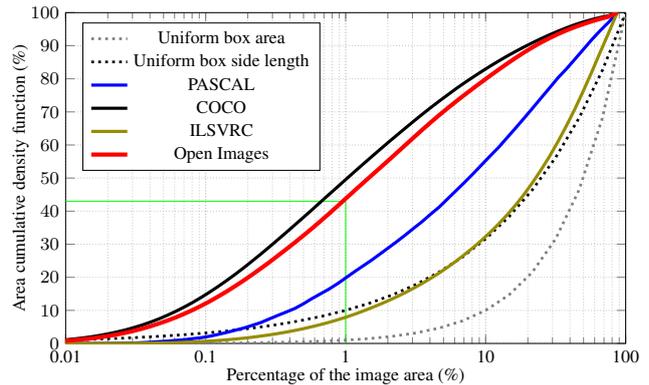
\begin{figure}[h]
\resizebox{\linewidth}{!}{%
\begin{tikzpicture}
\begin{axis}[width=1.3\linewidth,height=0.85\linewidth,
              ylabel=Area cumulative density function (\%),ylabel shift={-5pt},xlabel=Percentage of the image area (\%),xlabel shift={-3pt},
              ytick={0,10,...,100},
              xticklabels={0.001, 0.01,0.1,1,10,100},
              xmin=0.01,
              xmax=100,
              ymin=0,
              ymax=100,
              xmode=log,
              legend pos = north west,
              enlargelimits=false,grid=both,grid style=densely dotted]
  \draw [green,solid]({axis cs:1,0}) -- ({axis cs:1,43});
  \addplot[forget plot,green,solid,domain=0.01:1]{43};
  \label{fig:gt_areas:percentile}
    \addplot[gray, line width=1.3pt,dotted,domain=0.01:100]{x};
  \label{fig:gt_areas:uniform_area}
  \addlegendentry{Uniform box area}
  \addplot[black, line width=1.3pt,dotted,domain=0.01:100]{((x/100)^0.5)*100};
  \label{fig:gt_areas:uniform_side}
  \addlegendentry{Uniform box side length}
  \addplot+[blue,solid,mark=none, line width=1.5pt] table[x=X,y=F] {data/area_distrib/AreaDistrib_PASCAL_VOC2012_Main_train.txt};
  \addlegendentry{\pascal{}}
  \addplot+[black,solid,mark=none, line width=1.5pt] table[x=X,y=F] {data/area_distrib/AreaDistrib_COCO_train2017.txt};
  \addlegendentry{\coco{}}
    \addplot+[olive,solid,mark=none, line width=1.5pt] table[x=X,y=F] {data/area_distrib/AreaDistrib_ILSVRC-DET_train2014.txt};
  \addlegendentry{\ilsvrc{}}
  \addplot+[red,solid,mark=none, line width=2pt] table[x=X,y=F] {data/area_distrib/AreaDistrib_OpenImagesV4_train.txt};
  \addlegendentry{\oi{}}
\end{axis}
\end{tikzpicture}}
\caption{\textbf{Annotated objects area}: Cumulative distribution of the
percentage of image area occupied by the annotated objects of \pascal{}, \coco{}, and \oi{}; \ie{}, percentage of instances whose area is below a certain value.
As a baseline, we plot the function corresponding to boxes with uniformly distributed area and side length.
We ignore here boxes marked as crowd in \coco{} and marked as group in \oi{}.
Train set for all datasets.}
\label{fig:gt_areas}
\end{figure}

Figure~\ref{fig:gt_areas} displays the cumulative density function of the bounding box sizes in \oi{}, \pascal{}, \coco{}, and \ilsvrc{}.
The function represents the percentage of bounding boxes (vertical axis) whose area is below a certain percentage of the image area (horizontal axis).
As an example, the green lines (\ref{fig:gt_areas:percentile}) show that \num{43}\% of the bounding boxes in \oi{} occupy less than \num{1}\% of the image area.
Hence, the \oid{} offers a real challenge for object detection, supporting the development and evaluation of future detection models and algorithms.

We compare to two uniform distribution baselines: boxes with uniform area (\ref{fig:gt_areas:uniform_area}) or with uniform side length (\ref{fig:gt_areas:uniform_side}) (i.e.  the square root of their area is uniformly distributed).
Interestingly, \ilsvrc{} closely follows the distribution of the latter.
In contrast, the other datasets have a greater proportion of smaller objects:
\oi{} has a similar distribution to \coco{}, both having many more small objects than \pascal{}.

\begin{figure}
    \includegraphics[width=\linewidth]{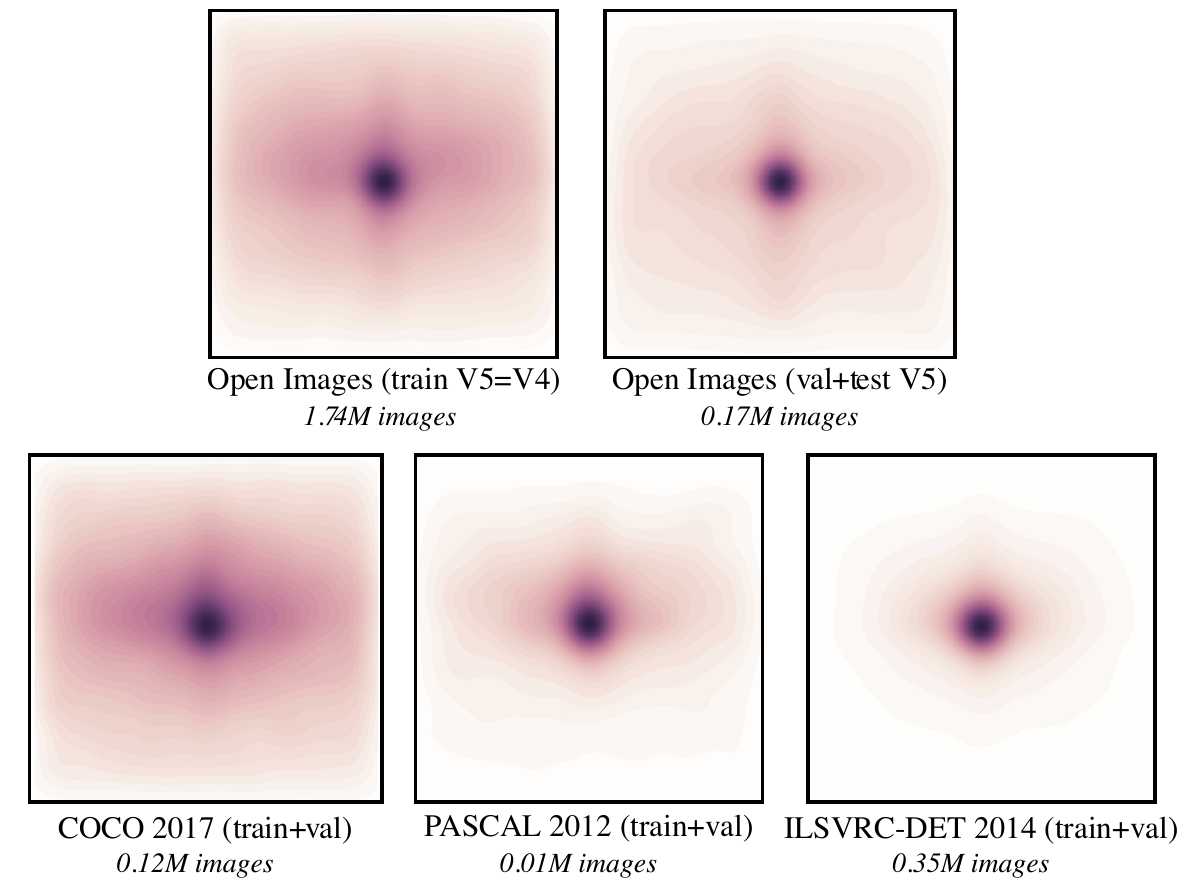}
    \caption{\textbf{Distribution of object centers} for various splits of \oi{} and other related datasets.}
    \label{fig:object_centers}
\end{figure}

\smallsection{Box center statistics}
As another way to measure the complexity and diversity of the boxes, Figure~\ref{fig:object_centers} shows the distributions of object
centers\footnote{We thank Ross Girshick for suggesting this type of visualization.}
in normalized image coordinates for \oi{} and other related datasets.
The \oi{} train set, which contains most of the data, shows a rich and diverse distribution of a complexity in a similar ballpark to that of \coco{}.
Instead, \pascal{} and \ilsvrc{} exhibit a simpler, more centered distribution.
This confirms what we observed when considering the number of objects per image (Fig.~\ref{fig:num_objs_hist}) and their area distribution (Fig.~\ref{fig:gt_areas}).

\smallsection{Validation and test V5}
The number of boxes per image (Tab.~\ref{tab:subsets_bboxes}) in \oi{} V4 is significantly higher in the train split than in validation and test.
In the next version of \oi{} (V5) we increased the density of boxes for validation and test to be closer to that of train (Tab.~\ref{tab:v4_vs_v5}).

\begin{table}[h]
    \centering
    \resizebox{\linewidth}{!}{%
    \begin{tabular}{l@{\hspace{20mm}}rrr}
        \toprule
        & Train & Validation & Test \\
        \midrule
        Boxes V4                     &	\num{14610229}	&   \num{204621}	&   \num{625282}	   \\
        \rule{3mm}{0mm}\it per image & \textit{8.4} & \textit{4.9} & \textit{5.0} \\
        Boxes V5                     &	\num{14610229}	&   \num{303980}	&   \num{937327}	   \\
        \rule{3mm}{0mm}\it per image & \textit{8.4} & \textit{7.3} & \textit{7.5} \\
        \bottomrule
    \end{tabular}}
    \caption{\textbf{Number of boxes for \oid{} V4 versus V5}.}
    \label{tab:v4_vs_v5}
\end{table}

Fig.~\ref{fig:object_centers} shows the distributions of object
centers. While the smaller val and test splits are still simpler than train, they are considerably richer than \ilsvrc{} and also slightly better than \pascal{}.

\subsection{Visual Relationships}
\label{sec:stats:vis_res}

\begin{figure*}[h]
\setlength{\fboxsep}{0pt}
\resizebox{\linewidth}{!}{%
\fbox{\includegraphics[height=2cm]{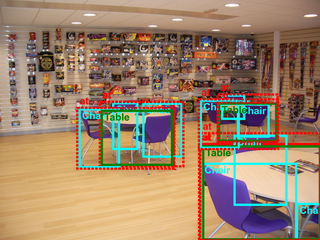}}%
\hspace{0.1pt}
\fbox{\includegraphics[height=2cm]{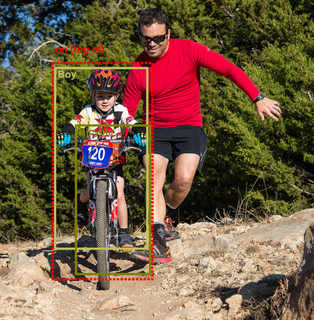}}%
\hspace{0.1pt}
\fbox{\includegraphics[height=2cm]{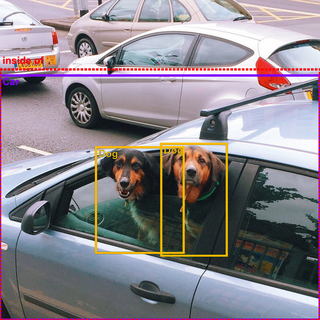}}}%
\\[1mm]
\resizebox{\linewidth}{!}{%
\fbox{\includegraphics[height=2cm]{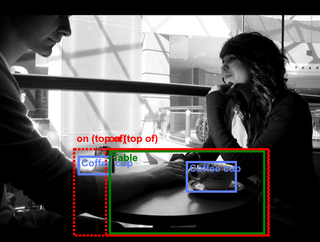}}%
\hspace{0.1pt}
\fbox{\includegraphics[height=2cm]{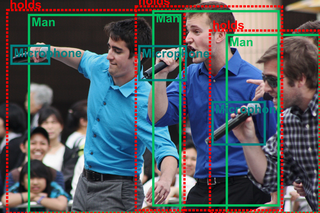}}%
\hspace{0.1pt}
\fbox{\includegraphics[height=2cm]{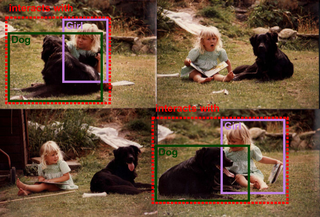}}%
}
\caption{\textbf{Examples of positively verified relationships}. Note how we annotated all instances of a relationship triplet (\eg{}multiple \lara{\cls{man}, holds, \cls{microphone}} in the same image).
}\label{fig:relationships}
\end{figure*}

The \oid{} was annotated in a very controlled manner. First, we produced image-level labels verified by human annotators. Afterwards, we annotated bounding boxes on {\em all instances} of each positive image-level label for 600 classes.
Now we expanded the annotations beyond object bounding boxes: we precisely defined a set of relationships between objects and then verified their presence for each pair of annotated objects in the dataset (Sec.~\ref{sec:vis_rel}).
In the end of the process we obtained \num{375}k annotations of \num{329} distinct relationship triplets, involving \num{57} different object classes.
Figure~\ref{fig:relationships} shows example annotations and Table~\ref{tab:vrd_oid_perset_stats} contains datasets statistics per split (train, validation, test).

\begin{table}[h] 
\centering
\resizebox{\linewidth}{!}{%
\begin{tabular}{lccc}
\toprule
& Train & Val & Test \\
\midrule
Number of VRD annotations \rule{10mm}{0mm} & $374,768$ & $3,991$ & $12,314$ \\
\bottomrule
\end{tabular}}
\caption[Caption]{Number of annotated visual relationship instances for the train, validation and test splits of \oi{}.\footnotemark}
\label{tab:vrd_oid_perset_stats}
\end{table}

On the other end of the annotation spectrum is data collection as proposed by the creators of Visual Genome (VG) and Visual Relationship Detection (VRD) datasets~\cite{krishna17ijcv, lu2016eccv}.\footnotetext{Visual relationships of validation and test sets are annotated using the boxes of V5 release since increased density of V5 allowed much denser relationships annotations.}
Their focus was on obtaining as much variety of relationships as possible by asking annotators to give a free-form region description, and annotate objects and relationships based on those descriptions.
The annotations from several annotators were then merged and combined using various language and quality models.

The difference in the two approaches naturally leads to difference in the properties of the two datasets:
while VG and VRD contain higher variety of relationship prepositions and object classes
(Tab.~\ref{tab:vis_rel_comparison}) they also have some shortcomings.
First, previous work shows that many of those are rather obvious, i.e. \lara{\cls{window}, on \cls{building}}~\cite{zellers18cvpr}.
Table~\ref{tab:most_freq_rel} compares the top-$10$ most frequent relationship triplets in all three datasets:
in both VG and VRD the most frequent relationships can be predicted from object co-occurrence and spatial proximity,  while \oi{} is more challenging in this respect.
Second, as follows from the free-form annotation process and lack of precise predefinitions, the annotations on VG and VRD contain multiple relationships with the same semantic meaning:
for example, the difference between relationships `near' and `next to' is not clear. This leads to annotation noise as multiple instances of conceptually the same relationship have different labels. Since \oi{}  annotations were collected in a very controlled setting this kind of noise is much lower.
Finally, in VG and VRD annotations within an image are sometimes incomplete (\eg{}if there are two chairs at a table in the same image, only one of them might be annotated).
Instead, thanks to the controlled annotation process for image-level labels, boxes, and relationships, in \oi{} for each image it is possible to know exactly if two objects are connected by a certain relationship or not.
This makes \oi{} better suited for evaluating the performance of visual relationship detection models, and also facilitates negative samples mining during training.

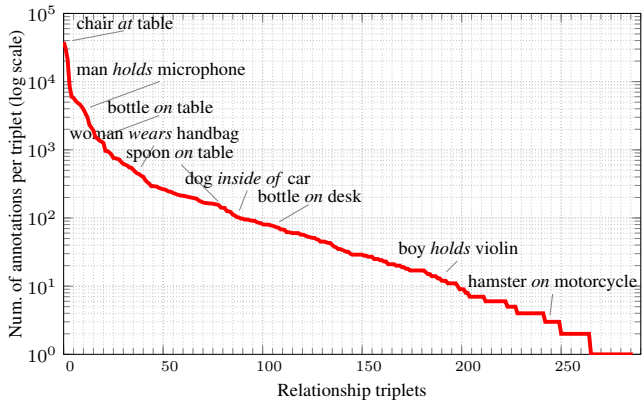
\begin{figure}
\pgfplotsset{every x tick label/.append style={font=\scriptsize, xshift=0.8ex}}
  \resizebox{\linewidth}{!}{%
\begin{tikzpicture}
\begin{axis}[width=1.3\linewidth,height=0.85\linewidth,
              ylabel=Num. of annotations per triplet (log scale),ylabel shift={-5pt},
              xlabel=Relationship triplets,
              ytick={1,1e1,1e2,1e3,1e4, 1e5},
              minor xtick={0,5,...,290},
              ymode=log,
              xmin=0,
              xmax=290,
              ymin=1,
              ymax=100000,
              enlargelimits=false,grid=both,grid style=densely dotted]
  \putpin{(1,38326)}{8.5mm}{10}{chair \textit{at} table}
  \putpin{(10,3889)}{12mm}{20}{man \textit{holds} microphone}
  \putpin{(17,1464)}{9.5mm}{20}{bottle \textit{on} table}

  \putpin{(37,451)}{4mm}{60}{woman \textit{wears} handbag}
  \putpin{(81,140)}{8.5mm}{135}{spoon \textit{on} table}
  \putpin{(87,103)}{3.5mm}{70}{dog \textit{inside of} car}
  \putpin{(105,77)}{5.5mm}{30}{bottle \textit{on} desk}
  \putpin{(189,13)}{2.5mm}{45}{boy \textit{holds} violin}
   \putpin{(244,3)}{3.5mm}{85}{hamster \textit{on} motorcycle}

  \addplot+[red,solid,mark=none,line width=1.8pt] table[x=index,y=freq,col sep=tab] {data/vrd/triplets_count_oid.tsv};
\end{axis}
\end{tikzpicture}}
  \caption{Number of annotations per triplet on \oi{} (two-object relationships only, without attributes).}\label{fig:stats_class_per_triplet}
\end{figure}

As one can expect from object class distribution on \oi{}, the distribution of the number of relationship annotations among triplets is highly imbalanced (Fig.~\ref{fig:stats_class_per_triplet}).  Hence, the \oi{} dataset includes both  rare and very frequent relationship triplets. This suggests that to be able to effectively detect triplets that have very small number of annotations, it will not be enough to train a monolithic detector for each triplet. We expect that a successful detector will have to be compositional.

Figures~\ref{fig:stats_class_per_triplet_aligned_vg} and~\ref{fig:stats_class_per_triplet_aligned_vrd} provide a comparison between the number of annotations in \oi{} vs in VG/VRD for the semantic triplets they have in common (considering only two-object relationships, not attributes). As the plots shows, the \oid{} has more annotations for several triplets than VG/VRD, which shows it can complement them\footnote{To find the triplets in common between two datasets we matched the class names based on Lexicographical %
comparison and aggregated annotations in VG based on relationship; since VG contains somewhat inconsistent relationship names, we use loose string matching to match relationships}.
Moreover, \oi{} contains new relationship triplets than are not in VG at all, \eg{}\lara{\cls{man}, play, \cls{flute}},
\lara{\cls{dog}, inside of, \cls{car}}, \lara{\cls{woman}, holds, \cls{rugby ball}}.

\begin{table*}[h]
\centering
\begin{tabular}{l|rrrr}
\toprule
   & Num. classes (/ num. attributes) & Num. distinct triplets & Num. annotations \\
\midrule
Visual Relationship Detection \cite{lu2016eccv} & \num{100} & \num{6672} & \num{30355} \\
\midrule
Visual Genome \cite{krishna17ijcv}  & \num{67123} / \num{4279} &\num{727635} & \num{2578118} \\
\qquad\textit{two-object relationships}  & \num{65398} & \num{675378} & \num{2316104} \\
\qquad\textit{attributes}  & \num{7100} / \num{4279} &\num{52257} & \num{262014} \\
\midrule
Open Images   & \num{57} / \num{5} & \num{329} & \num{374768} \\
\qquad\textit{two-object relationships}  & \num{57} & \num{287} & \num{180626} \\
\qquad\textit{attributes}  & \num{23} / \num{5} & \num{42} & \num{194142} \\
\bottomrule
\end{tabular}
\caption{Comparison with the existing visual relationship detection datasets.
}\label{tab:vis_rel_comparison}
\end{table*}

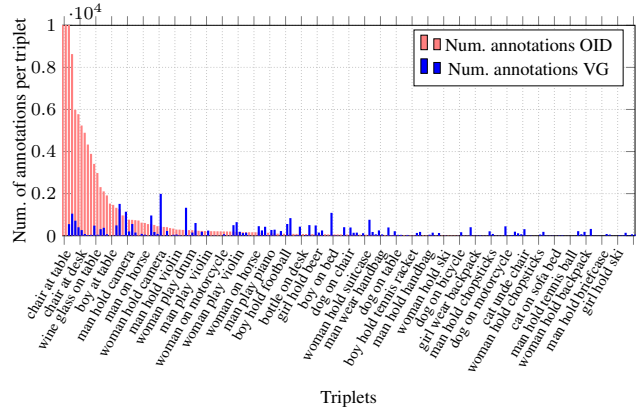
\begin{figure}[h]
\pgfplotsset{every x tick label/.append style={font=\scriptsize, xshift=0.8ex}}
  \resizebox{\linewidth}{!}{%
\begin{tikzpicture}
 \begin{axis}[ybar,width=1.3\linewidth,height=0.6\linewidth,
              ylabel=Num. of annotations per triplet,ylabel shift={-5pt},
              xlabel=Triplets,
              xtick={-0.5,4.5,...,180},
              xticklabels={chair at table,chair at desk,wine glass on table,boy at table,man hold camera,man on horse,woman hold camera,man hold violin,woman play drum,man play violin,woman on motorcycle,woman play violin,
              woman on horse,man play piano,boy hold football,bottle on desk,girl hold beer,boy on bed,dog on chair,woman hold suitcase,man wear handbag,dog on table,boy hold tennis racket,
              man hold handbag,woman hold ski,dog on bicycle,girl wear backpack,man hold chopsticks,dog on motorcycle,cat unde chair,woman hold chopsticks,cat on sofa bed,
              man hold tennis ball,woman hold backpack,man hold briefcase,girl hold ski},
              x tick label style={rotate=60,anchor=east},
              xmin=-1,
              xmax=180,
              ymin=0,
              ymax=10000,
              enlargelimits=false,grid=both,grid style=densely dotted]
  \addplot [ybar, bar width=1pt,draw=none, fill=red!50] table[x=index,y=oid_freq,col sep=tab] {data/vrd/oid_vg_comparison.tsv};
  \addplot [ybar, bar width=1pt,draw=none, fill=blue!100] table[x=index,y=vg_freq,col sep=tab] {data/vrd/oid_vg_comparison.tsv};
  \legend{Num. annotations OID, Num. annotations VG}
\end{axis}
\end{tikzpicture}}
  \caption{\textbf{Comparison of the number of triplet annotations} on \oi{} dataset vs. Visual Genome dataset for $195$ triplets found in common (two-object relationships only, without attributes).}\label{fig:stats_class_per_triplet_aligned_vg}
\end{figure}
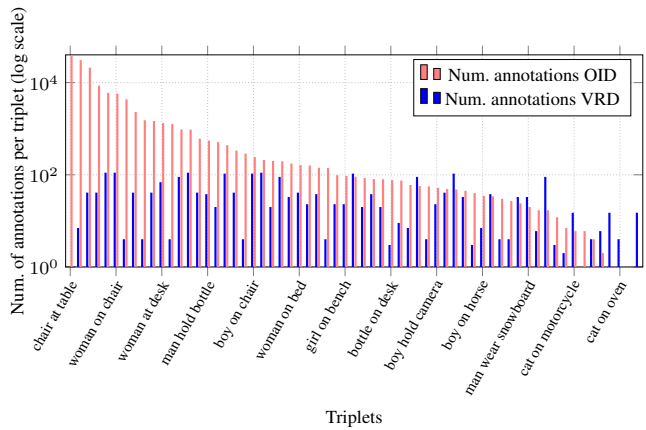
\begin{figure}[h]
\pgfplotsset{every x tick label/.append style={font=\scriptsize, xshift=0.8ex}}
  \resizebox{\linewidth}{!}{%
\begin{tikzpicture}
 \begin{axis}[ybar,width=1.3\linewidth,height=0.6\linewidth,
              ylabel=Num. of annotations per triplet (log scale),ylabel shift={-5pt},
              xlabel=Triplets,
              ymode=log,
              xtick={-0.5,4.5,...,180},
              xticklabels={chair at table,woman on chair,woman at desk,man hold bottle,boy on chair,woman on bed,girl on bench,bottle on desk,boy hold camera,boy on horse,man wear snowboard,
              cat on motorcycle,cat on oven},
              x tick label style={rotate=60,anchor=east},
              xmin=-1,
              xmax=62,
              ymin=0,
              ymax=40000,
              enlargelimits=false,grid=both,grid style=densely dotted]
  \addplot [ybar, bar width=1pt,draw=none, fill=red!50] table[x=index,y=oid_freq,col sep=tab] {data/vrd/oid_vrd_comparison.tsv};
  \addplot [ybar, bar width=1pt,draw=none, fill=blue!100] table[x=index,y=vrd_freq,col sep=tab] {data/vrd/oid_vrd_comparison.tsv};
  \legend{Num. annotations OID, Num. annotations VRD}
\end{axis}
\end{tikzpicture}}
  \caption{\textbf{Comparison of the number of triplet annotations} on the \oid{} versus the VRD dataset for $62$ triplets found in common (two-object relationships only, without attributes).}\label{fig:stats_class_per_triplet_aligned_vrd}
\end{figure}

In summary, \oi{} visual relationship annotations are not as diverse as in VG and VRD, but are better defined, avoid obvious relationships, have less annotation noise, and are more completely annotated. Moreover, \oi{} offers some complementary annotations to VG/VRD, both by the number of samples for some the triplets they have in common, and by some entirely new triplets.

\begin{table}[h]
\resizebox{\linewidth}{!}{%
\begin{tabular}{ccc}
\toprule
VRD dataset  & Visual Genome & Open Images \\
\midrule
person \textit{wear} shirt & window \textit{on} building & chair \textit{at} table \\
person \textit{wear} pants & clouds \textit{in} sky & man \textit{at} table \\
person \textit{next} to person & man \textit{wearing} shirt & woman \textit{at} table \\
person \textit{wear} jacket & cloud \textit{in} sky & man \textit{on} chair  \\
person \textit{wear} hat & sign \textit{on} pole & woman \textit{on} chair \\
person \textit{wear} glasses & man \textit{wearing} hat & chair \textit{at} desk \\
person \textit{has} shirt & leaves \textit{on} tree & man \textit{holds} guitar \\
person \textit{behind} person & man \textit{wearing} pants & man \textit{plays} guitar \\
person \textit{wear} shoes & man \textit{has} hair & chair \textit{at} coffee table \\
shirt \textit{on} person & building \textit{has} window & girl \textit{at} table \\
\bottomrule
\end{tabular}}
\caption{Top-$10$ most frequent relationships in the VRD, Visual Genome and Open Images datasets. \new{Note that some of the Open Images relations are not mutually exclusive (\eg{}``man holds guitar'' and ``man plays guitar'').
In these cases, we have annotated all relationships that occur in each particular sample (see Section~\ref{ss:annotating-rels}).}}\label{tab:most_freq_rel}
\end{table}

\section{Quality}
\label{sec:quality}

\subsection{Quality of bounding boxes}
\label{sec:box_quality}

We performed an extensive analysis of the quality of bounding box annotations.
We had a human expert examine 100 images for each of the first 150 boxable classes sorted by alphabetical order, containing a total of more than \num{26000} boxes.
We measured the quality of both the boxes and their attributes.

\begin{figure}[htpb]
\pgfplotsset{every x tick label/.append style={font=\scriptsize, xshift=0.8ex}}
  \resizebox{\linewidth}{!}{%
\begin{tikzpicture}
\begin{axis}[y filter/.code={\pgfmathparse{#1*100}\pgfmathresult},
			  width=1.3\linewidth,height=0.85\linewidth,
              ylabel=Semantic error percentage (\%),ylabel shift={-5pt},
              ytick={0,10,...,100},
              xtick={0,5,...,150},
              xticklabels={Bidet,Cake,Carnivore,Cocktail shaker,Baked goods,Briefcase,Cantaloupe,Artichoke,Barge,Coffee,Belt,Bus,Accordion,Corded phone,Bookcase,Bee,Calculator,Chest of drawers,Bread,Chicken,Ball,Coin,Candle,Candle,Caterpillar,Caterpillar,Bed,Barrel,Armadillo,Car,Billiard table},
              x tick label style={rotate=60,anchor=east},
              minor ytick={0,2.5,...,100},              
              minor grid style={white!85!black},
              major grid style={white!60!black},
              xmin=0,
              xmax=150,
              ymin=0,
              ymax=100,
              enlargelimits=false,grid=both,grid style=densely dotted]
  \putpin{(1,0.863)}{4mm}{20}{Bidet}
  \putpin{(2,0.552)}{8mm}{40}{Cello}
  \putpin{(3,0.347)}{12mm}{45}{Coffee table}
  \putpin{(4,0.276)}{10mm}{30}{Cucumber}
  \putpin{(5,0.251)}{22mm}{24}{Cocktail}
  \putpin{(6,0.239)}{22mm}{14}{Cake}
  \putpin{(7,0.216)}{18mm}{6}{Cheetah}
  \putpin{(8,0.178)}{20mm}{2}{Crutch}
  \putpin{(9,0.169)}{24mm}{-5}{Canary}
  
  \addplot+[blue,solid,mark=none,line width=1.8pt] table[x=Index,y=Error,col sep=tab] {data/class_errors.tsv};
\end{axis}
\end{tikzpicture}}
  \caption{\textbf{Percentage of boxes which have a semantic error} for each of the 150 examined classes. Every fifth class name is displayed on the horizontal axis. Moreover, we provide the names of the 9 classes with the highest percentage of errors directly on the curve.
}
  \label{fig:semantic_errors}
\end{figure}
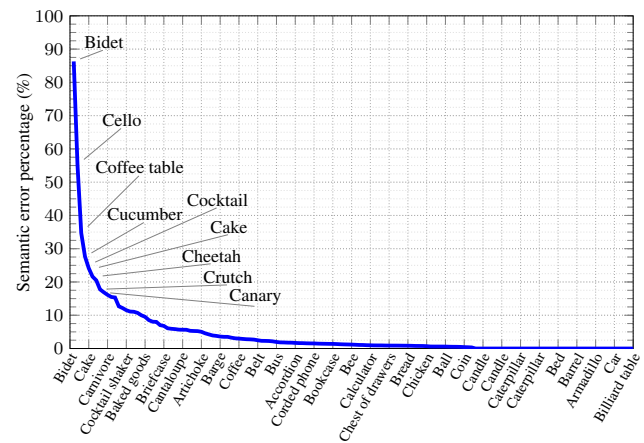

Results for box quality are shown in Table~\ref{tab:box_precision_recall}.
Both precision and recall are very high at 97.7\% and 98.2\%, respectively.
The missing precision is mostly caused by boxes which are geometrically imprecise (1.1\%),
and boxes with wrong semantic class labels (1.1\%). Imprecise boxes are quite evenly spread over classes.
However, while half of the classes have fewer than 1\% semantic errors, other (often rare) classes have more (Fig.~\ref{fig:semantic_errors}). Most notably, the three most problematic classes are \cls{bidet} (86\% errors, confused with toilets), \cls{cello} (55\% errors, confused with violins), and \cls{coffee table} (35\% errors, confused with other kinds of tables). The first two mistakes are caused by cultural differences. \cls{coffee table} is an ambiguous class.

\begin{table}[htpb]
  \centering
\begin{tabular}{ccc|c}
\toprule
\multicolumn{3}{c|}{precision}                    & recall     \\
\multicolumn{3}{c|}{97.7\%}                       & 98.2\%     \\
\midrule
imprecise & wrong class & multiple objects & \cellcolor{gray!15} \\
1.1\%           & 1.1\%       & 0.1\% & \cellcolor{gray!15}\\
\bottomrule
\end{tabular}
\caption{\textbf{Analysis of box quality}. Conditioned on a given class label for an image, we report precision and recall. We break down precision errors into three different types: an geometrically imprecise box, a box with the wrong class label, and a box which unjustifiably captures multiple objects.
}
\label{tab:box_precision_recall}
\end{table}

Table~\ref{tab:attribute_errors} shows results for attribute quality. Precision and recall are in the high nineties for most attributes. Especially the \cls{Occluded} and \cls{Truncated} attributes are very accurately annotated. For \cls{Depiction}, recall is 92\%.
For \cls{Inside}, precision is 67\%, which is mostly caused by several \cls{Bell peppers} incorrectly having this attribute when it was inside a container such as a shopping cart.
However, the \cls{Inside} attribute is extremely rare (0.4\% of all boxes) and only relevant for a few classes. 
The most frequent such class is \cls{Building}, for which precision and recall are good (100\% and 83\%, respectively).

\begin{table}[hptb]
\resizebox{\linewidth}{!}{%
\begin{tabular}{l@{\hspace{40mm}}ll}
\toprule
Attribute & Precision & Recall \\
\midrule
\cls{GroupOf}   & 94.2\%    & 95.3\% \\
\cls{Occluded}  & 98.6\%    & 98.4\% \\
\cls{Truncated} & 99.7\%    & 97.0\% \\
\cls{Depiction} & 96.7\%    & 92.2\% \\
\cls{Inside}    & 66.7\%    & 90.3\% \\
\bottomrule
\end{tabular}}
\caption{\textbf{Precision and recall for attributes}, with the error rates specified per attribute type.}
\label{tab:attribute_errors}
\end{table}

\subsection{Geometric agreement of bounding boxes}
\label{sec:box_ov_agreement}

Another way to measure the quality of bounding boxes is to draw them twice by different annotators, and then measuring their geometric agreement. We did that for both Extreme Clicking (Sec.~\ref{sec:extreme_clicking}) and Box Verification Series (Sec.~\ref{sec:box_verification}).

\paragraph{Extreme clicking.}

We randomly selected 50,000 boxes produced with extreme clicking, and had annotators redraw the box (without seeing the original box). We then measured human agreement as the average intersection-over-union (IoU) between the original box and the redrawn box on the same object.
We found this to be 0.87, which is very close to the human agreement of 0.88 IoU on \pascal~\cite{everingham15ijcv} reported by~\cite{papadopoulos17iccv}. The slight difference is mainly caused by objects being generally smaller in \oi{} (Fig.~\ref{fig:gt_areas}).

\paragraph{Box Verification Series.}

\begin{figure}[htpb]
  \centering
  \includegraphics[width=0.596\linewidth]{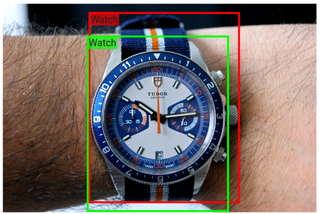}
  \includegraphics[width=0.394\linewidth]{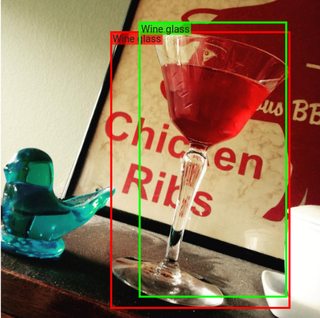}
  \caption{\textbf{Two examples of matching boxes with $IoU = 0.77$}, the average agreement between verified and drawn boxes. Verified boxes are in green, drawn boxes are in red. As these example shows, the verified boxes cover the object very well, but not perfectly.}
  \label{fig:examples_iou_77}
\end{figure}

For the boxes produced with box verification series, we had 1\% re-annotated using extreme clicking (again without showing the original box).
We then measured IoU between the original boxes and the newly manually drawn boxes. We found this to be 0.77 on average. As expected, this is higher than the threshold of IoU$>0.7$ for which we trained the annotators, and lower than the extreme clicking agreement of 0.87.
We underline that $0.77$ is widely considered as a good geometric accuracy (\eg{}the \coco{} Challenge calls IoU $>0.75$ a ``strict'' evaluation criterion).
To give a better feeling of the average quality of these boxes, Fig.~\ref{fig:examples_iou_77} shows two examples where a drawn box and box produced by box verification series have IoU$=0.77$.

\paragraph{Detectors: extreme clicking vs box verification series.}

The extreme clicking boxes are more accurate than those produced by box verification series.
But how does this influence contemporary object detection models?
To answer this question, we make use of the 1\% of re-annotated data from box verification series. This means that for the exact same object instances, we have both verified boxes and manually drawn boxes.
We train a Faster-RCNN~\cite{ren15nips} model based on Inception-ResNet-V2~\cite{szegedy17aaai} on each set and measure performance on the \oi{} test split.
We measure performance using the \oi{} Challenge metric $mAP_{OI}$ (Sec.~\ref{subsec:od_eval}),
which is a modified version of mean Average Precision commonly used for object detection~\cite{everingham10ijcv}.
Interestingly, we found that the difference in detection performance was smaller than 0.001 $mAP_{OI}$.
We conclude that boxes produced by box verification series make perfectly useful training data for contemporary detectors, as they lead to the same performance as training from manually drawn boxes.

\subsection{Recall of boxable image-level labels}
\label{sec:recal_im_level}

As described in Sec.~\ref{sec:image_labels}, we obtained image-level labels by verifying candidate labels produced automatically by a classifier. Here we estimate the general recall of this process for the 600 boxable classes. We randomly sampled 100 images and inspected each image independently by two human experts to identify all instances of boxable classes which were not annotated. This was done by displaying each image with all existing box annotations overlaid. For each non-boxed object, the expert typed multiple free-form words, each of which was mapped to the closest five \oi{} boxable classes through Word2Vec~\cite{mikolov13nips}.
Based on these, the expert then decided whether the object indeed belonged to a boxable class, and recorded it as a missing object.
Additionally, the object could be marked as `difficult' according to the \pascal{} standards~\cite{everingham10ijcv}, i.e. very small, severely occluded or severely truncated.
Afterwards, we took the set union of all labels of all missing objects recorded by the two experts.
We removed existing image-level labels from this set, to cover for the rare case when an object instance was not boxed even though its image-level label was available (high recall in Tab. \ref{tab:box_precision_recall}).
This results in the final set of classes present in the image but for which we do not have an image-level label.

When considering really all objects, the recall of image-level labels is 43\% in \oi{}. When disregarding `difficult' objects, the recall is 59\%.
While this is lower than the estimated recall of 83\% reported for \coco{}~\cite{lin14eccv}, \oi{} contains $7.5\times$ more classes, making it much harder to annotate completely. Importantly, this lack of complete annotation is partially compensated by having explicit negative image labels. These enable proper training of discriminative models.
Finally, we stress that for each positive image-level label we annotated bounding boxes for {\em all} instances of that class in the image. Along that dimension, the dataset is fully annotated (98.2\% recall, Tab.~\ref{tab:box_precision_recall}).

\section{Performance of baseline models}
\label{sec:baselines}

In this section we provide experiments to quantify the performance of baseline models for image classification (Sec.~\ref{subsec:image_classification}) and object detection (Sec.~\ref{sec:object_detection}) on the \oid{}.

\subsection{Image Classification}
\label{subsec:image_classification}

\new{Image classification has fostered some of the most relevant advances in computer vision in the
last decade, bringing new techniques whose impact has reached well beyond the task
itself~\cite{krizhevsky12nips,szegedy15cvpr,ioffe15icml,he16cvpr,szegedy17aaai}.
Here we train an image classification model with Inception-ResnetV2~\cite{szegedy17aaai},
a widely used high capacity network, and we empirically measure:}
\begin{itemize}
\item The impact of the number of human-verified labels on the quality of a classifier.
\item The impact of using negative human-verified labels.
\item Classification performance when restricted to boxable classes only.
\end{itemize}

\smallsection{Training}
For our experiments we use the model described in Section~\ref{sec:candidate_labels_valtest} to produce machine-generated labels on the train split. We consider each label predicted with confidence score above \num{0.5} as positive. This way we have two sets of labels for the train split: these machine-generated labels, and the human-verified labels as discussed in Section~\ref{sec:human_verification}.

In these experiments we consider only classes with at least 100 positive human-verified examples in the train split (\num{7186} classes).

We first pre-train the network from scratch using the machine-generated labels. We then fine-tune it with a mix of 90\% human-verified labels and 10\% machine-generated labels.

\smallsection{Evaluation}
For each class we calculate Average Precision on the test set, for the \num{4728} classes that are both in the trainable label set and have $>0$ samples in the test set.
During evaluation we take into account that the ground-truth annotations are not exhaustive (Sec.~\ref{sec:image_labels}), and do not penalize the model for predicting classes for which we do not have human verification on the test set. This metric is discussed in detail for Open Images in~\cite{Veitcvpr17}.

\smallsection{Number of human-verified labels}
To measure the impact of the human-verified labels we repeat the fine-tuning stage described above to build classification models using a varying fraction of the human-verified labels. In all case we start from the model pre-trained on machine-generated labels on the entire train split.
We then fine-tune on human-verified labels from random subsets of the train split containing \num{1}\%, \num{10}\%, \num{25}\%, \num{50}\%, \num{75}\%, and \num{100}\% of all images.

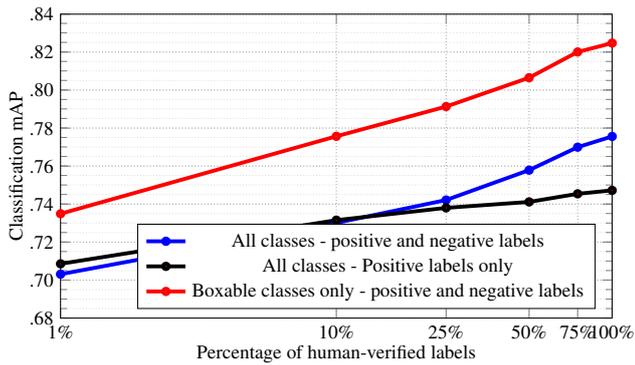
\begin{figure}[h]
\resizebox{\linewidth}{!}{%
\begin{tikzpicture}
\begin{axis}[width=1.2\linewidth,height=0.75\linewidth,
              ylabel=Classification mAP,ylabel shift={-3pt},xlabel=Percentage of human-verified labels,xlabel shift={-3pt},
              ytick={0.68,0.7,0.72,...,0.86},
              yticklabels={.68, .70,.72,.74,.76,.78,.80,.82,.84, .86},
	      xtick={.01, .1, .25, .50, .75, 1},
	      xticklabels={ 1\%, 10\%, 25\%, 50\%, 75\%, 100\%},
              xmin=0,
              xmax=1,
              ymin=0.68,
              ymax=0.84,
	      xmode=log,
	      minor ytick={0.68,0.685,...,0.84},
              legend pos = south east,
              minor grid style={white!85!black},
              major grid style={white!60!black},
              enlargelimits=false,grid=both,grid style=densely dotted]
  \addplot+[blue,solid,mark=*, mark size=1.2, line width=1.7pt] table[x=sample_ratio,y=map_all] {data/image_level/classifier_map.tsv};
  \addlegendentry{All classes - positive and negative labels}
  \addplot+[black,solid,mark=*, mark size=1.2, line width=1.7pt] table[x=sample_ratio,y=map_no_negatives] {data/image_level/classifier_map.tsv};
  \addlegendentry{All classes - Positive labels only}
  \addplot+[red,solid,mark=*, mark size=1.2, line width=1.7pt] table[x=sample_ratio,y=map_boxable] {data/image_level/classifier_map.tsv};
  \addlegendentry{Boxable classes only - positive and negative labels} %
\end{axis}
\end{tikzpicture}}
\caption{\textbf{Classifier performance versus amount of human-verified labels} in terms of the percentage of all available labels.}
\label{fig:classification_map}
\end{figure}

As shown in Fig.~\ref{fig:classification_map}, mAP increases as we increase the amount of human-verified labels, demonstrating that they directly improve model performance.
The absolute number of human-verified labels can be calculated using the values in Table~\ref{tab:image_level_rated_stats}.

\smallsection{Impact of negative labels}
To measure the impact of negative image-level labels we repeat the above experiment but train from positive labels only (Fig.~\ref{fig:classification_map}).
\new{For this we train our classification model while ignoring human-verified negative labels in the loss. Instead, we use as implicit negatives all labels that are not human-verified as positive (including candidate labels generated by the image classifier, and all other labels that are missed by it, Sec.~\ref{sec:image_labels}).}
We observe that, as saturation starts to occur when using a large number of positive labels, negative labels start to improve the model significantly.
\new{Hence, this experiment shows the value of explicit, human-verified negative labels.}

\smallsection{Boxable classes}
We also report mAP for the 600 boxable classes (using the same models as before).
As shown in Fig.~\ref{fig:classification_map}, mAP for these boxable classes is generally higher than mAP for all classes.
Boxable classes are generally easier for classification tasks, as they are all concrete objects defined clearly by their visual properties (as opposed to some of the classes in the wider \num{19794} set, \eg{}love and birthday). Also, they are mostly basic-level categories, whereas the wider set contains many fine-grained classes which are harder to distinguish (\eg{} breeds of dogs).
As before, we observe that a larger number of human-verified labels translates to a higher value of mAP.

\subsection{Object detection}
\label{sec:object_detection}
\new{
The advent of object detection came in the form classifiers applied densely to windows sliding
over the image (\eg{}based on boosting~\cite{viola01cvpr, viola01ijcv} or
Deformable Part Models~\cite{felzenszwalb10cvpr_b, felzenszwalb10pami}).
To reduce the search space, the concept of ``object proposals''~\cite{alexe10cvpr,alexe12pami,uijlings13ijcv} was then introduced, which enable to work on just a few thousand windows instead of a dense grid. 

R-CNN~\cite{girshick14cvpr} brought the advances in image classification using deep learning to
object detection using a two-stage approach: classify object proposal boxes~\cite{uijlings13ijcv}
into any of the classes of interest.
This approach evolved into Fast R-CNN~\cite{girshick15iccv} and Faster R-CNN~\cite{ren15nips}, which generates the proposals with a deep network too. Faster R-CNN stills provide very competitive results today in terms of accuracy.

More recently, single-shot detectors were presented to bypass the computational bottleneck of object proposals by regressing object locations directly from a predefined set of anchor boxes (\eg{}SSD~\cite{liu16eccv} and YOLO~\cite{redmon16cvpr,redmon17cvpr}). This typically results in simpler models that are easier to train end-to-end.

In this section we evaluate the performance of two modern object detectors on \oi{}
(the two-stage Faster-RCNN~\cite{ren15nips} and the single-shot SSD~\cite{liu16eccv}).
} %
We start by defining an evaluation metric that takes into account the characteristics of \oi{} in
Sec.~\ref{subsec:od_eval}.
Then we detail our evaluation setup and report results exploring various model architectures and
training set sizes in Section~\ref{subsubsec:capacity}.

\subsubsection{Evaluation metric}
\label{subsec:od_eval}

The standard metric used for object detection evaluation is \pascal{} VOC 2012 mean average precision (mAP)~\cite{pascal-voc-2012}.
However, this metric does not take into account several important aspects of the \oid{}:
non-exhaustive image-level labeling, presence of class hierarchy and group-of boxes.
We therefore propose several modifications to \pascal{} VOC 2012 mAP, which are discussed in detail below.

\mypar{Non-exhaustive image-level labeling.}
Each image is annotated with a
this is not the spirit of the point we are making, it is a boundary case; and anyway sets can be empty!
set of positive image-level labels,
indicating certain classes are present, and negative labels, indicating certain classes are absent.
All other classes are unannotated.
Further, for each positive image-level label, every instance of that object class is annotated with a ground-truth bounding box.
For fair evaluation we ignore all detections of unannotated classes.
A detection of a class with a negative label is counted as false positive.
A detection of a class with a positive label is evaluated as true positive or false positive depending on its overlap with the ground-truth bounding boxes (as in \pascal{} VOC 2012).
For a detection to be evaluated as true positive, its intersection-over-union with a ground-truth bounding box should be greater than $0.5$.

\mypar{Class hierarchy.}
\oi{} bounding box annotations are created according to a hierarchy of classes (Section~\ref{sec:hierarchical_dedup}).
For a leaf class in the hierarchy, AP is computed as normally in \pascal{} VOC 2012 (\eg{}`Football Helmet').
In order to be consistent with the meaning of a non-leaf class, its $AP$ is computed involving all its ground-truth object instances and all instances of its subclasses.
For example, the class \cls{Helmet} has two subclasses (\cls{Football Helmet} and \cls{Bicycle Helmet}).
These subclasses in fact also belong to \cls{Helmet}.
Hence, $AP_{\cls{Helmet}}$ is computed by considering that the total set of positive \cls{Helmet} instances are the union of all objects annotated as \cls{Helmet}, \cls{Football Helmet}, and \cls{Bicycle Helmet} in the ground-truth.
As a consequence, an object detection model should to produce a detection for each of the relevant classes, even if each detection corresponds to the same object instance.
For example, if there is an instance of \cls{Football Helmet} in an image, the model need to output detections for both \cls{Football Helmet} and for \cls{Helmet} in order to reach 100\% recall (see the semantic hierarchy visualization in Fig.\ref{fig:boxable_class_hierarchy}).
If only a detection with \cls{Football Helmet} is produced, one true positive is scored for \cls{Football Helmet} but the \cls{Helmet} instance will not be detected (false negative).

\mypar{Group-of boxes.}
A group-of box is a single box containing several object instances in a group (i.e. more than $5$ instances which are occluding each other and are physically touching).
The exact location of a single object inside the group is unknown.

We explore two ways to handle group-of boxes. These can be explained in a unified manner, differing in the value of a parameter weight $w \in \{0,1\}$.
If at least one detection is inside a group-of box, then a single true positive with weight $w$ is scored. Otherwise, the group-of box is counted as a single false negative with the same weight $w$.
A detection is inside a group-of box if the area of intersection of the detection and the box divided by the area of the detection is greater than $0.5$.
Multiple correct detections inside the same group-of box still count as a single true positive.

When $w=0$ group-of boxes act like ignore regions: the detector is not required to detect them, and if it does output detections inside them, they are ignored.
Instead, when $w=1$ the detector is required to output at least one detection inside a group-of box.
In our final evaluation metric, we use $w=1$.

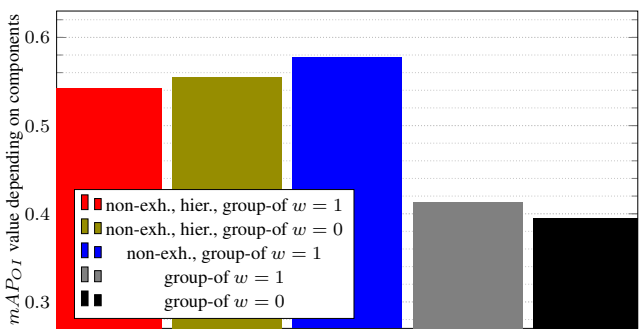
\begin{figure}[h]
\resizebox{\linewidth}{!}{%
\begin{tikzpicture}
 \begin{axis}[ybar,width=1.3\linewidth,height=0.8\linewidth,
              ylabel=$mAP_{OI}$ value depending on components,ylabel shift={-5pt},
              xticklabels = {},
              grid=both,
              xtick style={draw=none},
              xmajorgrids=false,
              minor ytick={0,0.02,...,0.8},
              ybar=5pt,
              ymin=0.3,
              ymax=0.6,
              xmin=-2.6,
              xmax=4.6,
              bar width=50pt,
              legend pos = south west,
              enlargelimits=true,grid style=densely dotted]
  \addplot[ybar, draw=none, fill=red!100] coordinates {(1,0.543012994097)};
  \addlegendentry{non-exh., hier., group-of $w=1$}
  \addplot[ybar, draw=none, fill=olive!100] coordinates {(1,0.555570313039)};
  \addlegendentry{non-exh., hier., group-of $w=0$}
  \addplot[ybar, draw=none, fill=blue!100] coordinates {(1,0.577391873141)};
  \addlegendentry{non-exh., group-of $w=1$}
  \addplot[ybar, draw=none, fill=gray!100] coordinates {(1,0.412960070066)};
  \addlegendentry{group-of $w=1$}
  \addplot[ybar, draw=none, fill=black!100] coordinates {(1,0.395389193559)};
  \addlegendentry{group-of $w=0$}
\end{axis}
\end{tikzpicture}}
\caption{\textbf{Effect of the components of the \oi{} metric}.
The full metric (non-exh., hier., group-of $w=1$) and the effect of its components:
non-exhaustive labeling, presence of hierarchy, group-of box weight $w$.
The black bar represents a metric very close to the standard \pascal{} VOC 2012 mAP (with the only addition of ignoring detections inside a group-of box).
}
\label{fig:metric_comparison}
\end{figure}

\mypar{Effects.}
To evaluate the effect of proposed customized evaluation metric we show results on the Faster-RCNN detector~\cite{ren15nips} with Inception-ResNetV2 backbone~\cite{szegedy17aaai} using various versions of the metric.
The details of training are given in the following Section~\ref{subsubsec:capacity}.
As Figure~\ref{fig:metric_comparison} shows, the biggest difference is caused by ignoring detections of unannotated classes, and thus taking into account the non-exhaustiveness of the annotations. Without this, correct detections of objects from unannotated classes would be wrongly counted as false negatives.

\subsubsection{Performance evaluation}
\label{subsubsec:capacity}

We evaluate \AlinaReplaced{three}{two} modern object detection models with different capacities.
The first model is Faster-RCNN~\cite{ren15nips} with an Inception-ResNetV2 backbone, which performs feature extraction~\cite{szegedy17aaai}.
\AlinaReplaced{The second model is SSD~\cite{liu16eccv} with InceptionV2 feature extractor~\cite{szegedy16cvpr}.
We do not consider the combination of SSD and Inception-ResNetV2 since SSD is designed to speedup detection performance while Inception-ResNetV2 is slow at inference.}{}
The \AlinaReplaced{third}{second} model is SSD~\cite{liu16eccv} with MobileNetV2~\cite{sandler18cvpr} feature extractor with depth multiplier $1.0$ and input image size $300\times300$ pixels.
We report in Table~\ref{tab:detector_cap} the number of parameters and inference speed for each detection model.

\begin{table}[h]
\centering
\resizebox{\linewidth}{!}{%
\begin{tabular}{lcc}
\toprule
 Detector      & Number of parameters & Inference time (s) \\
\midrule
Faster-RCNN with Inception-ResNetV2 & $63.947.366$ & $0.455$  \\
\AlinaReplaced{SSD with InceptionV2 & $93.171.799$ & $0.041$  \\}{}
SSD with MobileNetV2, dm=1.0 & $14.439.167$  & $0.024$  \\
\bottomrule
\end{tabular}}
\caption{\textbf{Detector capacity}: number of parameters and inference speed measured on a Titan X Pascal GPU. %
}\label{tab:detector_cap}
\end{table}

We consider four increasingly large subsets of the \oi{} train set, containing $10$k, $100$k, $1$M and $14.6$M bounding boxes.
We train \AlinaReplaced{all}{both} detectors on exactly the same subsets and test on the publicly released \oi{} test set.
All feature extractors are pre-trained on \ilsvrc{}-2012~\cite{russakovsky15ijcv} for image classification until convergence.
Then, the models are trained for object detection on \oi{} for \num{8}M-\num{20}M steps until convergence on $8$-$24$ NVidia GPUs (Tesla P100, Tesla V100).
For the Faster-RCNN architecture we use momentum optimizer~\cite{qian99nn}, %
while for the SSD architecture we used RMS-prop %
All hyperparameters are kept fixed across all training sets.

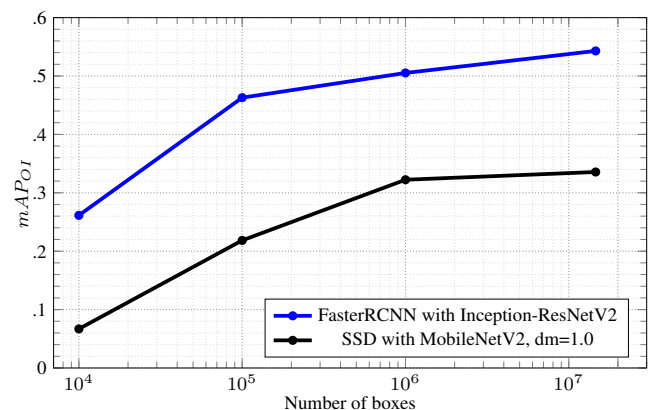
\begin{figure}[h]
\resizebox{\linewidth}{!}{%
\begin{tikzpicture}
\begin{axis}[width=1.3\linewidth,height=0.85\linewidth,
              ylabel=$mAP_{OI}$,
              ylabel shift={-5pt},
              xlabel=Number of boxes,
              xlabel shift={-3pt},
              ytick={0,0.1,...,1.0},
              yticklabels={0,.1,.2,.3,.4,.5,.6},
              xmin=7e3,
              xmax=3e7,
              ymin=0,
              ymax=0.6,
              xmode=log,
              enlargelimits=false,
              grid=both,
              grid style=densely dotted,
              legend pos = south east,
              minor ytick={0,0.02,...,0.6},
              minor grid style={white!85!black},
              major grid style={white!60!black},
              extra x ticks={2e7},
              extra x tick labels={},
              extra tick style={white!85!black}
              ]
  \addplot+[blue,solid,mark=*, mark size=1.2, line width=1.7pt, mark options={fill=blue}] table[x=TrainingSet,y=mAP] {data/mAP_detector/mAP_CHmetric_fasterRCNN_1.0_hier.txt};
  \addlegendentry{FasterRCNN with Inception-ResNetV2}
  \addplot+[black,solid,mark=*, mark size=1.2, line width=1.7pt, mark options={fill=black}] table[x=TrainingSet,y=mAP] {data/mAP_detector/mAP_CHmetric_MobileNetV2_1.0_hier.txt};
  \addlegendentry{SSD with MobileNetV2, dm=1.0}
\end{axis}
\end{tikzpicture}}
\caption{\textbf{Detector performance vs training set size}.
}\label{fig:detector_performance}
\end{figure}

Figure~\ref{fig:detector_performance} reports the results for each combination of deteciton model and training subset size. Generally, the performance of all detectors continuously improves as we add more training data. Faster-RCNN with InceptionV2 \AlinaReplaced{and SSD with Inception-ResNetV2} improve all the way to using all $14.6$M boxes, showing that the very large amount of training data offered by \oi{} is indeed very useful.
The smaller SSD with MobileNetV2 detector saturates at $1$M training boxes.
This suggests that for smaller models \oi{} provides more than enough training data to reach their performance limits.

\new{
\subsection{Visual relationship detection}
Many works have been proposed to tackle the visual relationship detection task in a fully supervised
scenario~\cite{lu2016eccv,liang17cvpr,dai2017cvpr,zhang17cvpr-visual,li17cvpr-vip},
in a weakly supervised setting~\cite{peyre17cvpr,zhang17iccv},
or focusing on human-object relations~\cite{gupta09pami,yao10cvpr-mutual,prest12pami}.
Recently, high-performing models based on deep convolutional neural networks are dominating the field~\cite{gupta2015arxiv,gkioxari17cvpr,gao18bmvc,kolesnikov18arxiv}.

In this section we evaluate two frequency baselines~\cite{zellers18cvpr} as well as a
state-of-the-art visual relationship detection model~\cite{kolesnikov18arxiv}.

\mypar{Tasks and evaluation.}
Traditionally, three main tasks were considered in the visual relationship detection community: relationship detection, phrase detection and preposition detection~\cite{lu2016eccv}.
The first task is the most challenging one, while the other tasks are a relaxation of it.
The performance of visual relationship detection models is often measured either as Recall@50, @100, etc~\cite{lu2016eccv}, or using mean average precision (mAP) as for object detection~\cite{gupta2015arxiv}.
However, until recently VRD datasets did not provide exhaustive annotations (except human-centric datasets~\cite{gupta2015arxiv}). Unfortunately this makes the mAP metric deliver an overly pessimistic assessment, since correct model predictions are sometimes scored as false positives due to missing annotations.
Open Images provides image-level labels that indicate if a given object class was annotated in an image (either as present or absent, Sec.~\ref{sec:image_labels}).
If a class was annotated, then all its object instances are annotated with bounding boxes, and also all its occurrences in visual relationship triplets are also exhaustively annotated.
Hence, mAP can be computed in similar manner as object detection mAP, while ignoring predictions on images where annotation are not present according to the image-level labels (Section \ref{subsec:od_eval}). Thus, there is no risk of incorrectly over-counting false-positives.

In the next paragraphs we provide performance evaluation of several baselines using two metrics:
\begin{itemize}
\item mAP for the visual relationship detection task only (taking into account image-level labels to not penalize correct predictions if ground-truth annotations are missing).
\item the Open Images Challenge metric\footnote{https://storage.googleapis.com/openimages/web/evaluation.html}, which is a weighted average of three metrics: mAP for relationship detection, mAP for phrase detection, and Recall@50.
\end{itemize}

\mypar{Frequency baselines.}
We compute two frequency baselines inspired by~\cite{zellers18cvpr}. As in~\cite{zellers18cvpr}, we name them FREQ and FREQ-OVERLAP.
Let $S$ indicate the subject, $O$ the object and $P$ the relationship preposition connecting two objects.
We first model the probability distribution $\p(S, P, O | I)$ that
a triplet $\langle S, P, O \rangle$ is a correct visual relationship in the input image $I$, using the chain rule of probability.
The joint probability distribution can be decomposed into:
\begin{equation}
 \p(S, P, O | I) = \p(S| I) \cdot \p(P | O, S, I) \cdot \p(O | S, I)  \label{eq:baselines}   
\end{equation}
In the simplest case $\p(P | O, S, I)$ can be computed from the training set distribution as the prior probability to have a certain relationship given a bounding box from a subject $S$ and object $O$, without looking at the image, \ie $\p(P | O, S, I) = \p(P | O, S)$. For the FREQ baseline it is computed using all pairs of boxes in the train set. FREQ-OVERLAP instead is computed using only overlapping pairs of boxes. 
Further, assuming the presence of $O$ is independent of $S$, then $\p(O | S, I) = \p(O | I)$.

To compute the $\p(O | I)$ and $\p(S | I)$ factors we use the FasterRCNN with Inception-ResNetV2 object detection model from Section~\ref{sec:object_detection},
and RetinaNet~\cite{lin17iccv} with ResNet50 that serves as a base model for BAR-CNN.
After the set of detections is produced, we derive the final score for each pair of detections according to Eq.~\eqref{eq:baselines} and using the prior (FREQ baseline).
For the FREQ-OVERLAP baseline, only overlapping pairs of boxes are scored using the corresponding prior. 
In a summary, these baselines use an actual object detector applied to the input image to determine the location of the subject and object boxes, but then determines their relationship based purely on prior probabilities, as learned on the training set.

\mypar{BAR-CNN baseline.}
The BAR-CNN model~\cite{kolesnikov18arxiv} is a conceptually simple model that first predicts all potential subjects in an image and then uses an attention mechanism to attend to each subject in turn and predict all objects connected with it by a relationship. This model is shown to deliver state-of-the-art results despite its simplicity.
We train BAR-CNN with ResNet50 backbone and focal loss~\cite{lin17iccv} on the training set with bounding boxes of and then fine-tune it for the visual relationship detection task using visual relationship annotations.
In contrast to the frequency baselines, BAR-CNN considers the input image also for predicting the relationship, and detects the object conditioned on the subject.

\subsubsection{Performance evaluation}

The evaluation results are presented in Table~\ref{tab:vrd_performance}.
The FREQ and FREQ-OVERLAP baselines score relatively low on the task, even when based on a strong object detector. This indicates that visual relationship detection requires more than simply object detection plus relationship priors.
BAR-CNN instead performs much better than the frequency baselines.
That indicates that there is a lot of extra visual information needed to correctly identify visual relationships on an image.
The result of BAR-CNN can be regarded as a reference for further improvements on the Open Images dataset.

\begin{table}[h]
\resizebox{\linewidth}{!}{%
\begin{tabular}{l@{\hspace{15mm}}cc}
\toprule
 Baseline      & mAP & score \\
\midrule
FREQ & $5.36$ & $18.93$  \\
FREQ-OVERLAP & $8.19$ & $21.47$ \\
FREQ (RetinaNet+ResNet50) & $5.68$ & $19.01$ \\
FREQ-OVERLAP (RetinaNet+ResNet50) & $8.12$ & $21.02$ \\
BAR-CNN (RetinaNet+ResNet50) & $14.63$ & $27.60$ \\
\bottomrule
\end{tabular}}
\caption{\textbf{VRD baselines}: performance of the VRD models on the test set of Open Images. %
}\label{tab:vrd_performance}
\end{table}

} %

\section{The Power of a Unified Dataset}
\label{sec:unification-experiments}
Unification is one of the distinguishing factors of the Open Images dataset, in that the annotations
for image classification, object detection, and visual relationship detection all coexist on the
same images.
In this section we present two experiments that take advantage of the different types of
annotations present in the same images.

\subsection{Fine-grained object detection by combining image-level labels and object bounding boxes}
\label{subsec:finegrained}

The Open Images dataset contains \num{19794} classes annotated at the image-level and \num{600} classes annotated at the box-level.
Bounding boxes provide more precise spatial localization but image-level labels are often semantically
more fine-grained and specific.
Since all classes in Open Images are part of a unified semantic hierarchy%
, we can find the image-level
classes that are more specific (children) than a certain bounding-box class (parent).
As an example, there are image-level classes such as \cls{Volkswagen} or \cls{Labrador} that are
more specific than the bounding-box classes \cls{Car} or \cls{Dog}, respectively.
The experiment in this section shows how we can create a fine-grained object detector (\eg{}\cls{Volkswagen} or \cls{Labrador}) by combining the two types of annotations.

\mypar{Creating fine-grained detection data.}
Given a bounding-box class \cls{cls}, we denote the set of all image-level classes more
specific than \cls{cls} as \cls{C\!(cls)}.
We then look for images where there are boxes of class \cls{cls} and image-level
labels of any class of \cls{C\!(cls)}.
In those images which have only one bounding box of class \cls{cls}, we transfer the more
fine-grained labels \cls{C\!(cls)} to it.
This transfer is safe, as there is only one possible object in that image.

We looked for bounding box classes with a significant number of more specific image-level classes and
selected the following four to experiment with: \cls{Car}, \cls{Flower}, \cls{Cat}, and \cls{Dog}.
We use the procedure above to create fine-grained box labels for these four classes. Statistics are presented in Table~\ref{tab:more_specific_counts}. 
Finally, we use stratified sampling to divide our data into 90\% training images and 10\% test images. 

\begin{table*}
\centering
\resizebox{\linewidth}{!}{%
  \begin{tabular}{lrrl}
    \toprule
    Bounding-box    & Number of         & Number of    & Examples of frequent and infrequent subclasses \cls{C\!(cls)}\\
    class \cls{cls} & subclasses        & samples      & (number of samples in parentheses) \\
    \midrule
    \cls{Car}       & 62                & 4254         & \cls{Ford} (354), \cls{Chevrolet} (223), \dots,
                                                         \cls{Frazer Nash} (2), \cls{Riley Motor} (1)\\
    \cls{Cat}       & 77                & 1340         & \cls{Arabian Mau} (62), \cls{American Wirehair} (56), \dots,
                                                         \cls{Donskoy} (1), \cls{Minskin} (1)\\
    \cls{Dog}       & 39                & 4405         & \cls{Terrier} (245), \cls{Pinscher} (166), \dots,
                                                         \cls{Malshi} (2), \cls{Beaglier} (1)\\
    \cls{Flower}    & 218               & 2541         & \cls{Orchids} (119), \cls{Buttercups} (114), \dots,
                                                         \cls{Tidy tips} (1), \cls{Aechmea `Blue Tango'} (1)\\
    \bottomrule
  \end{tabular}}
  \caption{\new{Statistics of the more specific classes for \cls{Car}, \cls{Flower}, \cls{Cat}, and
  \cls{Dog}. In all cases we report the total number of samples over the training and test sets.}}
  \label{tab:more_specific_counts}
\end{table*}

\begin{table*}
  \centering
  \resizebox{\linewidth}{!}{%
  \begin{tabular}{l@{\hspace{20mm}}rrrrr}
    \toprule
    \multirow{2}{*}{General class \cls{cls}}   &  Num.\ of fine-grained  & Uniform random  & Most common     & Prior-based random  & Image label transfer \\
                     &  classes ($\geq 5$ samples) & sub-class (mAP) & sub-class (mAP) & sub-class (mAP)     & (ours) (mAP)         \\
    \midrule                                                             
    \cls{Car}        &   57                    & \num{0.008}     & \num{0.002}     & \num{0.011}         & \num{0.287}          \\
    \cls{Cat}        &   61                    & \num{0.010}     & \num{0.002}     & \num{0.041}         & \num{0.231}          \\
    \cls{Dog}        &   33                    & \num{0.018}     & \num{0.011}     & \num{0.010}         & \num{0.272}          \\
    \cls{Flower}     &  102                    & \num{0.002}     & \num{0.000}     & \num{0.009}         & \num{0.594}          \\
    \bottomrule
  \end{tabular}}
  \caption{\new{Results on fine-grained detection over subclasses of \cls{Car}, \cls{Flower}, \cls{Cat}, and
  \cls{Dog}.}}
  \label{tab:fine_grained_detection}
\end{table*}

\mypar{Experimental setup.}
We evaluate on fine-grained classes which have at least 4 training samples and at least 1 test sample.
We train a single Faster-RCNN detector~\cite{ren15nips} with an Inception-ResNetV2
backbone~\cite{szegedy17aaai} (like in Sec.~\ref{subsubsec:capacity}) on the
fine-grained classes.
We apply this detector on the test set and report the Average Precision at IoU $>0.5$,
averaged over the fine-grained classes within each general class
(mAP over \cls{Car}, \cls{Cat}, \cls{Dog}, \cls{Flower}).

We also report the performance of three baselines, all based on the same Faster-RCNN architecture as above but trained on the four general classes (\cls{Car}, \cls{Cat}, \cls{Dog}, \cls{Flower}).
The first baseline assigns each detection of a general class to one of its subclasses sampled uniformly at random.
The second baseline instead assigns each general detection to its the most frequent subclass.
Finally, the third baseline assigns each general detection to a random subclass sampled according to their prior probabilities (as observed on the training set).
Note that our second and third baselines require statistics of the subclasses, which are not available when considering only the box-level classes.

\begin{figure*}
\setlength{\fboxsep}{0pt}
  \resizebox{\linewidth}{!}{\mbox{%
  \fbox{\includegraphics[height=5cm]{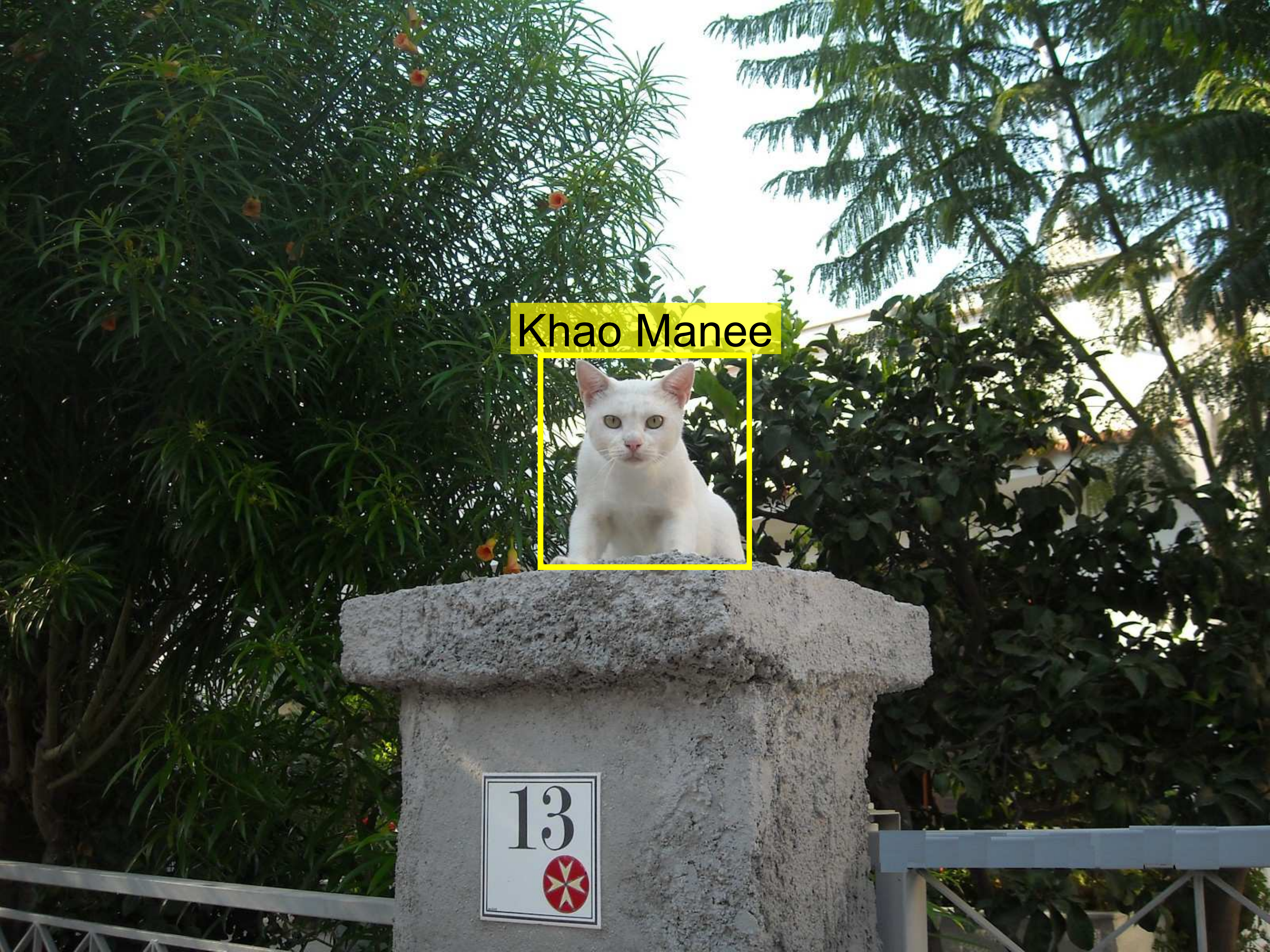}}\hspace{0.4mm}
  \fbox{\includegraphics[height=5cm]{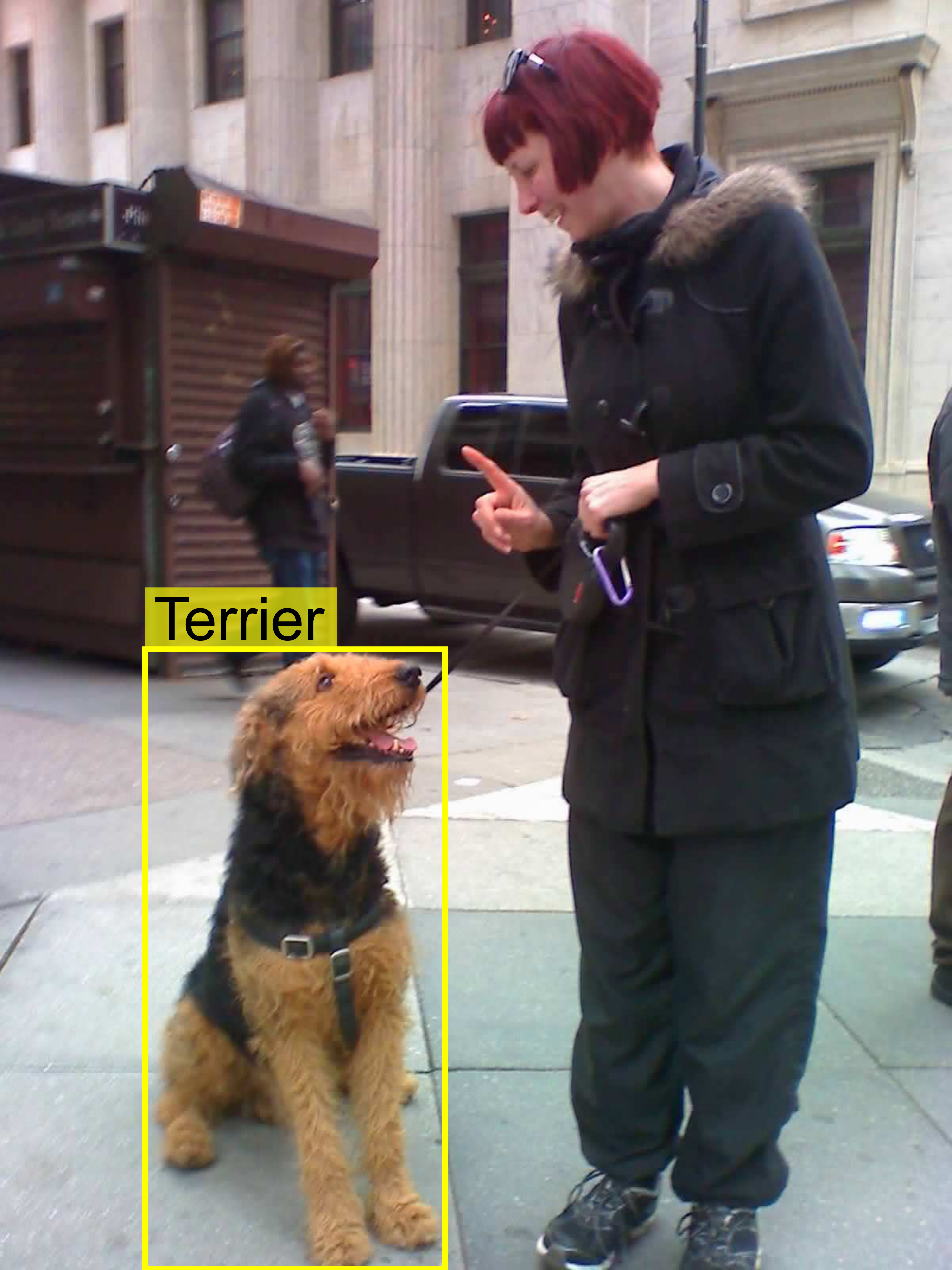}}\hspace{0.4mm}
  \fbox{\includegraphics[height=5cm]{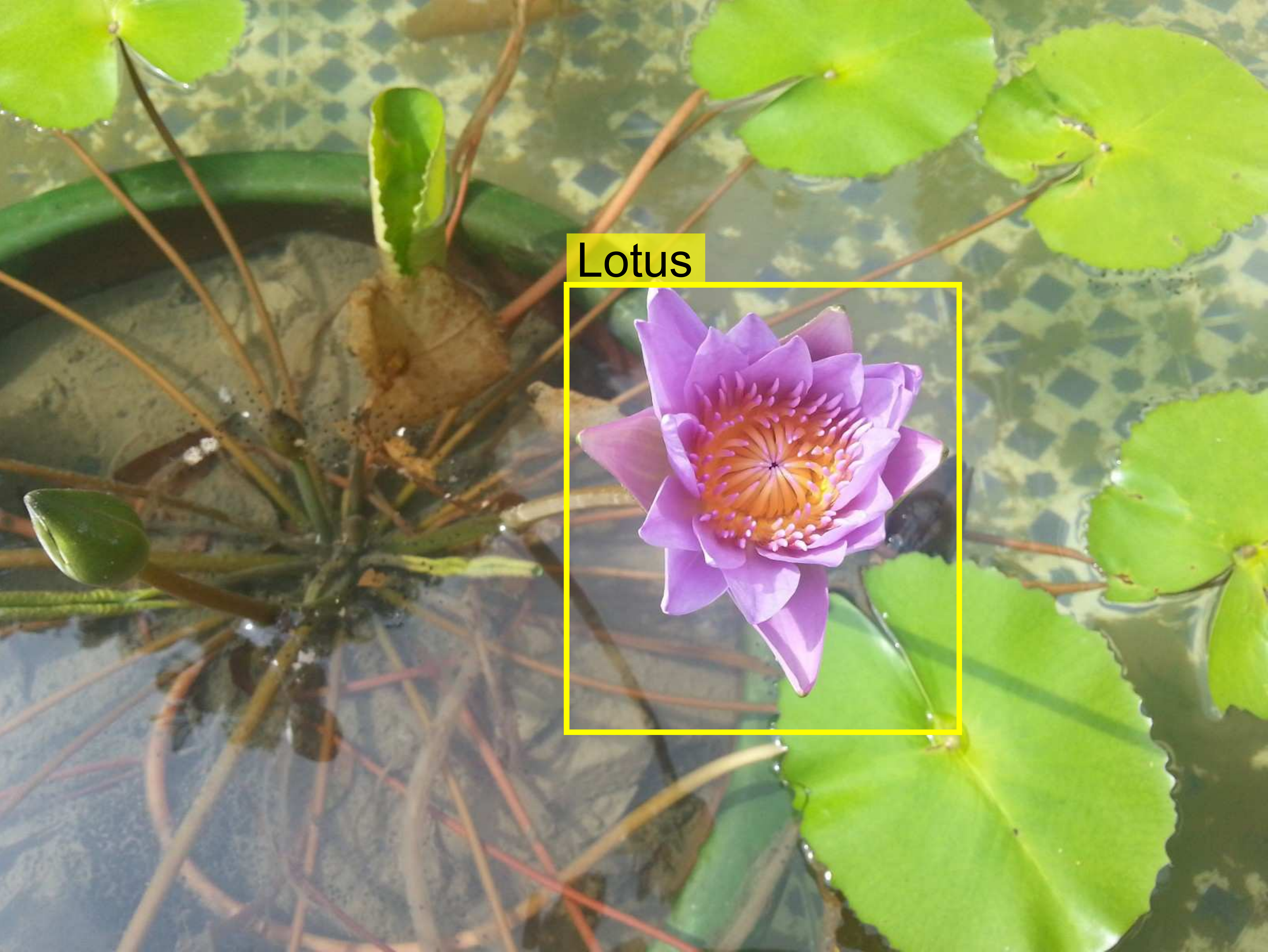}}}}\\[0.8mm]
    \resizebox{\linewidth}{!}{\mbox{%
  \fbox{\includegraphics[height=5cm]{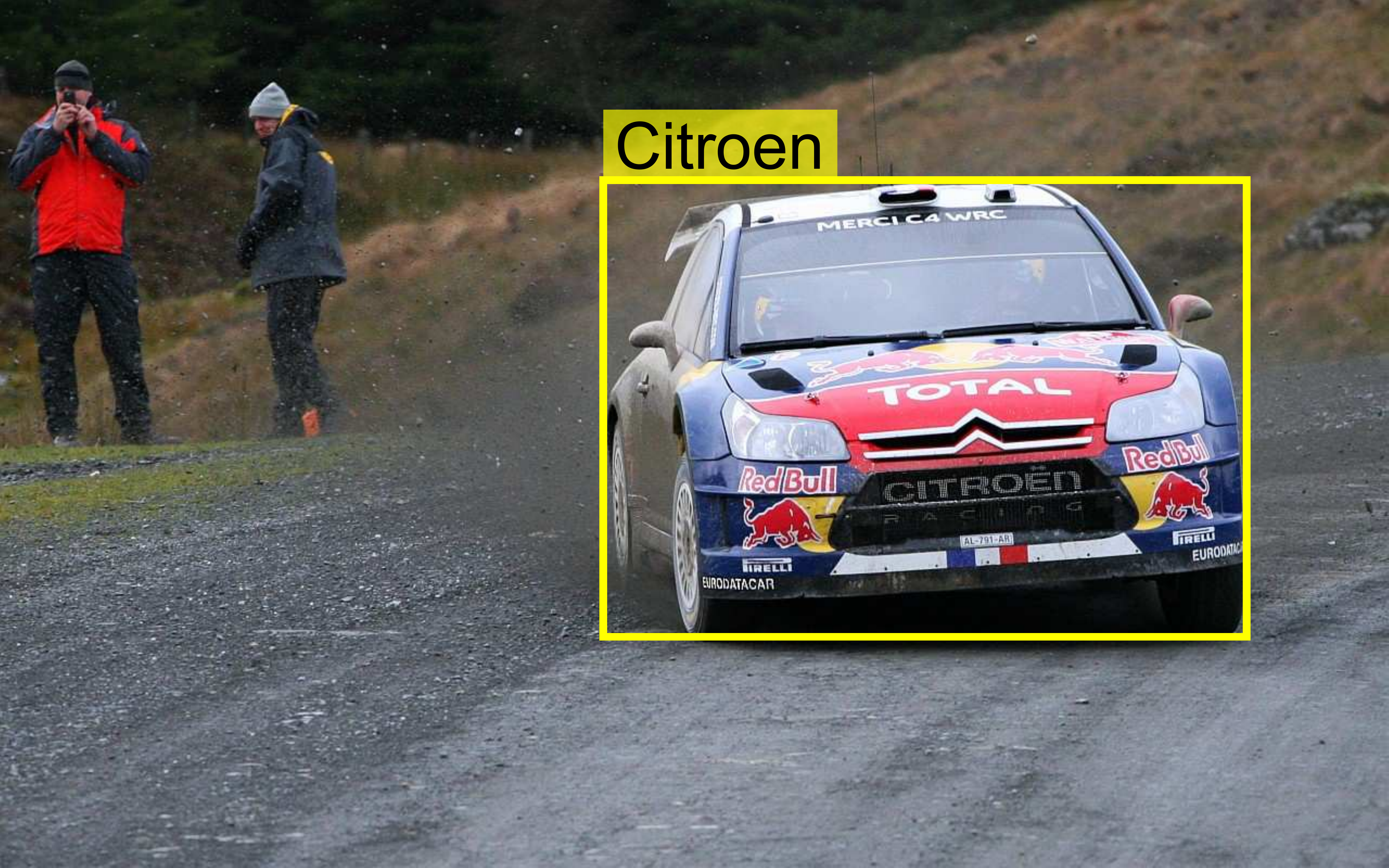}}\hspace{0.8mm}
  \fbox{\includegraphics[height=5cm]{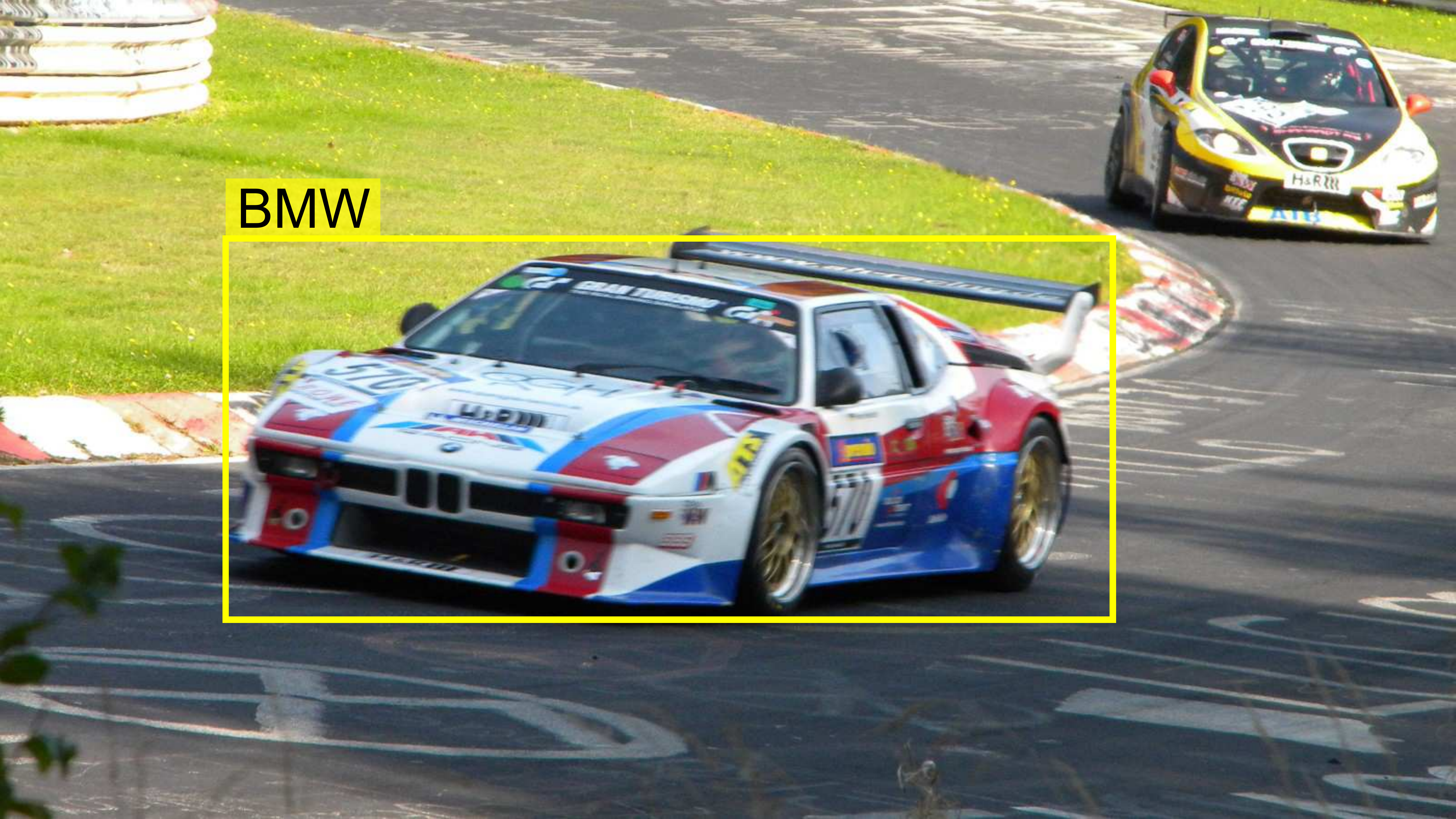}}\hspace{0.8mm}
  \fbox{\includegraphics[height=5cm]{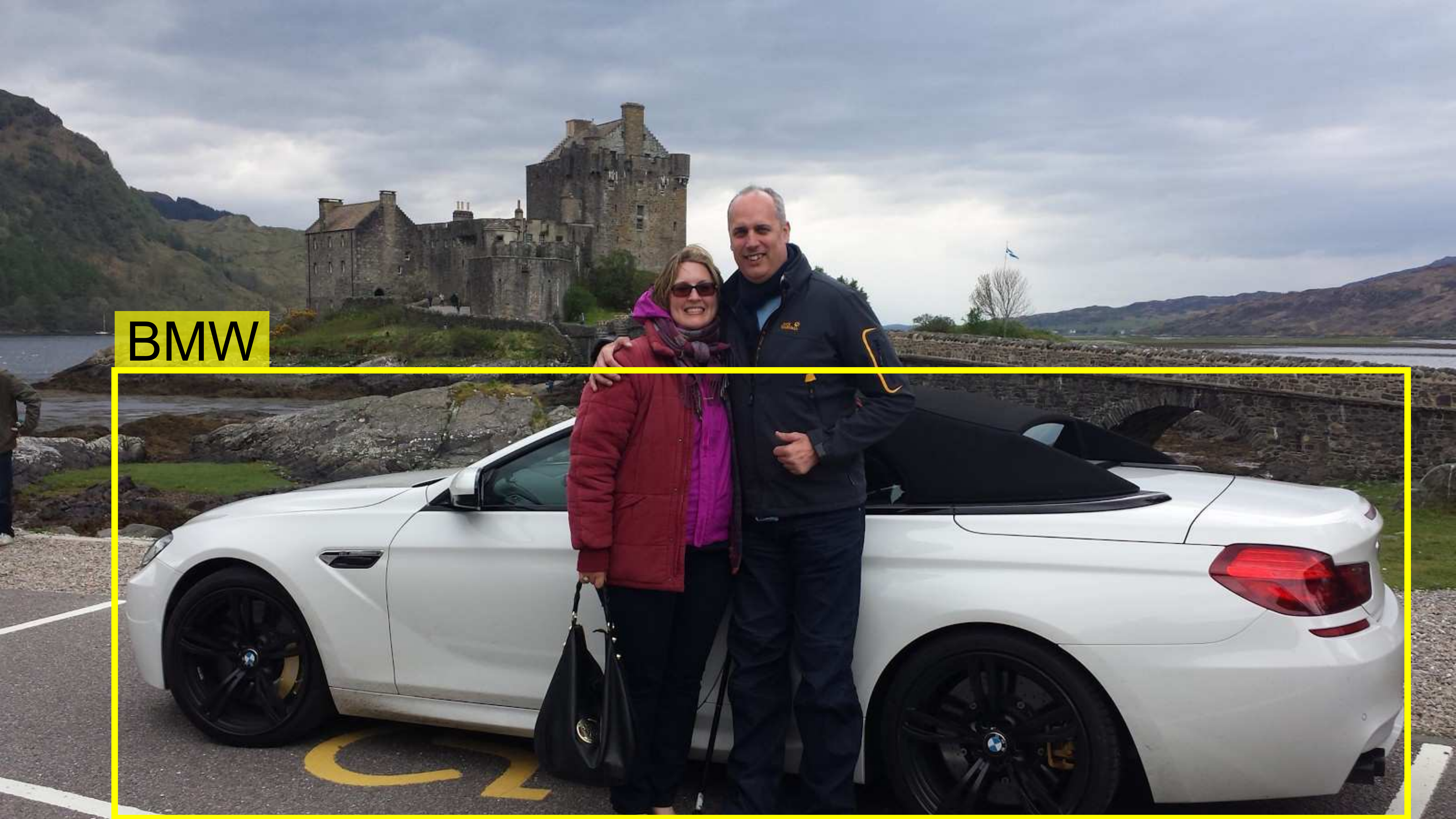}}}}
  \caption{\new{Example output of our fine-grained detectors.
  Note that our detector correctly identifies the individual car brands even in
  race cars whose appearance has been heavily modified (\eg{}two bottom left examples).}}
  \label{fig:examples_finegrained}
\end{figure*}

\mypar{Results.}
Results are presented in Table~\ref{tab:fine_grained_detection}.
While all baselines yield poor results below $< 0.05$ mAP, our method delivers
decent fine-grained detectors, with mAP ranging from \num{0.231} over the \num{61}
subclasses of \cls{Cat}, to \num{0.594} over the 102 subclasses of \cls{Flower}.
Interestingly, for several subclasses we have very good results suggesting that
these classes are very distinctive, \eg{}\cls{Ferrari} (0.638 mAP),
\cls{Land Rover} (0.620 mAP), and \cls{Schnauzer} (0.542 mAP).
Several example detections are shown in Figure~\ref{fig:examples_finegrained}.
These results demonstrate that the unified annotations of \oi{} enable to train
object detectors for fine-grained classes despite having only bounding box
annotations for their parent class.

\subsection{Zero-shot visual relationship detection by combining object bounding
boxes and visual relationships}\label{subsec:vrd_zeroshot}

\begin{table*}[th]
\resizebox{\linewidth}{!}{%
  \begin{tabular}{l@{\hspace{40mm}}rrrr}
  \toprule
   Metric & R@50 (all triplets) & R@100 (all triplets) & R@50 (zero-shot)  & R@100 (zero-shot) \\
  \midrule
  Relationship detection & $40.61$ & $40.93$ & $7.68$ & $7.70$ \\
  Phrase detection & $43.65$ & $43.86$ & $10.98$ & $11.08$ \\
  \bottomrule
  \end{tabular}}
  \caption{\textbf{Zero-shot visual relationship detection results}.
  We evaluate a single model trained on the existing VRD annotations and box annotations.
  We report performance on all test set annotations including supervised and zero-shot triplets, as well as the performance on zero-shot-only triplets in terms of recall for relationship and predicate detection~\cite{lu2016eccv}.}
  \label{tab:vrd_zeroshot}
\end{table*}
\begin{table}
   \centering
  \resizebox{\linewidth}{!}{%
  \begin{tabular}{lr}
    \toprule
    Relationship & Zero-shot triplets \\
    \midrule   
    plays ($26$) & \begin{tabular}{@{}c@{}}
    \lara{\cls{Girl}, plays, \cls{\textbf{Cello}}}, \lara{\cls{Man}, plays, \cls{\textbf{Saxophone}}} , \dots \\
    \end{tabular} \\    
    holds ($70$) & \begin{tabular}{@{}c@{}}
    \lara{\cls{Man}, holds, \cls{\textbf{Bicycle wheel}}}, \lara{\cls{Boy}, holds, \cls{\textbf{Skateboard}}} , \dots \\
    \end{tabular} \\      
    wears ($76$) & \begin{tabular}{@{}c@{}}
    \lara{\cls{Woman}, wears, \cls{\textbf{Scarf}}}, \lara{\cls{Man}, wears, \cls{\textbf{Glasses}}} , \dots, \\
    \lara{\cls{Girl}, wears, \cls{\textbf{Necklace}}}
    \end{tabular} \\      
    on ($10$) & \begin{tabular}{@{}c@{}}
    \lara{\cls{Spoon}, on, \cls{\textbf{Cutting board}}}, \lara{\cls{Dog}, on, \cls{\textbf{Dog bed}}} , \dots \\
    \end{tabular} \\
    inside of ($7$) &  \begin{tabular}{@{}c@{}}
 \lara{\cls{Woman}, inside of, \cls{\textbf{Golf cart}}}, \lara{\cls{Girl}, inside of, \cls{\textbf{Bus}}}, \dots   \\   
\end{tabular} \\
at ($4$) & \begin{tabular}{@{}c@{}}
    \lara{\cls{Boy}, at, \cls{\textbf{Billiard Table}}}, \lara{\cls{Girl}, at, \cls{\textbf{Billiard Table}}}, \\
    \lara{\cls{Woman}, at, \cls{\textbf{Billiard Table}}}, \lara{\cls{Man}, at, \cls{\textbf{Billiard Table}}}  
    \end{tabular} \\
under ($1$) & \begin{tabular}{@{}c@{}}
    \lara{\cls{Cat}, under, \cls{\textbf{Coffee Table}}} \\
    \end{tabular} \\    
    \bottomrule
  \end{tabular}}
  \caption{\textbf{Examples of zero-shot relationship triplets}, involving the new classes in bold.
  In parentheses, the total number of zero-shot triplets for each relationship.}
 \label{tab:zeroshot_vrd}
\end{table}

In the classical zero-shot Visual Relationship Detection (VRD) task, the
zero-shot triplets consist only of new combinations of classes appearing in
other annotated relationships~\cite{lu2016eccv,liang18aaai}, \eg{} detecting
\lara{\cls{Cat}, under, \cls{Table}} when the training set contains
\lara{\cls{Cat}, behind, \cls{Door}} and
\lara{\cls{Person}, under, \cls{Table}}.
We propose to detect relationships also for new classes that are not
present in any relationship annotations, by leveraging the bounding box annotations
that co-exists in the same images as relationship annotations.
This showcases the benefits of the unified annotations in \oi{}.
In particular, we combine training data for related triplets that have
the same subject and relationship with bounding boxes for a new object, \eg{} detecting
\lara{\cls{Cat}, under, \cls{Car}} when we only have
\lara{\cls{Cat}, under, \cls{Table}} annotated as relationship and \cls{Car} as
bounding boxes.
Analogously, we also introduce new subjects.

\begin{figure*}
  \setlength{\fboxsep}{0pt}
  \resizebox{\linewidth}{!}{\mbox{%
  \fbox{\includegraphics[height=5cm]{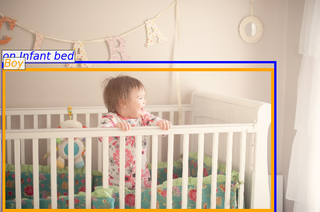}}\hspace{0.4mm}
  \fbox{\includegraphics[height=5cm]{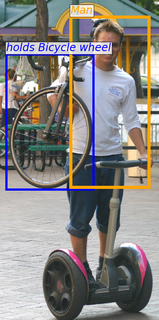}}
  \fbox{\includegraphics[height=5cm]{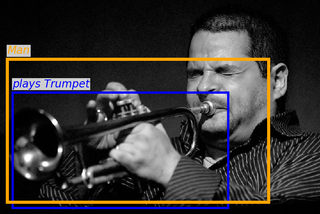}}\hspace{0.8mm}
  \fbox{\includegraphics[height=5cm]{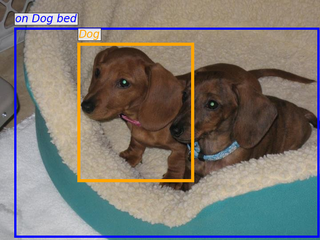}}}}
  \caption{\textbf{Example zero-shot detections} on the \oi{} test set.
  Yellow boxes denote the subjects and blue boxes denote the objects and relationships.
  For these examples, object classes are zero-shot.}
  \label{fig:examples_vrdzeroshot}
\end{figure*}

\begin{figure}
  \setlength{\fboxsep}{0pt}
  \resizebox{\linewidth}{!}{\mbox{%
  \fbox{\includegraphics[height=5cm]{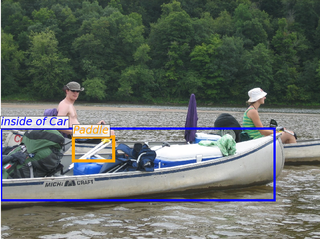}}\hspace{0.4mm}
  \fbox{\includegraphics[height=5cm]{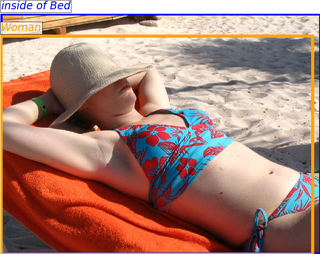}}\hspace{0.4mm}
  }}\\[0.8mm]
  \caption{\textbf{Interesting wrong predictions of the zero-shot model}.
  \textbf{Left image}: BAR-CNN detects the subject and relationship preposition correctly
  but the object label is incorrect (\cls{Boat} is a zero-shot class).
  \textbf{Right image}: BAR-CNN only detects the subject correctly but
  both the relationship and the object are very close in meaning to what is shown in
  the image.}
  \label{fig:examples_vrderrors}
\end{figure}

Specifically, we select new classes that are among the $600$ with boxes in
\oi{} and do not have relationship annotations, and use them as our zero-shot classes.
For evaluation purposes, we annotated the additional visual relationships for
the new classes on the test split.
We then filtered the set of new zero-shot classes as those that are frequent
enough (there are at least $10$ instances of visual relationships in the test
split that contain that class).
Overall, we obtain $194$ new triplets with a relationship preposition existing
in the training split, where either subject or object are from
the set of the new zero-shot classes.
As a result, we have \num{5983} new zero-shot triplet annotations on the test
split, covering $48$ new zero-shot classes, from which we have $284$k annotated
bounding boxes in the training split.
Table~\ref{tab:zeroshot_vrd} shows examples of the new object classes and the
relationship triplets they are involved with.

We train the BAR-CNN model~\cite{kolesnikov18arxiv} using the annotated
visual relationships on the training set and the additional set of $48$ classes
for which only box annotations are available.
During training, BAR-CNN accepts two types of samples: train samples
with a single subject class label and train samples with object and relationship
class labels.
A combination of classification and box regression loss is optimized,
as described in~\cite{lin17iccv}.
In BAR-CNN, a sigmoid cross-entropy loss is used to handle multi-class samples
with object and relationship class labels.
Since for the zero-shot object samples the relationship class labels are not available,
we mask out the sigmoid loss components for the object samples without
relationship class labels (those are samples derived from box annotations).

Table~\ref{tab:vrd_zeroshot} presents our quantitative evaluation.
The results show that the highly challenging setting of zero-shot VRD can be
tackled to a reasonable degree but the gap to the supervised results is still very large,
indicating potential for further exploration of the task with help of the unified
annotations of \oi{}.
Figure~\ref{fig:examples_vrdzeroshot} shows some examples of BAR-CNN zero-shot
predictions and Figure~\ref{fig:examples_vrderrors} shows
interesting prediction examples: incorrectly detected object
class and incorrectly detected relationship (according to defined triplets); note that in both cases general semantics is preserved.

\section{Conclusions}
\label{sec:conclusions}
This paper presented \oi{} V4, a collection of \num{9.2} million images annotated with unified ground-truth for image classification, object detection, and visual relationship detection.
We explained how the data was collected and annotated, we presented comprehensive dataset statistics, we evaluated its quality, and we reported the performance of several modern models for image classification and object detection.
We hope that the scale, quality, and variety of \oi{} V4 will foster further research and innovation even beyond the areas of image classification, object detection, and visual relationship detection.

\section{Credits}
This is a full list of the contributors to the \oid{}, which goes beyond the authors of this paper. Thanks to everyone!

\begin{itemize}
\item[\textit{Project Lead and Coordination}:] Vittorio Ferrari, Tom Duerig, and Victor Gomes.
\item[\textit{Image collection}:]
Ivan Krasin,
David Cai.
\item[\textit{Image-level labels}:]
Neil Alldrin,
Ivan Krasin,
Shahab Kamali,
Tom Duerig,
Zheyun Feng,
Anurag Batra,
Alok Gunjan.
\item[\textit{Bounding boxes}:]
Hassan Rom,
Alina Kuznetsova,
Jasper Uij\-lings,
Stefan Popov,
Matteo Malloci,
Sami Abu-El-Haija,
Vittorio Ferrari.
\item[\textit{Visual relationships}:]
Alina Kuznetsova,
Matteo Malloci,
Vittorio Ferrari.
\item[\textit{Website and visualizer}:]
Jordi Pont-Tuset.
\item[\textit{Classes and hierarchy}:]
Chen Sun,
Kevin Murphy,
Tom Duerig,
Vittorio Ferrari.
\item[\textit{Challenge}:]
Vittorio Ferrari,
Alina Kuznetsova,
Jordi Pont-Tuset,
Matteo Malloci,
Jasper Uijlings,
Jake Walker,
Rodrigo Benenson.
\item[\textit{Advisers}:]
Andreas Veit,
Serge Belongie,
Abhinav Gupta,
Dhyanesh Narayanan,
Gal Chechik.
\end{itemize}

\bibliographystyle{spbasic}      %
\bibliography{shortstrings,loco}   %

\begin{thebibliography}{59}
\providecommand{\natexlab}[1]{#1}
\providecommand{\url}[1]{{#1}}
\providecommand{\urlprefix}{URL }
\expandafter\ifx\csname urlstyle\endcsname\relax
  \providecommand{\doi}[1]{DOI~\discretionary{}{}{}#1}\else
  \providecommand{\doi}{DOI~\discretionary{}{}{}\begingroup
  \urlstyle{rm}\Url}\fi
\providecommand{\eprint}[2][]{\url{#2}}

\bibitem[{Alexe et~al.(2010)Alexe, Deselaers, and Ferrari}]{alexe10cvpr}
Alexe B, Deselaers T, Ferrari V (2010) What is an object? In: CVPR

\bibitem[{Alexe et~al.(2012)Alexe, Deselaers, and Ferrari}]{alexe12pami}
Alexe B, Deselaers T, Ferrari V (2012) Measuring the objectness of image
  windows. IEEE Trans on PAMI

\bibitem[{Chollet(2017)}]{chollet17cvpr}
Chollet F (2017) Xception: Deep learning with depthwise separable convolutions.
  In: CVPR

\bibitem[{Dai et~al.(2017)Dai, Zhang, and Lin}]{dai2017cvpr}
Dai B, Zhang Y, Lin D (2017) Detecting visual relationships with deep
  relational networks. In: CVPR

\bibitem[{Deng et~al.(2009)Deng, Dong, Socher, Li, Li, and
  Fei-fei}]{deng09cvpr}
Deng J, Dong W, Socher R, Li LJ, Li K, Fei-fei L (2009) {ImageNet}: {A}
  large-scale hierarchical image database. In: CVPR

\bibitem[{Everingham et~al.(2010)Everingham, Van~Gool, Williams, Winn, and
  Zisserman}]{everingham10ijcv}
Everingham M, Van~Gool L, Williams CKI, Winn J, Zisserman A (2010) {The
  {PASCAL} Visual Object Classes ({VOC}) Challenge}. IJCV

\bibitem[{Everingham et~al.(2012)Everingham, Van~Gool, Williams, Winn, and
  Zisserman}]{pascal-voc-2012}
Everingham M, Van~Gool L, Williams CKI, Winn J, Zisserman A (2012) The {PASCAL}
  {V}isual {O}bject {C}lasses {C}hallenge 2012 {(VOC2012)} {R}esults.
  http://www.pascal-network.org/challenges/VOC/voc2012/workshop/index.html

\bibitem[{Everingham et~al.(2015)Everingham, Eslami, van Gool, Williams, Winn,
  and Zisserman}]{everingham15ijcv}
Everingham M, Eslami S, van Gool L, Williams C, Winn J, Zisserman A (2015) The
  {PASCAL} visual object classes challenge: A retrospective. IJCV

\bibitem[{Fei-Fei et~al.(2006)Fei-Fei, Fergus, and Perona}]{fei-fei:pami06}
Fei-Fei L, Fergus R, Perona P (2006) One-shot learning of object categories.
  IEEE Trans on PAMI 28(4):594--611

\bibitem[{Felzenszwalb et~al.(2010{\natexlab{a}})Felzenszwalb, Girshick, and
  McAllester}]{felzenszwalb10cvpr_b}
Felzenszwalb P, Girshick R, McAllester D (2010{\natexlab{a}}) {Cascade Object
  Detection with Deformable Part Models}. In: CVPR

\bibitem[{Felzenszwalb et~al.(2010{\natexlab{b}})Felzenszwalb, Girshick,
  McAllester, and Ramanan}]{felzenszwalb10pami}
Felzenszwalb P, Girshick R, McAllester D, Ramanan D (2010{\natexlab{b}}) Object
  detection with discriminatively trained part based models. IEEE Trans on PAMI
  32(9)

\bibitem[{Gao et~al.(2018)Gao, Zou, and Huang}]{gao18bmvc}
Gao C, Zou Y, Huang JB (2018) {iCAN}: Instance-centric attention network for
  human-object interaction detection. In: BMVC

\bibitem[{Girshick(2015)}]{girshick15iccv}
Girshick R (2015) Fast {R-CNN}. In: ICCV

\bibitem[{Girshick et~al.(2014)Girshick, Donahue, Darrell, and
  Malik}]{girshick14cvpr}
Girshick R, Donahue J, Darrell T, Malik J (2014) Rich feature hierarchies for
  accurate object detection and semantic segmentation. In: CVPR

\bibitem[{Gkioxari et~al.(2018)Gkioxari, Girshick, Doll{\'a}r, and
  He}]{gkioxari17cvpr}
Gkioxari G, Girshick R, Doll{\'a}r P, He K (2018) Detecting and recognizing
  human-object interactions. CVPR

\bibitem[{Griffin et~al.(2007)Griffin, Holub, and Perona}]{caltech256}
Griffin G, Holub A, Perona P (2007) The {Caltech-256}. Tech. rep., Caltech

\bibitem[{Gupta et~al.(2009)Gupta, Kembhavi, and Davis}]{gupta09pami}
Gupta A, Kembhavi A, Davis L (2009) Observing human-object interactions: Using
  spatial and functional compatibility for recognition. In: IEEE Trans. on PAMI

\bibitem[{Gupta and Malik(2015)}]{gupta2015arxiv}
Gupta S, Malik J (2015) Visual semantic role labeling. arXiv preprint
  arXiv:150504474

\bibitem[{He et~al.(2016)He, Zhang, Ren, and Sun}]{he16cvpr}
He K, Zhang X, Ren S, Sun J (2016) Deep residual learning for image
  recognition. In: CVPR

\bibitem[{Hinton et~al.(2014)Hinton, Vinyals, and Dean}]{hinton14nips}
Hinton GE, Vinyals O, Dean J (2014) Distilling the knowledge in a neural
  network. In: NeurIPS

\bibitem[{Huang et~al.(2017)Huang, Rathod, Sun, Zhu, Korattikara, Fathi,
  Fischer, Wojna, Song, Guadarrama, and Murphy.}]{huang17cvpr}
Huang J, Rathod V, Sun C, Zhu M, Korattikara A, Fathi A, Fischer I, Wojna Z,
  Song Y, Guadarrama S, Murphy K (2017) Speed/accuracy trade-offs for modern
  convolutional object detectors. In: CVPR

\bibitem[{Ioffe and Szegedy(2015)}]{ioffe15icml}
Ioffe S, Szegedy C (2015) Batch normalization: Accelerating deep network
  training by reducing internal covariate shift. In: ICML

\bibitem[{Kolesnikov et~al.(2018)Kolesnikov, Kuznetsova, Lampert, and
  Ferrari}]{kolesnikov18arxiv}
Kolesnikov A, Kuznetsova A, Lampert C, Ferrari V (2018) Detecting visual
  relationships using box attention. arXiv 1807.02136

\bibitem[{Krishna et~al.(2017)Krishna, Zhu, Groth, Johnson, Hata, Kravitz,
  Chen, Kalantidis, Li, Shamma, Bernstein, and Fei-Fei}]{krishna17ijcv}
Krishna R, Zhu Y, Groth O, Johnson J, Hata K, Kravitz J, Chen S, Kalantidis Y,
  Li LJ, Shamma DA, Bernstein M, Fei-Fei L (2017) Visual genome: Connecting
  language and vision using crowdsourced dense image annotations. IJCV
  123(1):32--73

\bibitem[{Krizhevsky(2009)}]{krizhevsky09}
Krizhevsky A (2009) Learning multiple layers of features from tiny images.
  Tech. rep., University of Toronto

\bibitem[{Krizhevsky et~al.(2012)Krizhevsky, Sutskever, and
  Hinton}]{krizhevsky12nips}
Krizhevsky A, Sutskever I, Hinton GE (2012) Imagenet classification with deep
  convolutional neural networks. In: NeurIPS

\bibitem[{Li et~al.(2017)Li, Ouyang, Wang, and Tang}]{li17cvpr-vip}
Li Y, Ouyang W, Wang X, Tang X (2017) {ViP-CNN}: Visual phrase guided
  convolutional neural network. In: CVPR

\bibitem[{Liang et~al.(2018)Liang, Guo, Chang, and Chen}]{liang18aaai}
Liang K, Guo Y, Chang H, Chen X (2018) Visual relationship detection with deep
  structural ranking. In: AAAI

\bibitem[{Liang et~al.(2017)Liang, Lee, and Xing}]{liang17cvpr}
Liang X, Lee L, Xing EP (2017) Deep variation-structured reinforcement learning
  for visual relationship and attribute detection. In: CVPR

\bibitem[{{Lin} et~al.(2017){Lin}, {Goyal}, {Girshick}, {He}, and
  {Dollar}}]{lin17iccv}
{Lin} T, {Goyal} P, {Girshick} R, {He} K, {Dollar} P (2017) Focal loss for
  dense object detection. In: ICCV

\bibitem[{Lin et~al.(2014)Lin, Maire, Belongie, Bourdev, Girshick, Hays,
  Perona, Ramanan, Zitnick, and Doll\'{a}r}]{lin14eccv}
Lin TY, Maire M, Belongie S, Bourdev L, Girshick R, Hays J, Perona P, Ramanan
  D, Zitnick CL, Doll\'{a}r P (2014) Microsoft {COCO}: Common objects in
  context. In: ECCV

\bibitem[{Liu et~al.(2016)Liu, Anguelov, Erhan, Szegedy, Reed, Fu, and
  Berg}]{liu16eccv}
Liu W, Anguelov D, Erhan D, Szegedy C, Reed S, Fu CY, Berg AC (2016) {SSD}:
  Single shot multibox detector. In: ECCV

\bibitem[{Lu et~al.(2016)Lu, Krishna, Bernstein, and Fei-Fei}]{lu2016eccv}
Lu C, Krishna R, Bernstein M, Fei-Fei L (2016) Visual relationship detection
  with language priors. In: European Conference on Computer Vision

\bibitem[{Mikolov et~al.(2013)Mikolov, Sutskever, Chen, Corrado, and
  Dean}]{mikolov13nips}
Mikolov T, Sutskever I, Chen K, Corrado GS, Dean J (2013) Distributed
  representations of words and phrases and their compositionality. In: NeurIPS

\bibitem[{Papadopoulos et~al.(2016)Papadopoulos, Uijlings, Keller, and
  Ferrari}]{papadopoulos16cvpr}
Papadopoulos DP, Uijlings JRR, Keller F, Ferrari V (2016) We don't need no
  bounding-boxes: Training object class detectors using only human
  verification. In: CVPR

\bibitem[{Papadopoulos et~al.(2017)Papadopoulos, Uijlings, Keller, and
  Ferrari}]{papadopoulos17iccv}
Papadopoulos DP, Uijlings JR, Keller F, Ferrari V (2017) Extreme clicking for
  efficient object annotation. In: ICCV

\bibitem[{Peyre et~al.(2017)Peyre, Laptev, Schmid, and Sivic}]{peyre17cvpr}
Peyre J, Laptev I, Schmid C, Sivic J (2017) Weakly-supervised learning of
  visual relations. In: CVPR

\bibitem[{Prest et~al.(2012)Prest, Schmid, and Ferrari}]{prest12pami}
Prest A, Schmid C, Ferrari V (2012) Weakly supervised learning of interactions
  between humans and objects. IEEE Trans on PAMI

\bibitem[{Qian(1999)}]{qian99nn}
Qian N (1999) On the momentum term in gradient descent learning algorithms.
  Neural Networks 12(1):145--151

\bibitem[{Redmon and Farhadi(2017)}]{redmon17cvpr}
Redmon J, Farhadi A (2017) {YOLO9000}: better, faster, stronger. In: CVPR

\bibitem[{Redmon et~al.(2016)Redmon, Divvala, Girshick, and
  Farhadi}]{redmon16cvpr}
Redmon J, Divvala S, Girshick R, Farhadi A (2016) You only look once: Unified,
  real-time object detection. In: CVPR

\bibitem[{Ren et~al.(2015)Ren, He, Girshick, and Sun}]{ren15nips}
Ren S, He K, Girshick R, Sun J (2015) Faster {R-CNN}: Towards real-time object
  detection with region proposal networks. In: NeurIPS

\bibitem[{Russakovsky et~al.(2015)Russakovsky, Deng, Su, Krause, Satheesh, Ma,
  Huang, Karpathy, Khosla, Bernstein, Berg, and Fei-Fei}]{russakovsky15ijcv}
Russakovsky O, Deng J, Su H, Krause J, Satheesh S, Ma S, Huang Z, Karpathy A,
  Khosla A, Bernstein M, Berg A, Fei-Fei L (2015) {ImageNet} large scale visual
  recognition challenge. IJCV

\bibitem[{Sandler et~al.(2018)Sandler, Howard, Zhu, Zhmoginov, and
  Chen}]{sandler18cvpr}
Sandler M, Howard AG, Zhu M, Zhmoginov A, Chen L (2018) Mobilenetv2: Inverted
  residuals and linear bottleneck. In: CVPR

\bibitem[{Su et~al.(2012)Su, Deng, and Fei-Fei}]{su12aaai}
Su H, Deng J, Fei-Fei L (2012) Crowdsourcing annotations for visual object
  detection. In: AAAI Human Computation Workshop

\bibitem[{Sun et~al.(2017)Sun, Shrivastava, Singh, and Gupta}]{sun17iccv}
Sun C, Shrivastava A, Singh S, Gupta A (2017) Revisiting unreasonable
  effectiveness of data in deep learning era. In: ICCV

\bibitem[{Szegedy et~al.(2015)Szegedy, Liu, Jia, Sermanet, Reed, Anguelov,
  Erhan, Vanhoucke, and Rabinovich}]{szegedy15cvpr}
Szegedy C, Liu W, Jia Y, Sermanet P, Reed S, Anguelov D, Erhan D, Vanhoucke V,
  Rabinovich A (2015) Going deeper with convolutions. In: CVPR

\bibitem[{Szegedy et~al.(2016)Szegedy, Vanhoucke, Ioffe, Shlens, and
  Wojna}]{szegedy16cvpr}
Szegedy C, Vanhoucke V, Ioffe S, Shlens J, Wojna Z (2016) Rethinking the
  inception architecture for computer vision. In: CVPR

\bibitem[{Szegedy et~al.(2017)Szegedy, Ioffe, Vanhoucke, and
  Alemi}]{szegedy17aaai}
Szegedy C, Ioffe S, Vanhoucke V, Alemi A (2017) Inception-v4, inception-resnet
  and the impact of residual connections on learning. In: AAAI

\bibitem[{Uijlings et~al.(2018)Uijlings, Popov, and Ferrari}]{uijlings18cvpr}
Uijlings J, Popov S, Ferrari V (2018) Revisiting knowledge transfer for
  training object class detectors. In: CVPR

\bibitem[{Uijlings et~al.(2013)Uijlings, van~de Sande, Gevers, and
  Smeulders}]{uijlings13ijcv}
Uijlings JRR, van~de Sande KEA, Gevers T, Smeulders AWM (2013) Selective search
  for object recognition. IJCV

\bibitem[{Veit et~al.(2017)Veit, Alldrin, Chechik, Krasin, Gupta, and
  Belongie}]{Veitcvpr17}
Veit A, Alldrin N, Chechik G, Krasin I, Gupta A, Belongie S (2017) Learning
  from noisy large-scale datasets with minimal supervision. In: Proceedings of
  the IEEE Conference on Computer Vision and Pattern Recognition, pp 839--847,
  \urlprefix\url{http://openaccess.thecvf.com/content_cvpr_2017/papers/Veit_Learning_From_Noisy_CVPR_2017_paper.pdf}

\bibitem[{Viola and Jones(2001{\natexlab{a}})}]{viola01cvpr}
Viola P, Jones M (2001{\natexlab{a}}) Rapid object detection using a boosted
  cascade of simple features. In: CVPR

\bibitem[{Viola and Jones(2001{\natexlab{b}})}]{viola01ijcv}
Viola P, Jones M (2001{\natexlab{b}}) Robust real-time object detection. IJCV

\bibitem[{Xu et~al.(2017)Xu, Zhu, Choy, and Fei-Fei}]{xu2017scenegraph}
Xu D, Zhu Y, Choy C, Fei-Fei L (2017) Scene graph generation by iterative
  message passing. In: Computer Vision and Pattern Recognition (CVPR)

\bibitem[{Yao and Fei-Fei(2010)}]{yao10cvpr-mutual}
Yao B, Fei-Fei L (2010) Modeling mutual context of object and human pose in
  human-object interaction activities. In: CVPR

\bibitem[{Zellers et~al.(2018)Zellers, Yatskar, Thomson, and
  Choi}]{zellers18cvpr}
Zellers R, Yatskar M, Thomson S, Choi Y (2018) Neural motifs: Scene graph
  parsing with global context. In: CVPR

\bibitem[{Zhang et~al.(2017{\natexlab{a}})Zhang, Kyaw, Chang, and
  Chua}]{zhang17cvpr-visual}
Zhang H, Kyaw Z, Chang SF, Chua TS (2017{\natexlab{a}}) Visual translation
  embedding network for visual relation detection. In: CVPR

\bibitem[{Zhang et~al.(2017{\natexlab{b}})Zhang, Kyaw, Yu, and
  Chang}]{zhang17iccv}
Zhang H, Kyaw Z, Yu J, Chang SF (2017{\natexlab{b}}) {PPR-FCN}: weakly
  supervised visual relation detection via parallel pairwise {R-FCN}. In: ICCV

\end{thebibliography}

\end{document}